\documentclass[runningheads]{llncs}

\usepackage[year=2026,ID=7924]{eccv}

\usepackage{eccvabbrv}

\usepackage{graphicx}
\usepackage{booktabs}
\usepackage{amsmath,amssymb}
\usepackage[table]{xcolor}
\usepackage{tabularx}
\usepackage{adjustbox}
\usepackage{wrapfig}
\usepackage{comment}
\usepackage[section]{placeins}
\usepackage{amsmath, amssymb}

\usepackage[accsupp]{axessibility}

\definecolor{orangefirst}{HTML}{F6A23A}
\definecolor{orangesecond}{HTML}{F9C06A}
\definecolor{orangethird}{HTML}{FBD8A0}

\usepackage[pagebackref,breaklinks,colorlinks,citecolor=eccvblue]{hyperref}

\begin{document}

\title{Control-DINO: Feature Space Conditioning for Controllable Image-to-Video Diffusion} 

\titlerunning{Control-DINO}
\authorrunning{Dominici, Deixelberger, Vardis and Steinberger}

\author{Edoardo A. Dominici \inst{1}\thanks{The authors contributed equally to this work.} \and Thomas Deixelberger\inst{2,3}\protect\footnotemark[1] \and Konstantinos Vardis\inst{1} \and Markus Steinberger\inst{2,3}}
\institute{Huawei, Switzerland \and Huawei, Austria \and Graz University of Technology, Austria\\
{\color{blue}\href{https://dedoardo.github.io/projects/control-dino}{https://dedoardo.github.io/projects/Control-DINO}}}

\maketitle

{
\centering
\makebox[\textwidth]{\includegraphics[width=\textwidth]{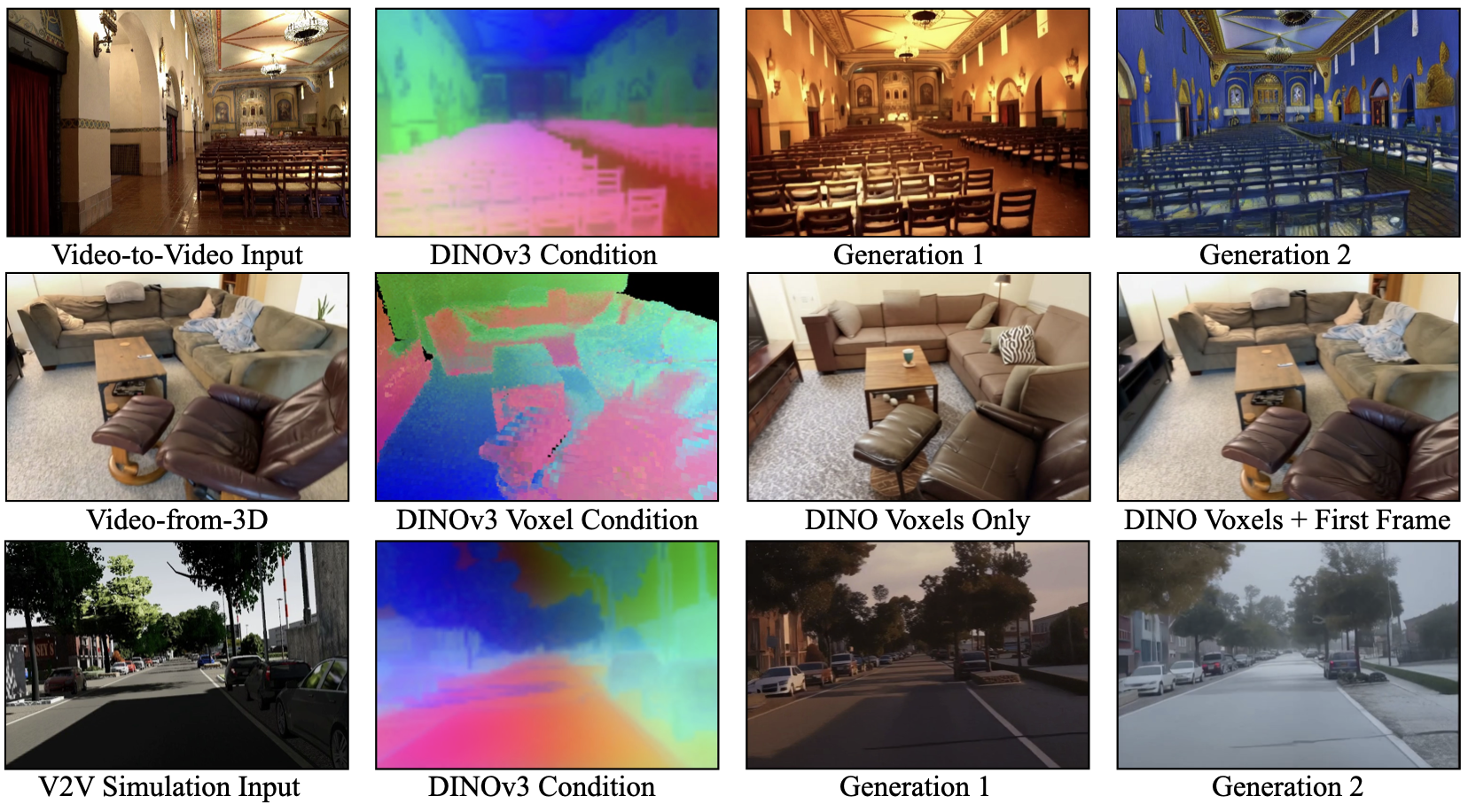}}
\captionof{figure}{We introduce a DINO based ControlNet for I2V diffusion models and apply it to video-to-video and video-from-3D tasks. It allows using dense DINOv3 features for structural and semantic guidance, in the image we show by row: (1) Style transfer on first frame propagated to later timesteps; (2) video-from-3D obtained by using only the voxel as conditioning with no first frame and with first frame conditioning; (3) VKITTI weather transfer where the first frame is obtained with FLUX.1  KREA~\cite{flux1kreadev2025}.}
\label{fig:teaser}
}

\begin{abstract}
Video diffusion models have recently been applied with success to problems in content generation, novel view synthesis, and, more broadly, world simulation. Many applications in generation and transfer rely on conditioning these models, typically through perceptual, geometric, or simple semantic signals, fundamentally using them as generative renderers. At the same time, high-dimensional features obtained from large-scale self-supervised learning on images or point clouds are increasingly used as a general-purpose interface for vision models. The connection between the two has been explored for subject specific editing, aligning and training video diffusion models, but not in the role of a dense conditioning signal for pretrained video diffusion models.
Features obtained through self-supervised learning like DINOv3, contain a lot of entangled information about style, lighting and semantics of the scene. This makes them great at reconstruction tasks but limits their generative capabilities. In this paper, we show how we can use the features for tasks such as video domain transfer and video-from-3D generation. We introduce a lightweight control architecture and training strategy that decouples appearance from other features that we wish to preserve, enabling robust control for appearance changes such as stylization and relighting. Furthermore, we show that low spatial resolution can be compensated by higher feature dimensionality, improving controllability in generative rendering from explicit spatial representations.
\end{abstract}

\section{Introduction}
\label{sec:intro}
Generative video models have advanced rapidly and now serve as strong foundations for editing, stylization, simulation, and world modeling~\cite{vdmsurvey2024,nvidia2025cosmosworldfoundationmodel,Robbyant2026LingBotWorld,Zhou2024DINOWM}. In many practical scenarios, we are not generating from scratch: we already have partial information about the scene, such as a reference video, sparse views, or 3D geometry, and we want generation to preserve this structure while still allowing controlled variation.

Most current control pipelines rely on low-level geometric or perceptual inputs, including edges, depth maps, colored points, pose skeletons and semantic masks~\cite{cvgsurvey2025,vdmsurvey2024}. These controls are often effective, but each captures only a narrow aspect of scene structure, and different tasks tend to require diverse control types. This fragmentation makes it difficult to build a unified conditioning interface that transfers cleanly across tasks like video restyling, video-from-3D, and long-horizon view synthesis.

\begin{figure}[t]
  \centering
  \setlength{\tabcolsep}{1pt}
  \renewcommand{\arraystretch}{0.5}
  \begin{tabular}{cccc}
    \tiny{Input image ($t{=}0$)} &
    \tiny{Condition ($T/2$)} &
    \tiny{Real training data ($T/2$)} &
    \tiny{Mixed training data ($T/2$)} \\
    \includegraphics[width=0.23\linewidth]{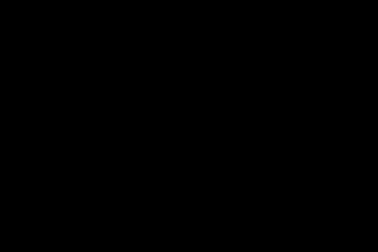} &
    \includegraphics[width=0.23\linewidth]{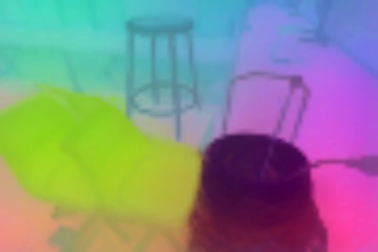} &
    \includegraphics[width=0.23\linewidth]{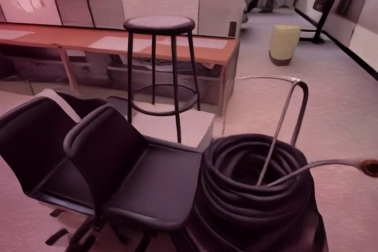} &
    \includegraphics[width=0.23\linewidth]{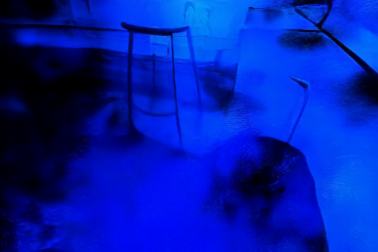}
  \end{tabular}
\caption{Naively training a ControlNet on top of an Image-to-Video model leaves DINO features prone to overfitting. We isolate the conditioning effect by dropping the first-frame conditioning, relying solely on the text embedding. With the prompt set to "blue", our method significantly reduces the ControlNet's bias toward the training data.}
   \label{fig:first_frame_overfit}

\end{figure}

At the same time, universal self-supervised features such as DINO~\cite{caron2021dino} are increasingly used as a common interface across vision models. Trained on large scale 2D and 3D data, these high-dimensional features encode semantics, geometry, and appearance while having some invariance to lighting thanks to the pretraining augmentations. They achieve strong performance in downstream tasks and have also shown promise as latent spaces for prediction~\cite{Zhou2024DINOWM,baldassarre2025back} or completion~\cite{Jevtic2025SceneDINO}. This makes them an attractive candidate for conditioning video diffusion models with richer and more reusable control signals than traditional low-level inputs.
 However, feature conditioning is not straightforward. The same expressivity that makes foundation features useful also makes them difficult to tame: appearance, geometry and other properties are often entangled, making the process challenging to generalize, especially when training data is limited (\autoref{fig:first_frame_overfit}). Recent work has shown promising subject-specific video transfer with DINO features~\cite{Huang2025DIVE}, but robust general-purpose conditioning remains underexplored.  

 We study DINO-based conditioning for video diffusion in a unified setting that spans both 2D and 3D feature sources. 
To extract structural guidance, the key idea is to use features of the original video as conditioning but train the diffusion model to denoise appearance-augmented versions of the same scene. Because the same features correspond to multiple appearances, the model learns to rely on spatial and semantic information rather than reproducing appearance cues. We further observe that low-spatial-resolution but high-dimensional features can provide robust control and similarly enable us to bridge 3D mesh renderings or other spatial representations to real videos.

Our contributions are as follows:
\begin{itemize}
  \item We introduce a general-purpose DINO-conditioned ControlNet for video diffusion that retains structural and semantic guidance while decoupling appearance from the conditioning signal.
  \item We apply the ControlNet to video transfer and video-from-3D tasks showing its versatility and robustness.
  \item We demonstrate new applications such as rendering low-resolution 3D voxel structures and connecting foundational 3D models to video diffusion.
\end{itemize}

\section{Related Work}
\label{sec:related}

\paragraph{Controllable diffusion via spatial conditioning.}
Modern generative video models~\cite{VDMHo2022,SVD2023} perform denoising in the latent space of a variational autoencoder, either with UNets or more recently diffusion transformers~\cite{Dit2023,yang2025cogvideox,Wan2024Wan22}. Within this paradigm, controllable generation is commonly achieved by injecting auxiliary signals into the denoising network interleaved between the transformer blocks. ControlNet~\cite{Zhang2023ControlNet} introduced a widely adopted design in which a trainable copy of the encoder injects residuals into the frozen generator via zero-initialized convolutions. Lighter-weight variants such as T2I-Adapter~\cite{Mou2023T2IAdapter} and IP-Adapter~\cite{Ye2023IPAdapter} offer more parameter-efficient or reference-based alternatives. Several works extend spatial conditioning to video by adapting image-level control architectures to temporal backbones~\cite{lin2025ctrladapter, sparsectrl2025, vace2025, Chen2023ControlAVideo, Liang2024FlowVid}, while alternative designs such as FullDiT~\cite{fulldit2025} tokenize all conditions for joint self-attention at the cost of full retraining. 
To date, these methods have predominantly been applied with geometric signals (edges, depth, pose), yet foundation model features remain underexplored as conditioning inputs. Recent methods that generate videos from 3D representations \cite{kim2025videofrom3d} also rely on edge conditioning for structure transfer. For single image domain transfer, a discriminator can also be adopted to finetune the model and decoder to generate samples closer to the target domain \cite{parmar2024one}.
 For a broader overview of controllable video generation, see~\cite{cvgsurvey2025}.

\paragraph{Self-supervised features for generation.}
Self-supervised vision models such as DINOv2~\cite{oquab2024dinov2} learn patch-level representations that capture object semantics and support tasks from segmentation to depth estimation, all from a single frozen backbone. DINOv3~\cite{simeoni2025dinov3} scales this family to denser features that preserve fine-grained local structure across model sizes. Multiple studies observe that self-supervised and diffusion features capture complementary information: the former provide high-level semantic structure, while the latter encode dense spatial coherence~\cite{Tang2023DIFT, stracke2025cleandift, luo2023diffhyper, zhang2023tale}. This motivates conditioning diffusion models on foundation features to supply semantic guidance beyond what the generation process captures on its own. REPA~\cite{yu2025repa} further shows that frozen DINOv2 features align with internal DiT representations, though as a training regularizer rather than an explicit conditioning signal. Directly conditioning on such features was first explored by Bordes et al.~\cite{bordes2022rcdm} using global DINO embeddings via FiLM~\cite{perez2018film} modulation. Graikos et al.~\cite{graikos2024learned} scale this to local, domain-specific SSL embeddings with cross-attention for large-image generation in medical and remote-sensing domains. AnyDoor~\cite{chen2024anydoor} injects DINOv2 patch tokens into a pretrained backbone for object identity, yet like the preceding works, targets images and not scene-level spatial control.
Self-supervised representations have also been extended to 3D, with methods such as Concerto~\cite{zhang2025concerto} and
Sonata~\cite{wu2025sonata} learning dense, multi-view consistent descriptors from point clouds that remain close to
DINO-style embeddings. We leverage this property by rendering such 3D features into 2D conditioning maps for our video diffusion
adapter.

\paragraph{Foundation features for video.}
Recent work has begun to explore foundation features as conditioning signals for video generation and editing. CAGE~\cite{davtyan2025cage} introduces a video model trained jointly with sparse DINOv2 patches to enable conditioning for unsupervised scene composition and object animation on synthetic scenes. DIVE~\cite{Huang2025DIVE} leverages DINOv2 features as implicit motion correspondences for subject-driven video editing through per-video MLP optimization, targeting appearance transfer rather than scene-level spatial control. Concurrent to our work, Driving with DINO (DwD)~\cite{chen2026dwd} injects DINOv3 features through a ControlNet-style branch into a frozen video backbone for sim-to-real driving video translation, introducing PCA-based spectral pruning tailored to its synthetic-to-real setting. However, none of these works targets dense foundation feature conditioning for different types of video transfer and rendering operations. We explore this direction with a ControlNet-style adapter that operates on spatially dense features without per-video optimization, and evaluate it across video transfer and 3D-guided video generation.

\section{Method}
\label{sec:method}

\begin{figure}[t]
  \centering
  \begin{minipage}[t]{0.49\linewidth}
    \centering
    \textbf{Training}\\[2pt]
    \includegraphics[width=\linewidth]{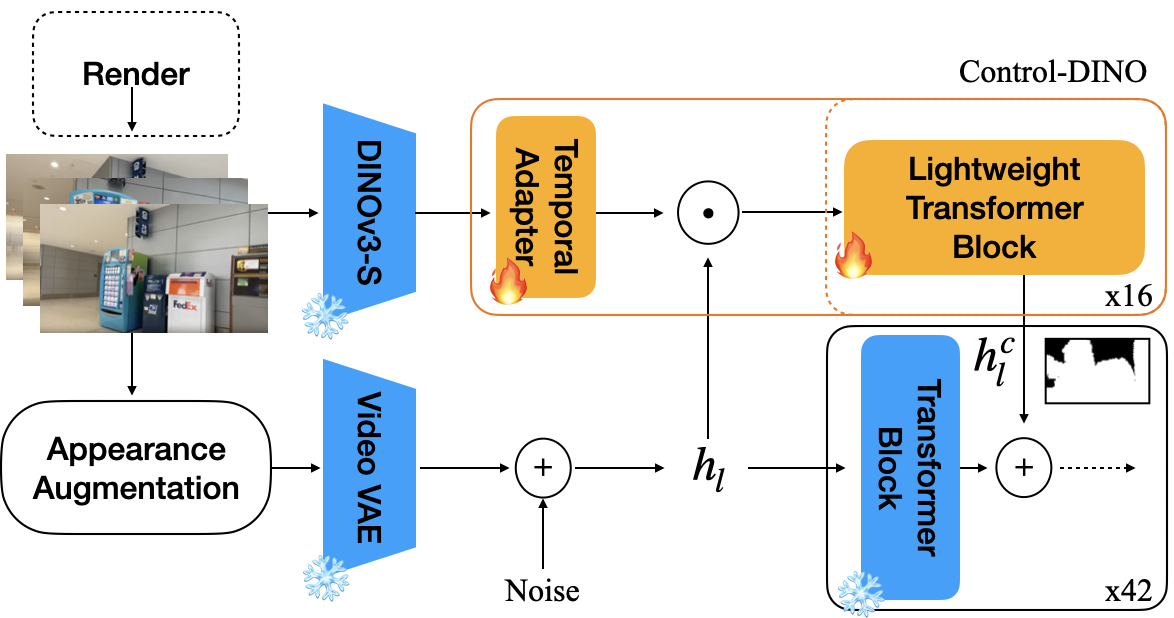}
  \end{minipage}\hspace{0.005\linewidth}%
  {\color{gray!45}\vrule width 0.5pt}%
  \hspace{0.005\linewidth}
  \begin{minipage}[t]{0.49\linewidth}
    \centering
    \textbf{Inference}\\[2pt]
    \includegraphics[width=\linewidth]{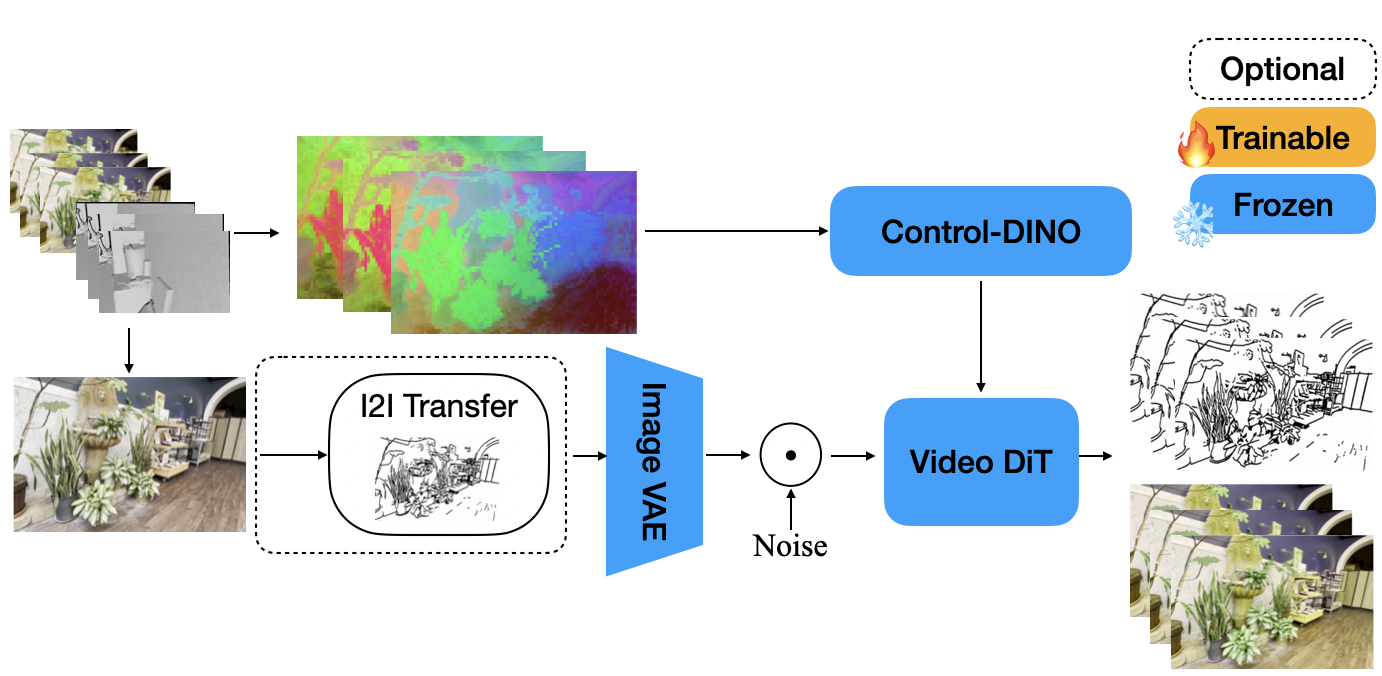}
  \end{minipage}
  \caption{(Left) During training, DINOv3 features from the original video condition the trainable Control-DINO branch, while the frozen
  backbone denoises appearance-augmented latents of the same scene. (Right) At inference, conditioning features (2D or 3D-rendered)
  guide generation through Control-DINO. A transferred first frame sets the target appearance.}
  \label{fig:training_inference}
\end{figure}

The goal of our model is to condition video diffusion on dense features while making their control practical (\autoref{fig:training_inference}). We achieve this by injecting foundation features into a frozen video transformer backbone through a ControlNet-style residual adapter (Sec.~\ref{subsec:conditioning}). We train on paired augmented data where the conditioning features come from the original video but the denoising target is a structurally similar version of the same scene, enabling the adapter to rely on geometry and semantics alone (Sec.~\ref{subsec:appearance}). A property of feature-level conditioning is
that it operates at low spatial resolution but high channel dimensionality, which means the same interface naturally extends to coarse 3D representations where pixel-level detail is unavailable. Since conditioned video models are increasingly used for video-from-3D generation~\cite{kim2025videofrom3d}, we describe how features can be obtained from 3D data and used as conditioning inputs to our adapter (Sec.~\ref{subsec:rendered_features}).

\subsection{Diffusion and Conditioning}
\label{subsec:conditioning}
Let $x_{1:T}$ denote an RGB video and $z_{1:T}$ its latent representation under the video VAE encoder. We follow the standard diffusion formulation:
\begin{align}
  z_t = \alpha_t z_0 + \sigma_t \epsilon, \qquad \epsilon \sim \mathcal{N}(0, I),
\end{align}
where $t$ denotes the diffusion timestep and subscript ranges $1{:}T$ index video frames.

The denoiser $v_\theta$, a video transformer, is trained to predict the velocity $v_t = \alpha_t \epsilon - \sigma_t z_0$ by minimizing $\mathcal{L} = \mathbb{E}_{z_0,\epsilon,t}\left[\|v_t - v_\theta(z_t, t, c)\|_2^2\right]$, where $c$ denotes conditioning.
We consider per-frame conditioning signals $c_{1:T}$ (edges, depth, semantic maps, or feature maps) pixel aligned with the output frames. Conditioning signals that are encoded with the video VAE (e.g., depth or edges) match the latent-space temporal resolution directly, while signals that bypass the VAE are rescaled spatially and temporally. With the spatial resolution adapted, we use a temporal adapter with the same causal architecture as the video encoder to map the conditioning temporal length.

We condition the video diffusion model by injecting residual connections from the output of the transformer block into its first 16 layers. This stream concatenates the noisy latent with the conditioning features along the feature dimension and processes them with a copy of the transformer block($\mathcal{Z}$) at each layer $\mathcal{Z}^{light}_\ell$, yielding the conditioning hidden states $h^{c}_\ell$.
The projections $\mathcal{Z}^{light}_\ell$ are zero initialized. We optionally modulate each injection by a mask $M \in [0,1]^{60 \times 90}$, broadcast over the feature dimension, where $M = 1$ recovers the unmasked case; $M$ can encode, for example, per-pixel visibility obtained by rendering geometry.
%
\begin{align}
  h_{\ell+1} = \mathcal{Z}_\ell(h_\ell) + M \odot \mathcal{Z}^{light}_\ell(h_\ell \cdot c), \qquad \ell = 1, \ldots, L.
\end{align}
Here $h_{\ell+1}$ is the current hidden state being denoised.

\subsection{Feature Conditioning} 
\label{subsec:appearance}
A side effect of DINO features encoding rich appearance information is that structural signals can be entangled with color, texture, and style. We therefore aim for the ControlNet to faithfully follow the spatial structure depicted in the conditioning features while remaining agnostic to the visual style of the source, enabling domain transfer applications. 

We extract dense patch-level features from a frozen DINOv3 vision encoder $D$ applied to each frame of the source video. For $x_{1:T} \in \mathbb{R}^{T \times 3 \times H \times W}$ we obtain per-frame feature maps
\begin{equation}
  F_i = D(x_i) \in \mathbb{R}^{384 \times 60 \times 90}, \quad i = 1, \ldots, T,
\end{equation}
for patch size $16$. 
To align conditioning with the video diffusion latent grid, we bicubic-upsample each input frame before DINO encoding so that $D$ operates at the transformer latent spatial scale, which coincides with the latent resolution of the video model's VAE. These feature maps serve as the conditioning signal $c = A(F_{1:T})$ for the ControlNet adapter described in Sec.~\ref{subsec:conditioning}, where A is the temporal adapter.

\paragraph{Training with augmentation.}
To force style-invariant conditioning, we construct training pairs $(F_{1:T}, \tilde{z}_{1:T})$, where $F_{1:T}$ are DINO features from the original video and $\tilde{z}_{1:T}$ are VAE latents of an appearance-augmented version of the same scene. We train with
$
  \mathcal{L} = \mathbb{E}_{\epsilon,t}\left[\|v_t - v_\theta(z_t, t, F_{1:T})\|_2^2\right],
$
where $z_t$ is obtained from $\tilde{z}_{1:T}$ using the forward diffusion process in Sec.~\ref{subsec:conditioning}.
During training we employ a hierarchy of augmentation strategies at varying levels of severity to mimic appearance changes expected at test time: (i) photometric augmentations (hue, grayscale, saturation, brightness, contrast, gamma, etc.) applied uniformly across frames; (ii) five GAN-based neural style transfer methods (cartoon, anime, etc.)~\cite{Chen2018CartoonGAN,Tachibana2018AnimeGANv2}; and (iii) small blur to simulate quality degradations. During training, we sample from a mixture uniformly distributed among real data and these three augmentation groups, all paired with features extracted from the original unstylized video. At inference, we apply a more generic transfer function only to the first frame while conditioning on features from the original source domain (real or rendered)~\cite{wang2024instantstylefreelunchstylepreserving,zhang2025scaling}. As long as the first-frame transfer preserves structure and semantics, the model can follow the target appearance consistently over time. Because the same features correspond to multiple appearances, the model learns to rely on spatial and semantic information rather than appearance cues in the features. We compare this strategy against PCA-based feature projection in Sec.~\ref{subsec:ablations}.

\subsection{3D Feature Conditioning}
\label{subsec:rendered_features}

Although DINO features are extracted on coarse patch grids, each token is high-dimensional and semantically rich, such that an $h \times w$ grid with $C{=}384$ channels (here $60 \times 90$) preserves object identity, part-level layout, and scene semantics even when fine pixel detail is absent. This makes the features complementary to geometry controls: geometry anchors \emph{where} content appears, while DINO channels encode \emph{what} it is and how regions relate semantically. We therefore align both signals at the same low spatial resolution and condition on them jointly. In our method the geometric prior is a voxel grid, which is stable across views and time: voxel geometry enforces camera-consistent structure while DINO features stabilize subject and background semantics. Because both inputs are low-resolution and high-level, conditioning remains inexpensive while still guiding temporally coherent generation. High-resolution geometry is not required, and low-resolution voxel cues from LiDAR reconstructions or occupancy prediction models suffice.

\begin{figure}[t]
  \centering
  \setlength{\tabcolsep}{1pt}
  \renewcommand{\arraystretch}{0.5}
    \scriptsize 
  \textbf{ScanNet++ Scene} (09c1414f1b / 0d2ee665be)\\[2pt]
  \begin{tabular}{cccccc}
    \raisebox{4ex}{\rotatebox{90}{\tiny $t{=}0$}} &
    \includegraphics[width=0.18\linewidth]{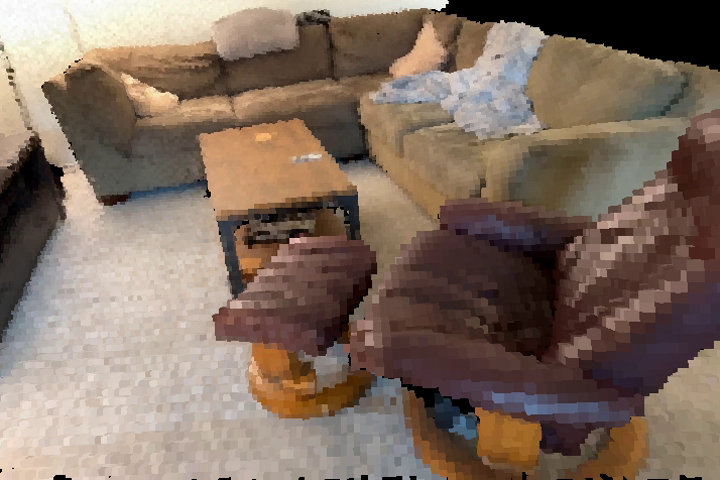} &
    \includegraphics[width=0.18\linewidth]{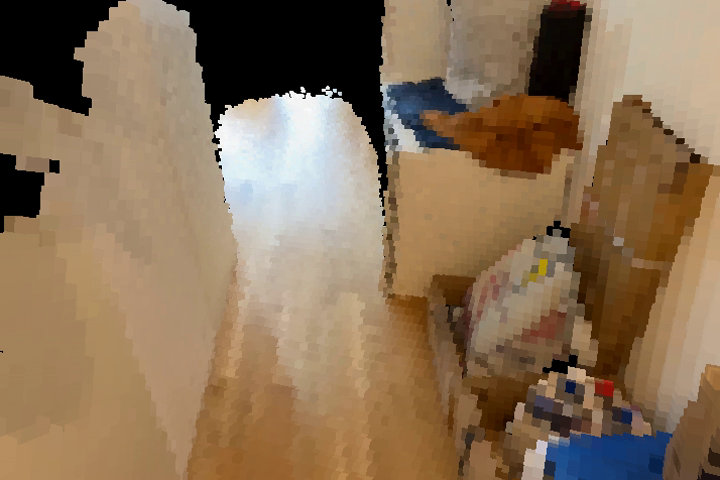} &
     \includegraphics[width=0.18\linewidth]{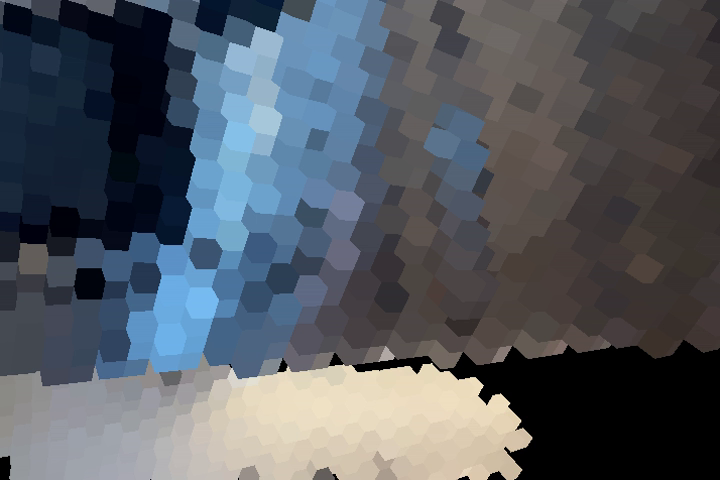} &
    \includegraphics[width=0.18\linewidth]{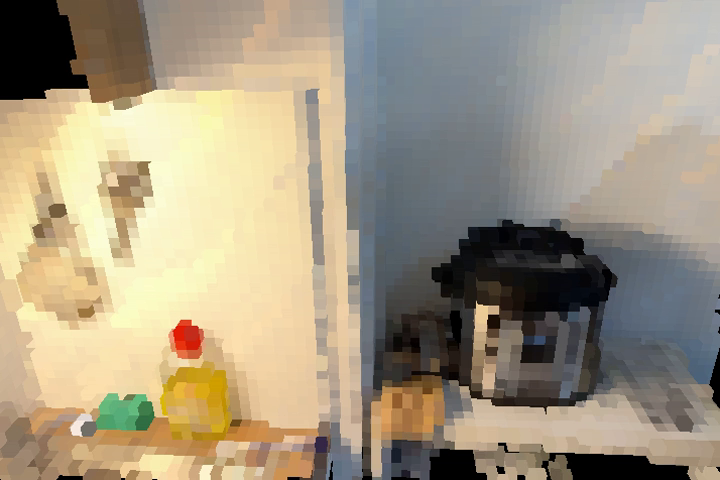} &
        \includegraphics[width=0.18\linewidth]{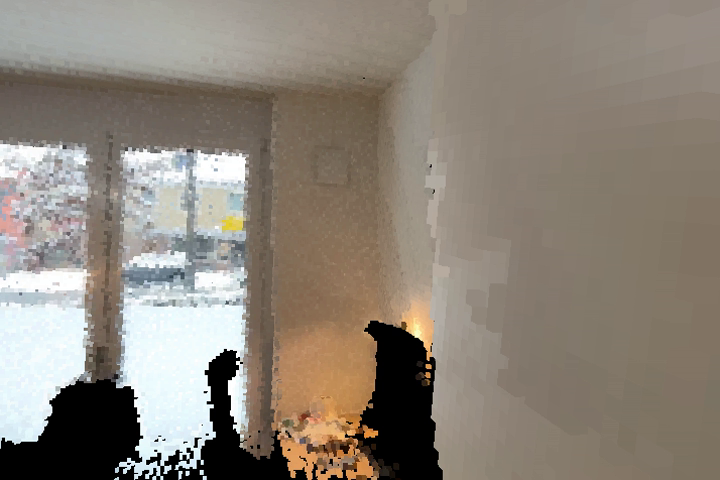}
        \\
        \raisebox{4ex}{\rotatebox{90}{\tiny $t{=}0$}} &
    \includegraphics[width=0.18\linewidth]{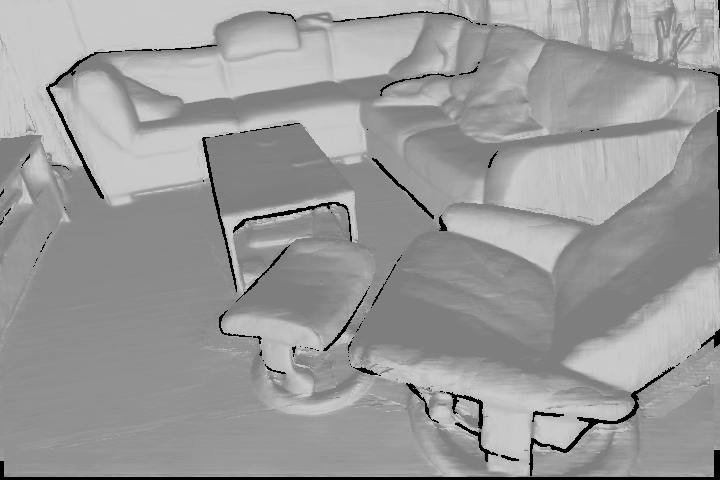} &
    \includegraphics[width=0.18\linewidth]{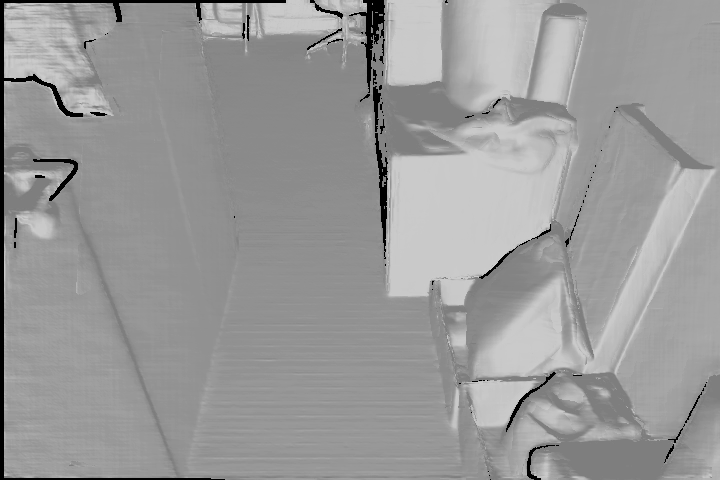} &
    \includegraphics[width=0.18\linewidth]{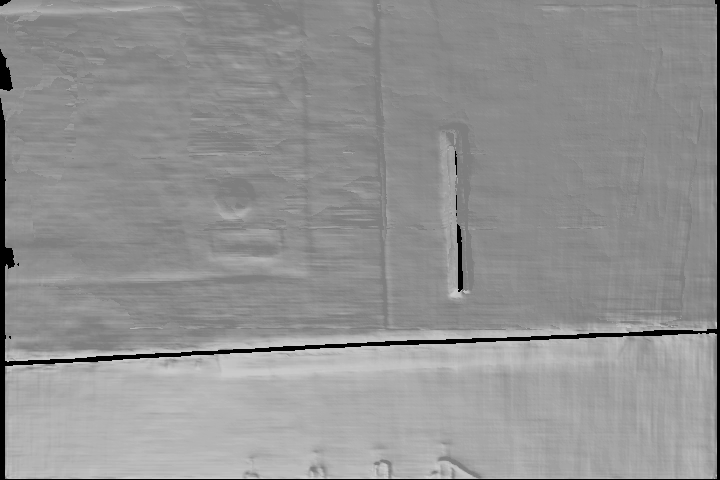} &
    \includegraphics[width=0.18\linewidth]{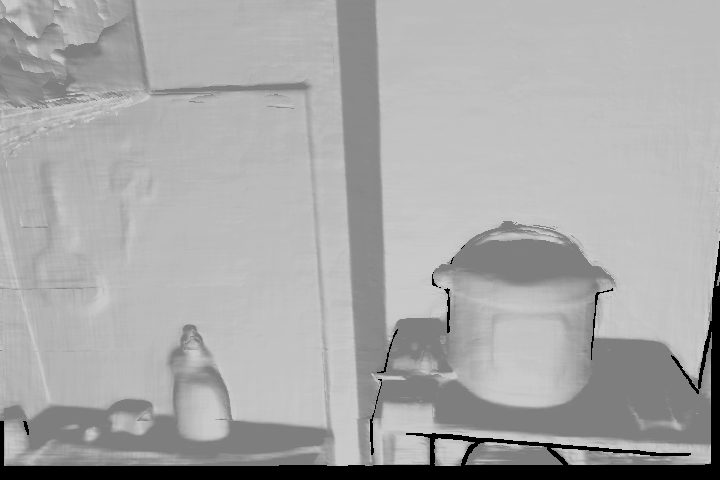} &
        \includegraphics[width=0.18\linewidth]{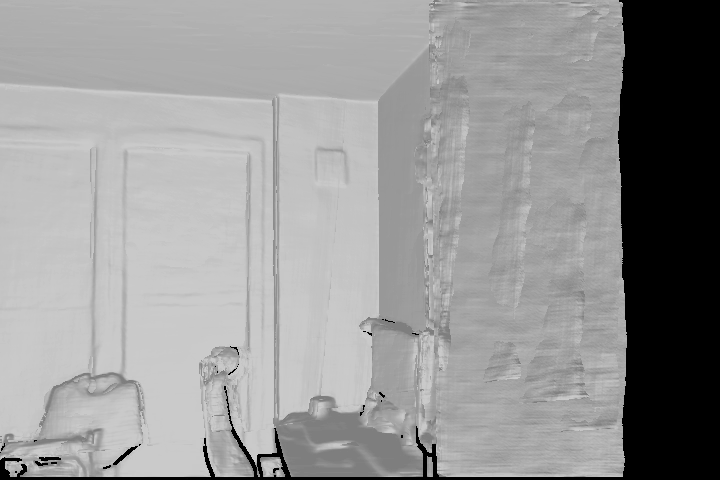}
  \end{tabular}
  \caption{3D structure visualization. Top: RGB voxels. Bottom: mesh renderings.}
  \label{fig:visualisation_3d_structure}
\end{figure}

\paragraph{DINO-like features from uncolored geometry.}
Importantly, useful feature priors do not require RGB textures at inference time. Recent 2D-3D self-supervised methods, such as Concerto~\cite{zhang2025concerto}, can produce dense geometric descriptors from uncolored geometry like point clouds that remain close to DINO-style semantic embeddings after projection. We therefore treat these geometry-derived descriptors as a drop-in conditioning signal: they can be rendered to the target view, projected to the ControlNet token grid, and fused with standard controls exactly as in \autoref{fig:visualisation_3d_structure}. This enables controllable generation in settings where only geometry is available, while retaining semantic consistency typically associated with appearance-based foundation features.

\section{Experiments}
\label{sec:experiments}

\begin{figure*}[t]
  \centering
  \setlength{\tabcolsep}{1pt}
  \renewcommand{\arraystretch}{0.5}
  \newcommand{\panel}[1]{\includegraphics[width=0.125\linewidth]{figures/comparison_styles0/#1}}
  \begin{tabular}{cccccccc}
    & \tiny{Cond.} & \multicolumn{3}{c}{\tiny{Painting (InstantStyle)}}
      & \multicolumn{3}{c}{\tiny{Relighting}} \\
    & & \tiny{$t{=}0$} & \tiny{$t{=}T/2$} & \tiny{$t{=}T$}
      & \tiny{$t{=}0$} & \tiny{$t{=}T/2$} & \tiny{$t{=}T$} \\
    \raisebox{1ex}{\rotatebox{90}{\tiny C-DINO}} &
      \panel{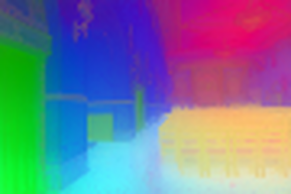} &
      \panel{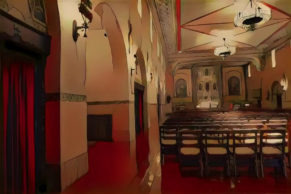} & \panel{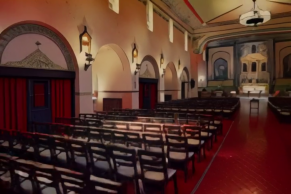} & \panel{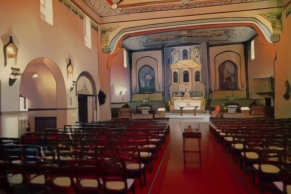} &
      \panel{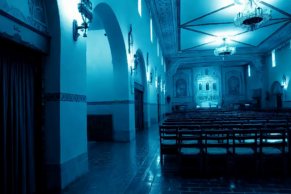} & \panel{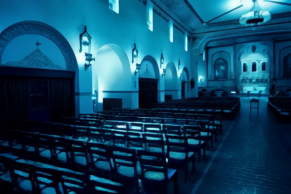} & \panel{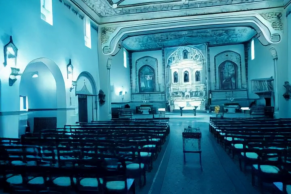} \\
    \raisebox{1ex}{\rotatebox{90}{\tiny Canny}} &
      \panel{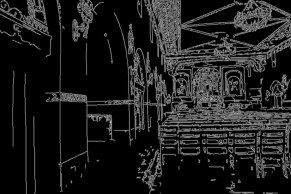} &
      \panel{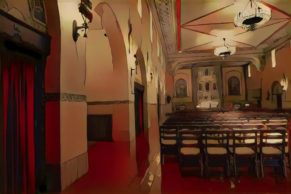} & \panel{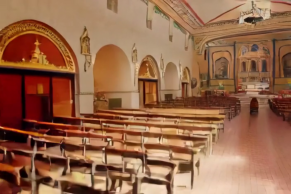} & \panel{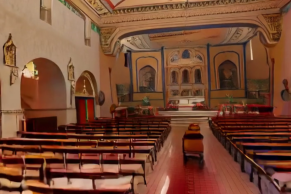} &
      \panel{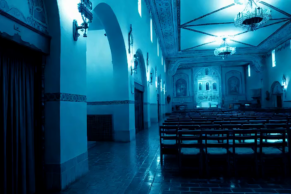} & \panel{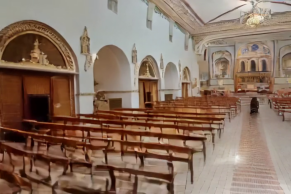} & \panel{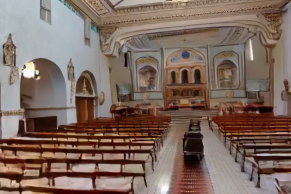} \\
    \raisebox{1ex}{\rotatebox{90}{\tiny Depth}} &
      \panel{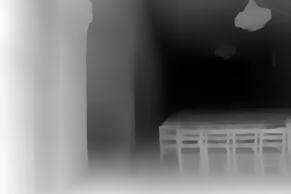} &
      \panel{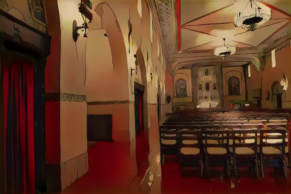} & \panel{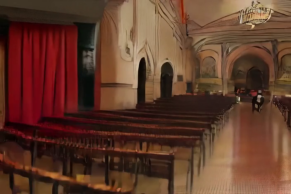} & \panel{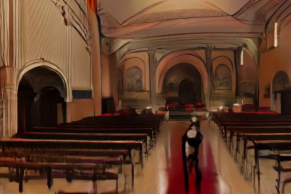} &
      \panel{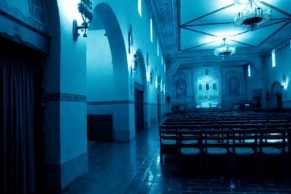} & \panel{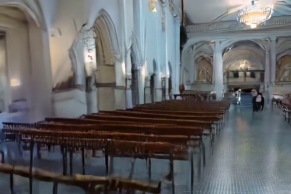} & \panel{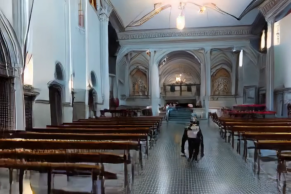} \\
    \raisebox{1ex}{\rotatebox{90}{\tiny AnyV2V}} &
      \panel{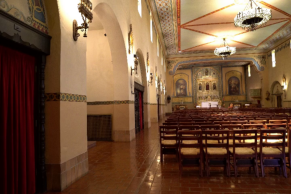} &
      \panel{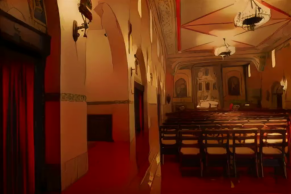} & \panel{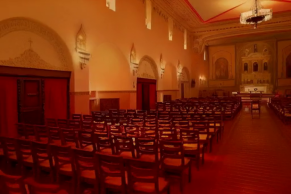} & \panel{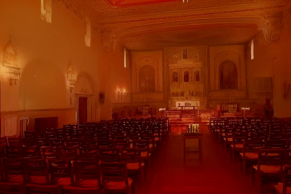} &
      \panel{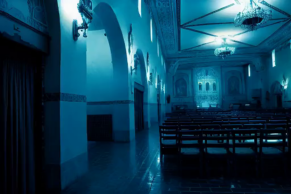} & \panel{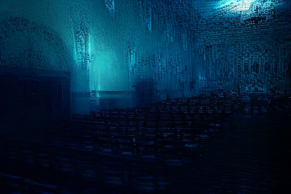} & \panel{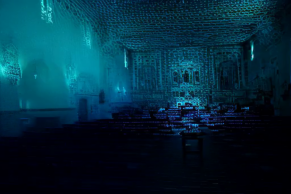} \\
  \end{tabular}
  \caption{We show here results for DINO stylization and dramatic light changes (GT Frame is in the bottom left of the image). Our method is able to strongly retain the original geometry and semantics while allowing for appearance and color to change. For brevity, \textit{Canny} refers to Wan 2.2 Fun Canny and \textit{Depth} to Wan 2.2 Fun Depth.}
  \label{fig:style_aug_ablation}
\end{figure*}

\begin{table}[t]
\centering
\caption{Quantitative comparison on T\&T Appearance Augmented. All models are trained on DL3DV and evaluated on T\&T (for two frame sampling strides) with a combination of 15 lighting or stylistic variations per scene different from the training data. Overall our method produces the most 3D consistent videos, especially for longer camera trajectories, while maintaining a competitive CLIP similarity with the first frame in the sequence. Our same model but trained only on real data performs much worse in style following.}
\label{tab:benchmark_tt}
\resizebox{\linewidth}{!}{
\begin{tabular}{lccccccc}
\multicolumn{8}{c}{\textbf{ Tanks and Temples - Out of domain - Transfer}} \\
\toprule
& \multicolumn{2}{c}{Consistency} & \multicolumn{3}{c}{Video Quality (VBench)} & \multicolumn{1}{c}{CLIP Similarity} & \multicolumn{1}{c}{COLMAP} \\
\cmidrule(lr){2-3} \cmidrule(lr){4-6} \cmidrule(lr){7-7} \cmidrule(lr){8-8}
Method & Subject $\uparrow$ & Background $\uparrow$ & Motion Smoothness $\uparrow$ & Aesthetic $\uparrow$ & Imaging $\uparrow$ & CLIP Sim. $\uparrow$ & Registration $\uparrow$ \\
\midrule
\multicolumn{8}{c}{\textbf{Stride = 8}} \\
\midrule
Wan 2.2 Fun Depth~\cite{Wan2024Wan22}                   & \cellcolor{orangesecond}0.8853 & \cellcolor{orangesecond}0.8916 & 0.9421         & 0.5388         & 0.6880         & 0.4626         & 0.8430         \\
Wan 2.2 Fun Canny~\cite{Wan2024Wan22}                   & 0.8778         & 0.8848         & 0.9599         & 0.5325         & \cellcolor{orangefirst}0.7568 & \cellcolor{orangefirst}0.5020         & \cellcolor{orangefirst}0.9770 \\
AnyV2V~\cite{ku2024anyvv}                              & 0.8124         & 0.8818         & 0.9258         & 0.5390         & 0.6590         & \cellcolor{orangesecond}0.4875 & 0.5470         \\
\midrule
Control-DINO Real Only                                             & 0.8767             & 0.8465        & \cellcolor{orangefirst}0.9680            & \cellcolor{orangesecond}0.5533             & \cellcolor{orangesecond}0.7372             & 0.4408             & 0.8840             \\
Control-DINO                                      & \cellcolor{orangefirst}0.8913 & \cellcolor{orangefirst}0.9060 & \cellcolor{orangesecond}0.9640 & \cellcolor{orangefirst}0.5610 & 0.7146 & 0.4717 & \cellcolor{orangesecond}0.9530 \\

\midrule
\multicolumn{8}{c}{\textbf{Stride = 16}} \\
\midrule
Wan 2.2 Fun Depth~\cite{Wan2024Wan22}                   & 0.8222         & 0.8527         & \cellcolor{orangefirst}0.9410 & 0.5044         & 0.6497         & 0.4804 & 0.4956         \\
Wan 2.2 Fun Canny~\cite{Wan2024Wan22}                   & 0.8226         & 0.8512         & 0.9357         & 0.5104         & \cellcolor{orangefirst}0.7235 & \cellcolor{orangefirst}0.4630         & 0.5918         \\
AnyV2V~\cite{ku2024anyvv}                              & 0.7621         & 0.8579         & 0.9162         & 0.5072         & 0.6299         & \cellcolor{orangesecond}0.4864 & 0.1875         \\
\midrule
Control-DINO Real Only                                  & \cellcolor{orangefirst}0.8544 & \cellcolor{orangefirst}0.8906 & \cellcolor{orangesecond}0.9396         & \cellcolor{orangesecond}0.5485 & 0.6908 & 0.4406 & \cellcolor{orangesecond}0.9911         \\
Control-DINO                                            & \cellcolor{orangesecond}0.8461 & \cellcolor{orangesecond}0.8862 & 0.9361         & \cellcolor{orangefirst}0.5508 & \cellcolor{orangesecond}0.7109 & 0.4707 & \cellcolor{orangefirst}0.9932 \\
\bottomrule
\end{tabular}
}
\end{table}

\begin{table}[t]
\centering
\caption{We compare against the only publicly available segmentation ControlNet. Our method produces competitive results, especially under larger camera motions, even against models larger in scale, training data, and architecture.}
\label{tab:i0t}
\resizebox{\linewidth}{!}{
\begin{tabular}{lccccccc}
\multicolumn{8}{c}{\textbf{Tanks and Temples - Out of domain - Transfer}} \\
\toprule
& \multicolumn{2}{c}{Consistency} & \multicolumn{3}{c}{Video Quality (VBench)} & \multicolumn{1}{c}{CLIP Similarity} & \multicolumn{1}{c}{Coverage} \\
\cmidrule(lr){2-3} \cmidrule(lr){4-6} \cmidrule(lr){7-7} \cmidrule(lr){8-8}
Method & Subject $\uparrow$ & Background $\uparrow$ & Motion Smoothness $\uparrow$ & Aesthetic $\uparrow$ & Imaging $\uparrow$ & CLIP Sim. $\uparrow$ & RegRate $\uparrow$ \\
\midrule
Cosmos-Segmentation~\cite{nvidia2025cosmostransfer1conditionalworldgeneration} (S=8) & \cellcolor{orangefirst}0.9128 & \cellcolor{orangefirst}0.9191 & \cellcolor{orangefirst}0.9678 & 0.4987 & 0.6666 & \cellcolor{orangefirst}0.7306 & 0.9374 \\
Control-DINO (S=8) & \cellcolor{orangesecond}0.8913 & \cellcolor{orangesecond}0.9060 & \cellcolor{orangesecond}0.9640 & \cellcolor{orangefirst}0.5610 & \cellcolor{orangefirst}0.7146 & \cellcolor{orangesecond}0.4717 & \cellcolor{orangesecond}0.9530 \\
\midrule
Cosmos-Segmentation~\cite{nvidia2025cosmostransfer1conditionalworldgeneration} (S=16) & 0.8315 & \cellcolor{orangefirst}0.8958 & \cellcolor{orangefirst}0.9475 & 0.4547 & 0.6190 & \cellcolor{orangefirst}0.7147 & 0.7755 \\
Control-DINO (S=16) & \cellcolor{orangefirst}0.8461 & 0.8862 & 0.9361 & \cellcolor{orangefirst}0.5508 & \cellcolor{orangefirst}0.7109 & \cellcolor{orangesecond}0.4707 & \cellcolor{orangefirst}0.9932 \\
\bottomrule
\end{tabular}
}
\end{table}

\begin{table}[t]
\centering
\caption{We illustrate the tradeoff between structural control signals and generation determinism, by analyzing how different seeds affect generation. Features jointly encode geometry and appearance, constraining the output more than other signals. }
\label{tab:benchmark}
\resizebox{\linewidth}{!}{
\begin{tabular}{l*{4}{c}*{3}{c}}
\multicolumn{8}{c}{\textbf{DL3DV140 - In domain - Real}} \\
\toprule
& \multicolumn{4}{c}{Reconstruction} & \multicolumn{3}{c}{Video Quality (VBench)} \\
\cmidrule(lr){2-5} \cmidrule(lr){6-8}
Method & FID $\downarrow$ & Latent Var. & CLIP Var.  & mIOU $\uparrow$ & Flickering $\uparrow$ & Aesthetic $\uparrow$ & Imaging $\uparrow$ \\
\midrule
Wan 2.2 Depth~\cite{Wan2024Wan22}                   & 47.98          & 0.5115         & 0.0151         & \cellcolor{orangesecond}0.0362         & \cellcolor{orangesecond}0.9192         & 0.4343         & 0.5469         \\
Wan 2.2 Canny~\cite{Wan2024Wan22}                   & \cellcolor{orangesecond}37.65          & 0.4787         & 0.0150         & 0.0322         & 0.9154         & \cellcolor{orangesecond}0.4496         & \cellcolor{orangesecond}0.6169         \\
\midrule
Control-DINO              & \cellcolor{orangefirst}20.23          & 0.1944         & 0.0051         & \cellcolor{orangefirst}0.2992         & 0\cellcolor{orangefirst}.9208         & \cellcolor{orangefirst}0.5133 & \cellcolor{orangefirst}0.6283 \\
\bottomrule
\end{tabular}
}
\end{table}

\begin{table}[t]
\centering
\caption{Video-from-3D comparison on ScanNet++. We evaluate geometry-driven 3D-derived conditioning methods. We compare quantitative results for image-space baselines (Wan and 2D Control-DINO) and report results for 3D-derived methods.}
\label{tab:video_from_3d}
\resizebox{\linewidth}{!}{
\begin{tabular}{lcccccccccc}
\multicolumn{11}{c}{\textbf{ScanNet++ - In domain - Video-from-3D}} \\
\toprule
& \multicolumn{4}{c}{Reconstruction} & \multicolumn{6}{c}{Video Quality (VBench)} \\
\cmidrule(lr){2-5} \cmidrule(lr){6-11}
Method & PSNR $\uparrow$ & SSIM $\uparrow$ & LPIPS $\downarrow$ & FID $\downarrow$ & Subject $\uparrow$ & Background $\uparrow$ & Flickering $\uparrow$ & Motion Smooth $\uparrow$ & Aesthetic $\uparrow$ & Imaging $\uparrow$ \\
\midrule
\multicolumn{11}{c}{\textbf{Image-space baselines}} \\
\midrule
Wan 2.2 Depth~\cite{Wan2024Wan22}  & 18.23 & 0.7237 & 0.2782 & 92.1072 & 0.9027 & 0.9234 & \cellcolor{orangefirst}0.9680 & 0.9881 & 0.4111 & 0.5714 \\
Wan 2.2 Canny~\cite{Wan2024Wan22}  & 19.82 & 0.7968 & 0.1987 & 72.1929 & 0.9001 & 0.9251 & 0.9590 & 0.9894 & 0.4100 & \cellcolor{orangefirst}0.6625 \\
Control-DINO from 2D (reference) & \cellcolor{orangefirst}27.93 & \cellcolor{orangefirst}0.9084 & \cellcolor{orangefirst}0.0963 & \cellcolor{orangefirst}38.5853 & 0.9183 & 0.9256 & \cellcolor{orangesecond}0.9662 & \cellcolor{orangefirst}0.9931 & 0.4187 & 0.5821 \\
\midrule
\multicolumn{11}{c}{\textbf{3D-derived methods (ours)}} \\
\midrule
Control-DINO from Mesh & 21.41 & 0.8010 & 0.1807 & 67.1730 & 0.9145 & 0.9242 & 0.9647 & 0.9919 & \cellcolor{orangesecond}0.4198 & 0.5852 \\
Control-DINO from Voxel & \cellcolor{orangesecond}22.80 & \cellcolor{orangesecond}0.8153 & \cellcolor{orangesecond}0.1572 & \cellcolor{orangesecond}65.3810 & \cellcolor{orangefirst}0.9216 & \cellcolor{orangefirst}0.9293 & 0.9658 & \cellcolor{orangesecond}0.9928 & 0.4188 & 0.5979 \\
Control-DINO Concerto & 21.19 & 0.7805 & 0.1752 & 67.9680 & \cellcolor{orangesecond}0.9195 & 0.9263 & 0.9638 & 0.9926 & \cellcolor{orangefirst}0.4200 & 0.6120 \\
Control-DINO Concerto (inpainted) & 19.66 & 0.7542 & 0.2060 & 78.8411 & 0.9122 & \cellcolor{orangesecond}0.9267 & 0.9635 & 0.9917 & 0.4184 & \cellcolor{orangesecond}0.6326 \\

\bottomrule
\end{tabular}
}
\end{table}

\subsection{Datasets}

Our work combines video diffusion conditioning from both 2D and 3D representations, with a focus on generative rendering. To emphasize generation quality and cross-dataset generalization, we train on DL3DV~\cite{Ling_dl3dv} and evaluate primarily on Tanks and Temples (T\&T)~\cite{Knapitsch2017TanksAndTemples}, which provides longer and smoother camera trajectories. From these trajectories we sample two evaluation settings, stride 8 and stride 16. For \autoref{tab:benchmark}, we also report results on DL3DV140~\cite{Ling_dl3dv}. In total, we train on about 4.5M unique frames and evaluate on 10k unique frames. The larger camera motion in Tanks and Temples stresses both generative quality and conditioning robustness. For 3D rendering experiments, we fine-tune the same base model on ScanNet++~\cite{yeshwanth2023scannetpp} and evaluate on parts of its official validation split. On ScanNet++, we sample training clips at stride 1, yielding nearly 8k clips, and evaluate video-from-3D pipelines that construct DINO and DINO-like conditioning features from explicit 3D geometry (\autoref{tab:video_from_3d}).

\subsection{Model Details}
\label{subsec:model_details}
We implement our approach on CogVideoX-5B-I2V~\cite{yang2025cogvideox} with residual connections~\cite{Zhang2023ControlNet} into the first $L{=}16$ transformer blocks, using dense DINOv3-ViT-S/16 patch features as input. These features are bicubically upscaled by a factor of 2 and then encoded. To align the latent space dimension to the one of the hidden state. The temporal adapter $A$ is composed of two causal Conv3D layers that temporally compress the sequence from 49 to 13 frames, matching the VAE latent temporal resolution (and architecture) and the result is concatenated channel-wise with the VAE latents before being fed into the conditioning branch. We set the hidden dimension of the conditioning branch to $256$ to further reduce the risk of overfitting. Zero-initialized output projections inject additive residuals at each layer of the base CogVideoX transformer, ensuring stable training from initialization. At inference time, we scale the conditioning residuals by $0.8$.

We train the temporal adapter and lightweight transformer blocks from scratch with AdamW ($\beta_1{=}0.9$, $\beta_2{=}0.95$) at a peak learning rate of $2{\times}10^{-4}$ with cosine annealing and 500 warmup steps, a batch size of 8, and gradient clipping at 1.0. We apply conditioning dropout during training, setting the conditioning input to zero with probability $0.1$. In the supplemental we show results for different video models.

\subsection{Metrics}
For the main benchmark on out-of-domain evaluation (\autoref{tab:benchmark_tt}) we assess video quality using VBench~\cite{VBench}, reporting consistency, motion and imaging scores but also measure 3D consistency through COLMAP-based reconstruction errors and registration success rate. We also quantify style transfer fidelity via CLIP style similarity with the first frame image. On the validation split of the training set (\autoref{tab:benchmark}) we additionally evaluate model generalizability by looking at the distribution of the output latent space and features. For video-from-3D, we additionally report standard reconstruction metrics (\autoref{tab:video_from_3d}).

\subsection{Baselines}
Our main baselines use alternative conditioning signals on image-to-video diffusion models with similar architecture and scale. We use the official Wan2.2-Fun-5B-Control (I2V) checkpoints with Canny-edge and depth conditioning~\cite{Wan2024Wan22}. We also compare against the publicly available semantic ControlNet from Cosmos-Transfer-2.5~\cite{nvidia2025cosmostransfer1conditionalworldgeneration}, while noting that Cosmos base model and ControlNet are substantially larger models.
For domain transfer, our setup is closest to AnyV2V~\cite{ku2024anyvv}, which uses DDIM-inverted latents~\cite{song2021ddim} for appearance transfer during diffusion. For first-frame appearance transfer baselines, we use IC-Light for relighting~\cite{zhang2025scaling}, InstantStyle for style transfer~\cite{wang2024instantstylefreelunchstylepreserving} picking 10 styles from the default ones, and five additional cartoon neural styles that were excluded from training.

\subsection{Experiments and Results}
\subsubsection{Representation and transfer}

\begin{figure}[t]
  \centering
  \setlength{\tabcolsep}{1pt}
  \renewcommand{\arraystretch}{1.0}
  \scriptsize
  \begin{tabular}{@{}c c c c@{}}
    & \tiny{Barn} & \tiny{Bear} & \tiny{Horse}\\
    \raisebox{4ex}{\rotatebox{90}{\tiny Input frame}} &
      \includegraphics[width=0.31\linewidth]{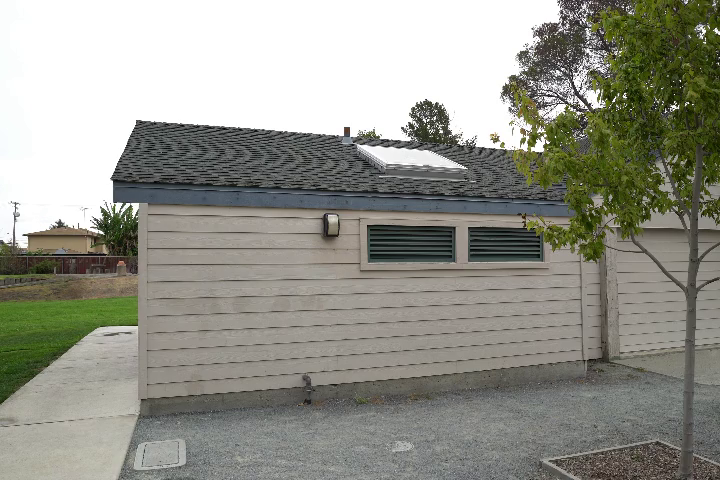} &
      \includegraphics[width=0.31\linewidth]{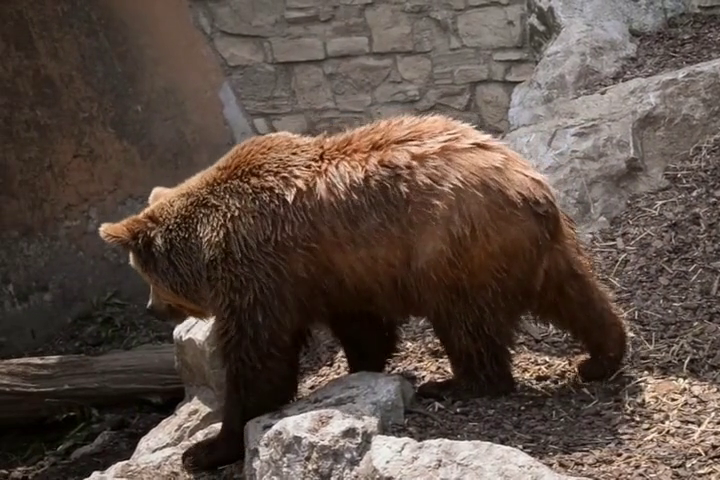} &
      \includegraphics[width=0.31\linewidth]{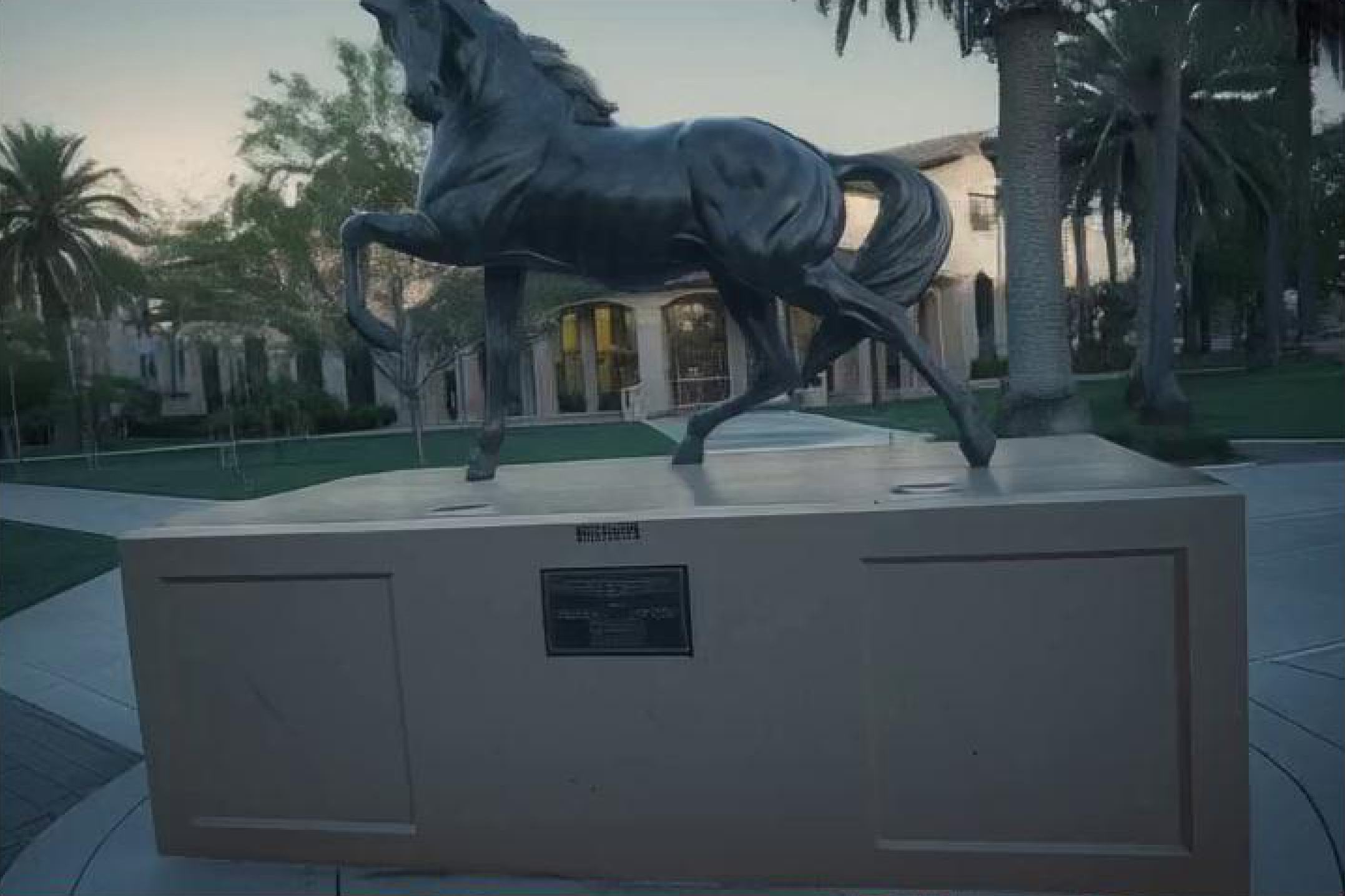}\\
    \raisebox{3.5ex}{\rotatebox{90}{\tiny Light-a-Video}} &
      \includegraphics[width=0.31\linewidth]{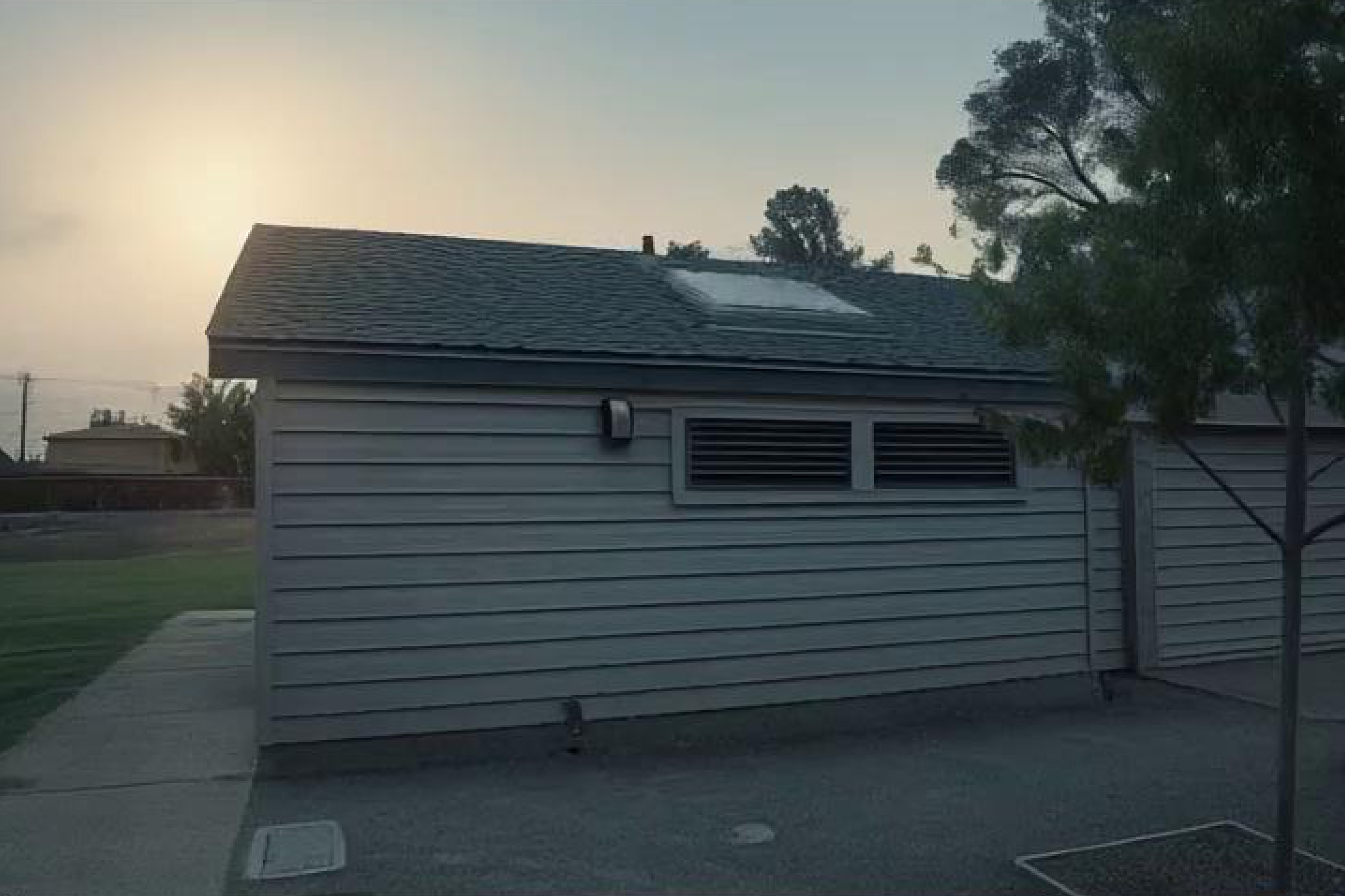} &
      \includegraphics[width=0.31\linewidth]{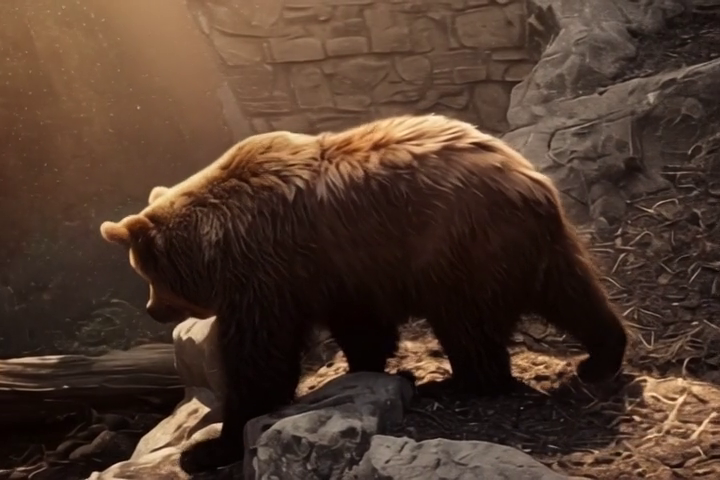} &
      \includegraphics[width=0.31\linewidth]{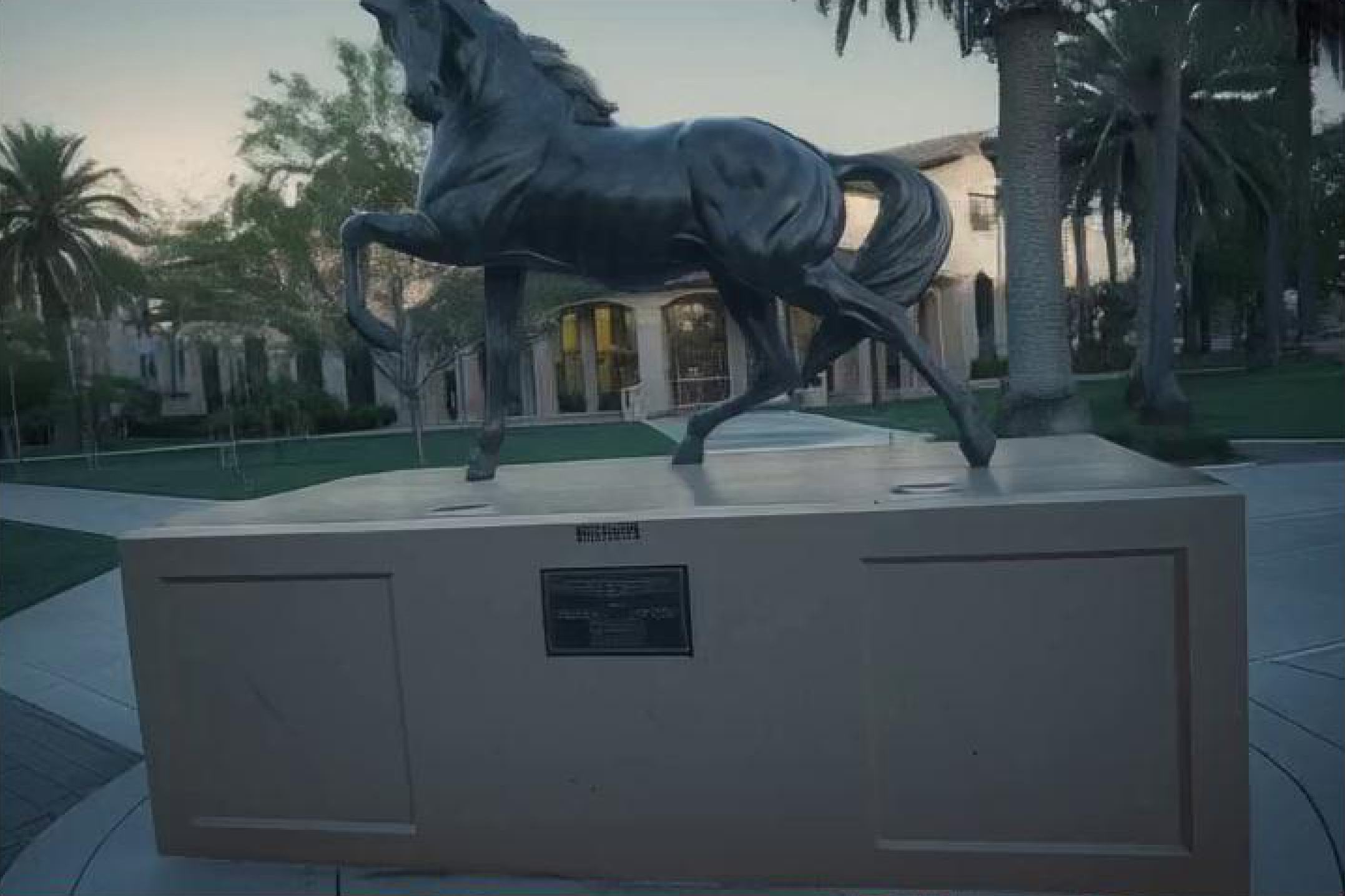}\\
    \raisebox{3.5ex}{\rotatebox{90}{\tiny Control-DINO}} &
      \includegraphics[width=0.31\linewidth]{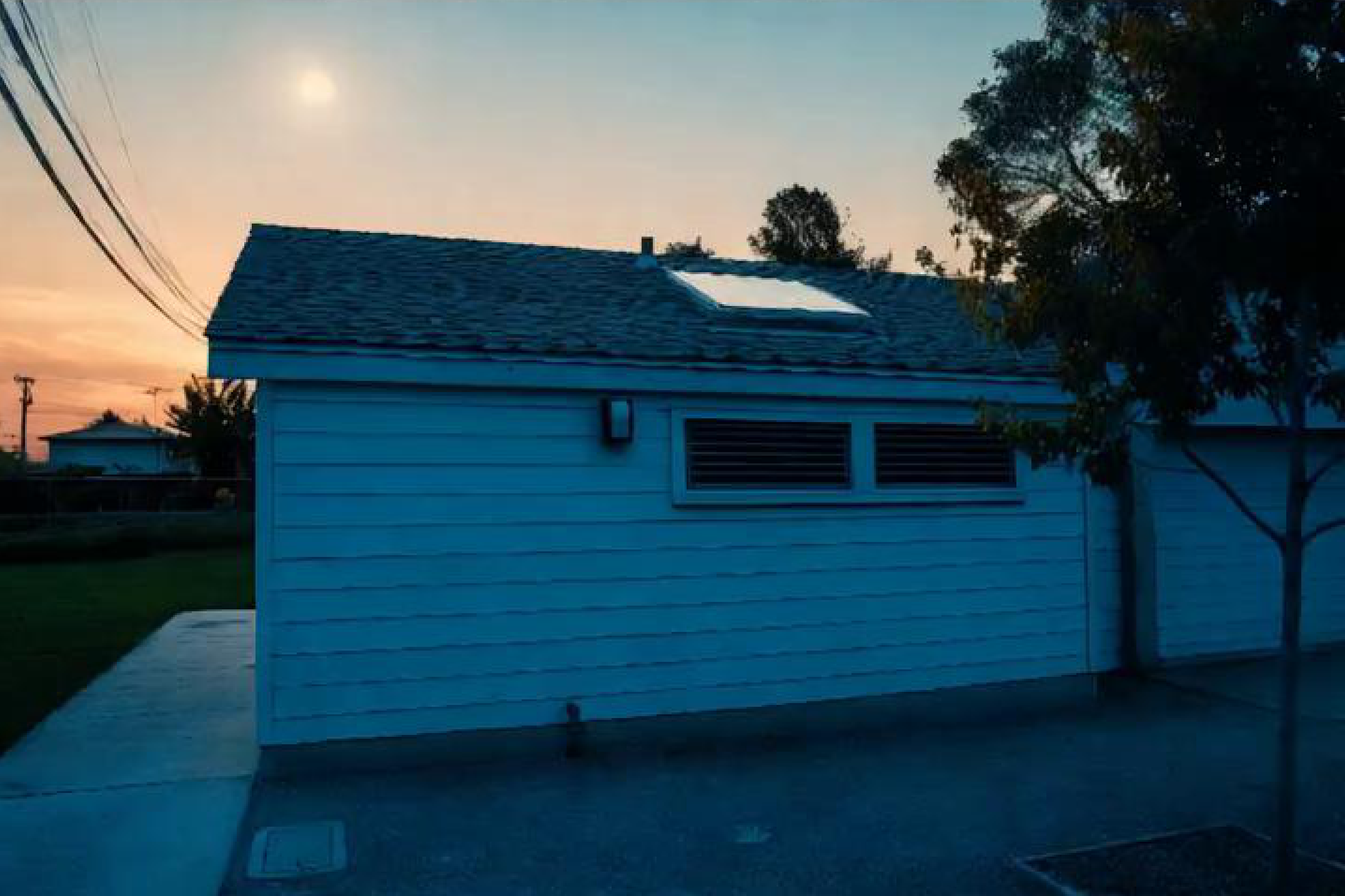} &
      \includegraphics[width=0.31\linewidth]{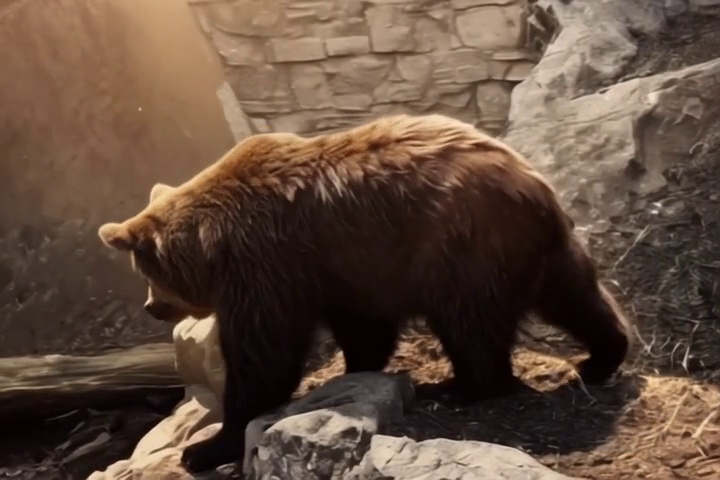} &
      \includegraphics[width=0.31\linewidth]{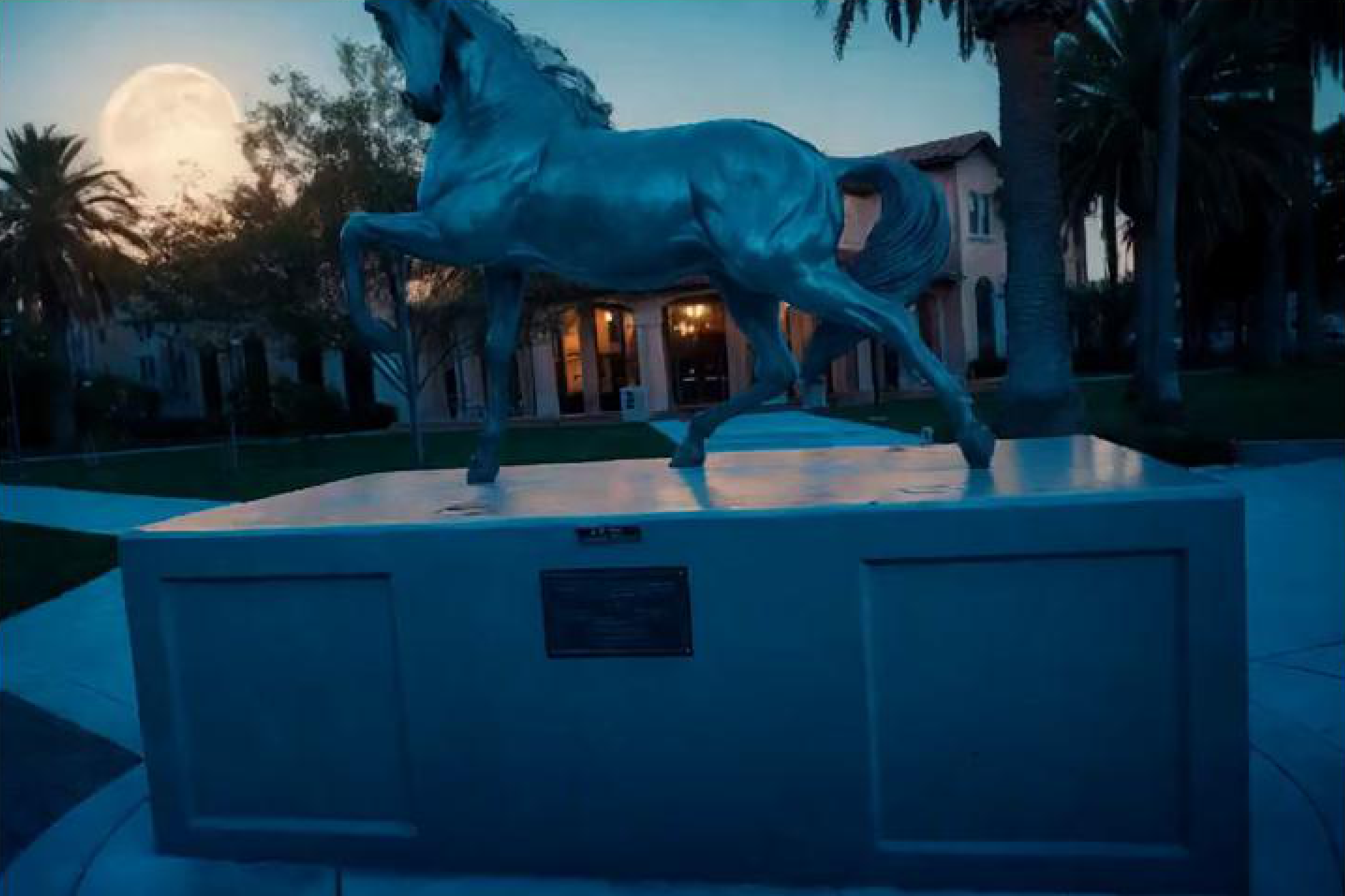}\\
  \end{tabular}
  \caption{We show intermediate frames for a relighting task compared to Light-a-Video \cite{zhou2025lightavideo} that applies the same transfer function (IC-Light) to all (key)frames as guidance, while we just apply it to the first frame. We obtain similar results qualitatively and in particular observe high similarity when evaluating on the Light-a-Video dataset, this is because both methods share the same frozen backbone.}
  \label{fig:qualitative_more_results_simple_3}
  
\end{figure}

We evaluate representation-level conditioning for long-horizon transfer on the out-of-domain Tanks and Temples benchmark. Quantitative results in \autoref{tab:benchmark_tt} and \autoref{tab:i0t} show that DINO-based conditioning provides strong and stable guidance under large camera motion, with improved temporal and structural consistency across both settings. As also illustrated in \autoref{fig:qualitative_more_results_simple_3}, our method has lower generative flexibility than weaker controls, but this is compensated by substantially higher output quality and consistency. Camera control is stronger because the model can offload part of the generative burden to the base diffusion prior, allowing the diffusion pathway to focus on finer structural details (\autoref{fig:style_aug_ablation}).

\begin{figure*}[htbp]
  \centering
  \setlength{\tabcolsep}{1pt}
  \renewcommand{\arraystretch}{1.0}
  \scriptsize
  \begin{tabular}{@{}c c c c c c c c@{}}
    & \tiny{$t{=}2$} & \tiny{$t{=}T/2$} & \tiny{$t{=}T{-}1$} & & \tiny{$t{=}2$} & \tiny{$t{=}T/2$} & \tiny{$t{=}T{-}1$}\\
    \raisebox{4ex}{\rotatebox{90}{\tiny GT}} &
      \includegraphics[width=0.155\linewidth]{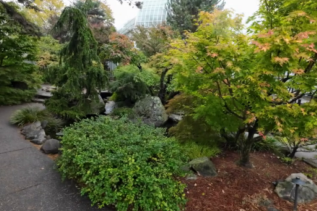} &
      \includegraphics[width=0.155\linewidth]{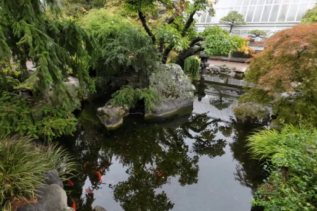} &
      \includegraphics[width=0.155\linewidth]{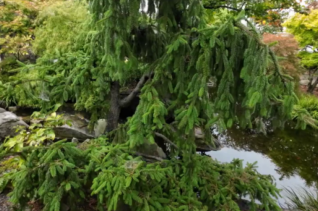} & &
      \includegraphics[width=0.155\linewidth]{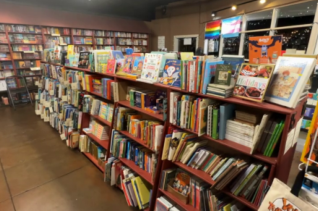} &
      \includegraphics[width=0.155\linewidth]{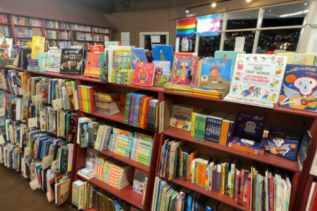} &
      \includegraphics[width=0.155\linewidth]{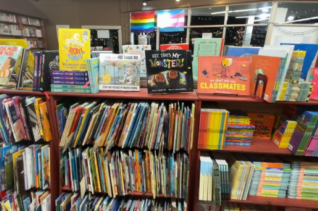}\\
    \raisebox{1.5ex}{\rotatebox{90}{\tiny C-DINO}} &
      \includegraphics[width=0.155\linewidth]{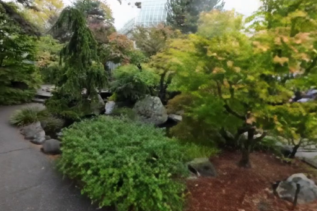} &
      \includegraphics[width=0.155\linewidth]{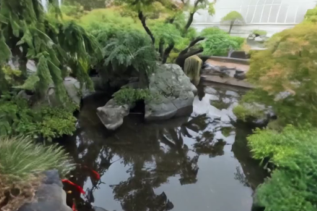} &
      \includegraphics[width=0.155\linewidth]{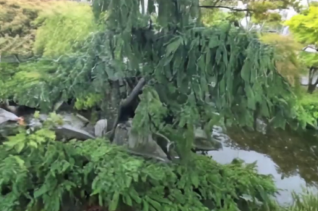} & &
      \includegraphics[width=0.155\linewidth]{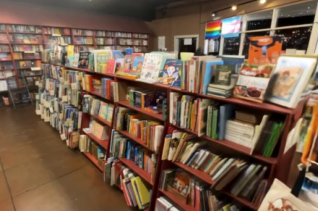} &
      \includegraphics[width=0.155\linewidth]{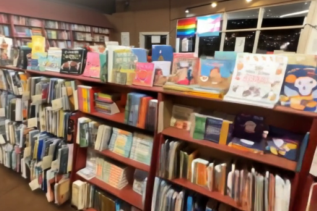} &
      \includegraphics[width=0.155\linewidth]{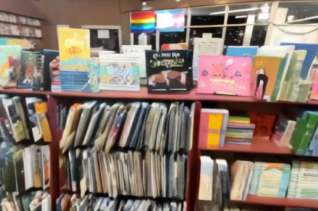}\\
    \raisebox{2.5ex}{\rotatebox{90}{\tiny Canny}} &
      \includegraphics[width=0.155\linewidth]{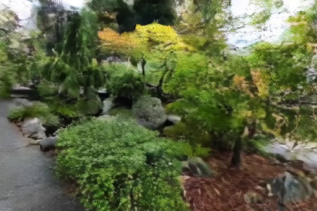} &
      \includegraphics[width=0.155\linewidth]{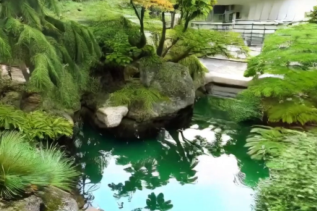} &
      \includegraphics[width=0.155\linewidth]{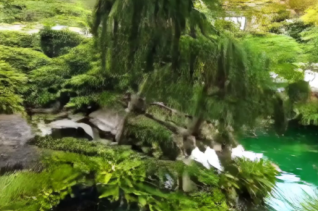} & &
      \includegraphics[width=0.155\linewidth]{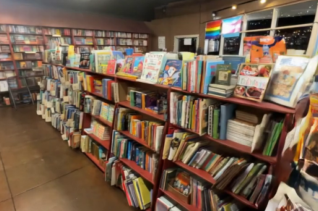} &
      \includegraphics[width=0.155\linewidth]{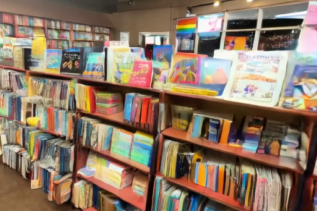} &
      \includegraphics[width=0.155\linewidth]{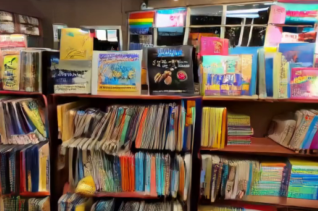}\\
    \raisebox{2.5ex}{\rotatebox{90}{\tiny Depth}} &
      \includegraphics[width=0.155\linewidth]{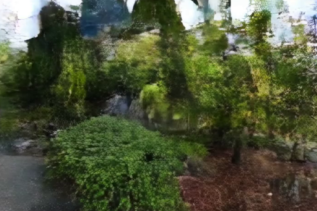} &
      \includegraphics[width=0.155\linewidth]{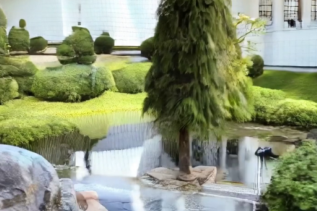} &
      \includegraphics[width=0.155\linewidth]{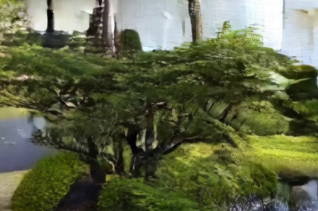} & &
      \includegraphics[width=0.155\linewidth]{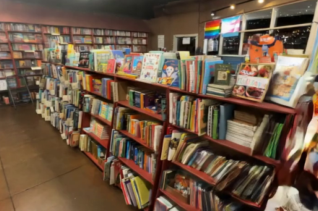} &
      \includegraphics[width=0.155\linewidth]{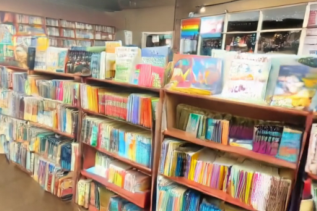} &
      \includegraphics[width=0.155\linewidth]{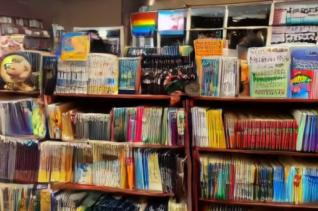}\\
  \end{tabular}
  \caption{We evaluate different methods on a reconstruction task on the DL3DV dataset, featuring high-frequency geometry such as foliage and books. We have found that our method tends to replicate the input more accurately, despite color and lighting shifts caused by the augmentation.}
  \label{fig:qualitative_2scenes}
\end{figure*}

\begin{figure}[t]
  \centering
  \setlength{\tabcolsep}{1pt}
  \renewcommand{\arraystretch}{1.0}
  \scriptsize
  \begin{tabular}{@{}c c c c @{\hspace{8pt}} c c c c@{}}
    & \tiny{$t{=}0$} & \tiny{$t{=}T/2$} & \tiny{$t{=}T{-}1$} &
    & \tiny{$t{=}0$} & \tiny{$t{=}T/2$} & \tiny{$t{=}T{-}1$}\\
    \raisebox{2ex}{\rotatebox{90}{\shortstack{\tiny Francis\\\tiny Drawing}}} &
      \includegraphics[width=0.15\linewidth]{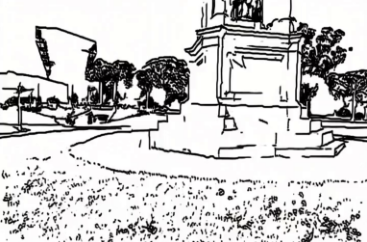} &
      \includegraphics[width=0.15\linewidth]{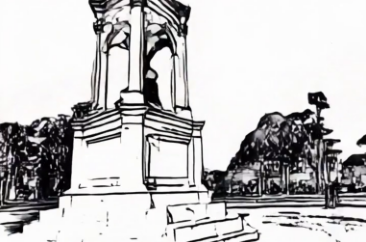} &
      \includegraphics[width=0.15\linewidth]{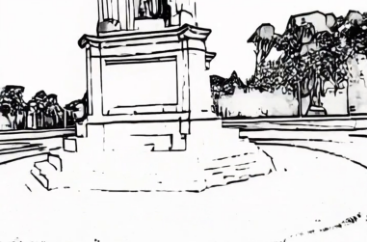} &
    \raisebox{2ex}{\rotatebox{90}{\shortstack{\tiny Truck\\\tiny Mosaic}}} &
      \includegraphics[width=0.15\linewidth]{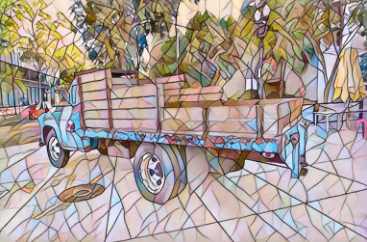} &
      \includegraphics[width=0.15\linewidth]{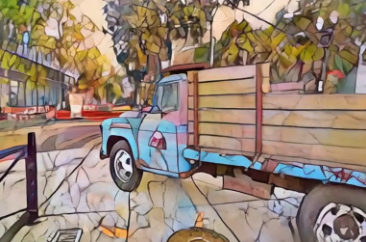} &
      \includegraphics[width=0.15\linewidth]{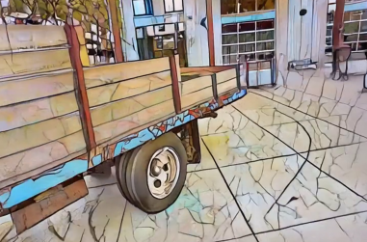}\\
    \raisebox{2.5ex}{\rotatebox{90}{\shortstack{\tiny Horse\\\tiny Studio}}} &
      \includegraphics[width=0.15\linewidth]{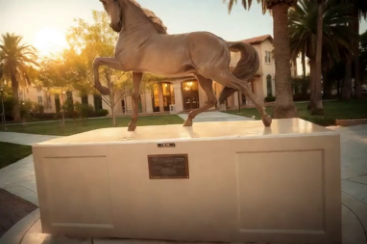} &
      \includegraphics[width=0.15\linewidth]{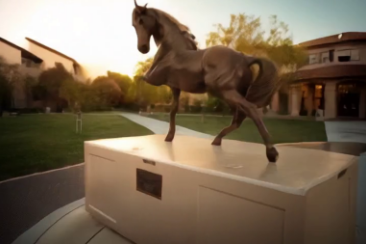} &
      \includegraphics[width=0.15\linewidth]{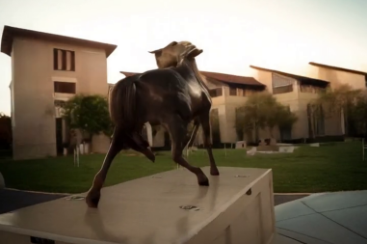} &
    \raisebox{1.5ex}{\rotatebox{90}{\shortstack{\tiny Temple\\\tiny Neon}}} &
      \includegraphics[width=0.15\linewidth]{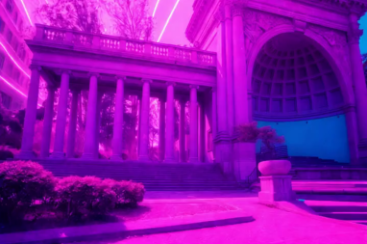} &
      \includegraphics[width=0.15\linewidth]{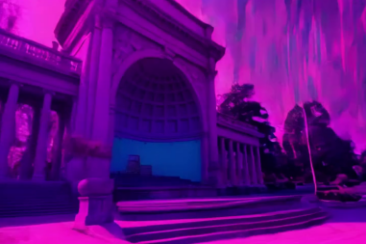} &
      \includegraphics[width=0.15\linewidth]{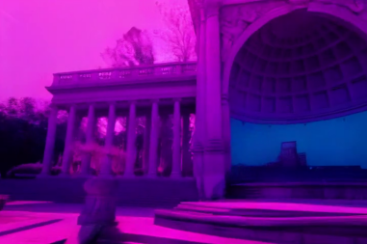}\\
    \raisebox{1.5ex}{\rotatebox{90}{\shortstack{\tiny Court\\\tiny Mondrian}}} &
      \includegraphics[width=0.15\linewidth]{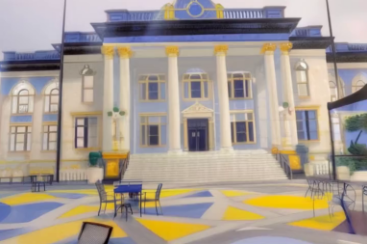} &
      \includegraphics[width=0.15\linewidth]{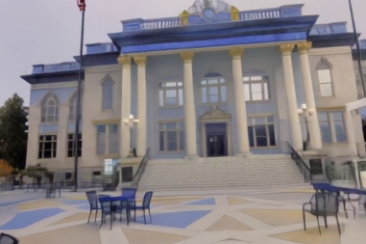} &
      \includegraphics[width=0.15\linewidth]{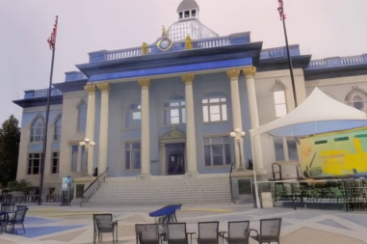} &
    \raisebox{2.5ex}{\rotatebox{90}{\shortstack{\tiny Train\\\tiny Arcane}}} &
      \includegraphics[width=0.15\linewidth]{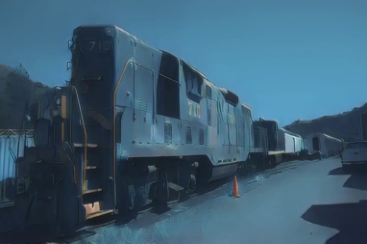} &
      \includegraphics[width=0.15\linewidth]{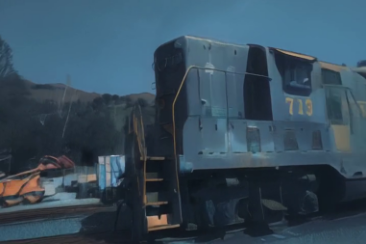} &
      \includegraphics[width=0.15\linewidth]{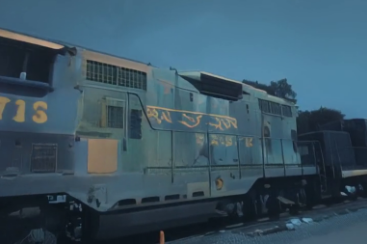}\\
  \end{tabular}
  \caption{Additional 2D transfer results on T\&T scenes with different validation styles.}
  \label{fig:qualitative_more_results_simple}
\end{figure}

Our appearance-decoupled training makes this representation practical for transfer: the same conditioning interface supports RGB-to-stylized transfer and geometry-driven transfer (e.g., mesh-to-real), as discussed in Sec.~\ref{subsec:rendered_features}. Qualitative examples are shown in \autoref{fig:qualitative_2scenes} and \autoref{fig:qualitative_more_results_simple_3}, with additional results in \autoref{fig:qualitative_more_results_simple}. We observe particularly strong relighting behavior, consistent with the built-in invariances of DINO features and the pixel-aligned geometric constraints of the task.

\begin{figure}[htbp]
  \centering
  \setlength{\tabcolsep}{1pt}
  \renewcommand{\arraystretch}{1.0}
  \scriptsize 
  \textbf{ScanNet++ Scene} (09c1414f1b)\\[2pt]
  \begin{tabular}{@{}c c c c c c c @{}}
    & \tiny{GT} & \tiny{2D} & \tiny{Voxel} & \tiny{Mesh} & \tiny{Concerto}\ & \shortstack{\tiny{Concerto}\\\tiny{inpainted}}\\
  \raisebox{3ex}{\rotatebox{90}{\tiny $t{=}35$}} &
      \includegraphics[width=0.16\linewidth]{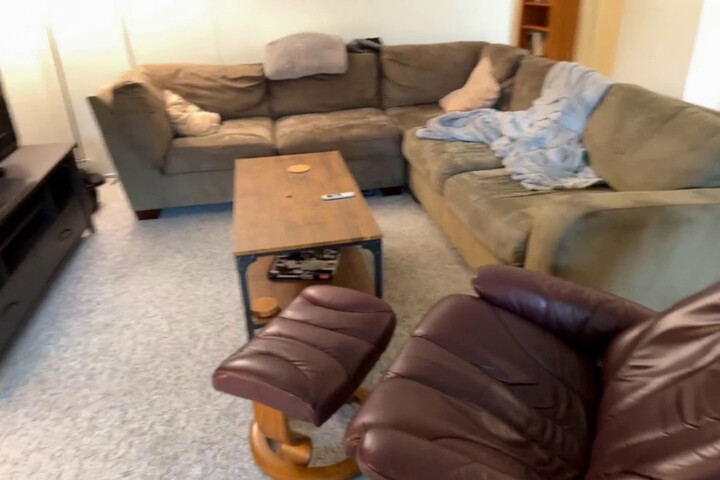} &
    \includegraphics[width=0.16\linewidth]{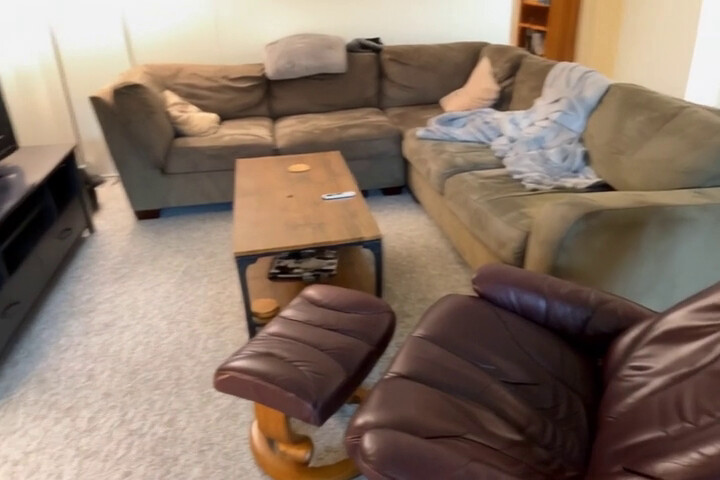} &
    \includegraphics[width=0.16\linewidth]{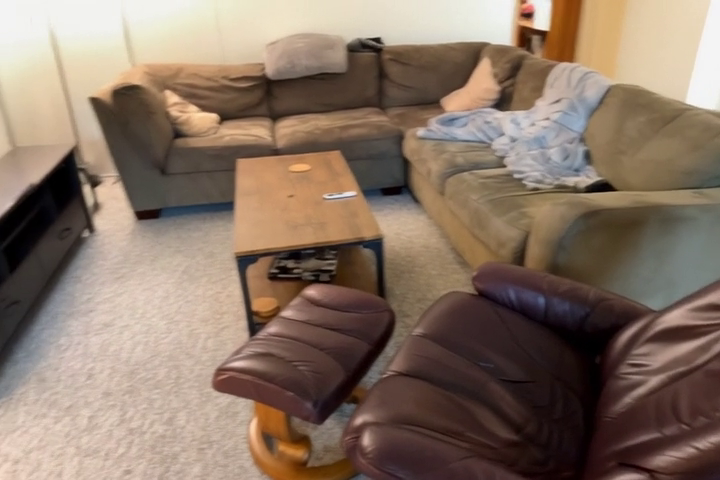} &
    \includegraphics[width=0.16\linewidth]{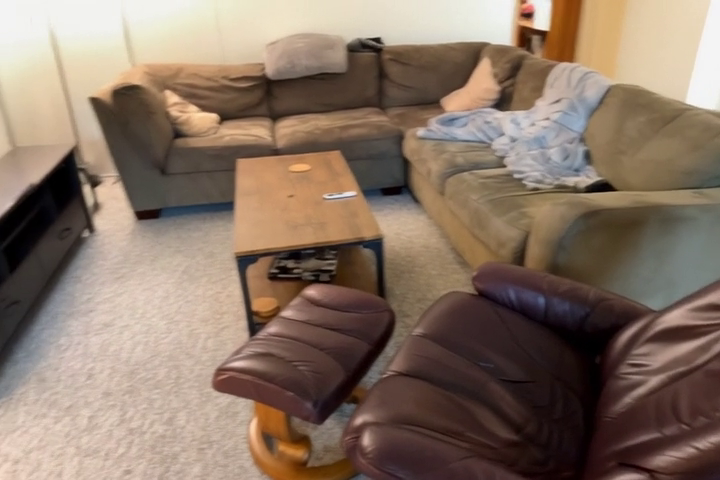} &
    \includegraphics[width=0.16\linewidth]{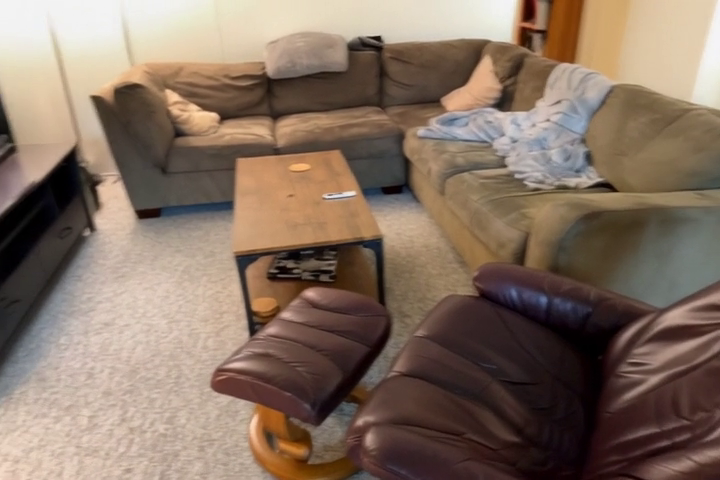} & 
    \includegraphics[width=0.16\linewidth]{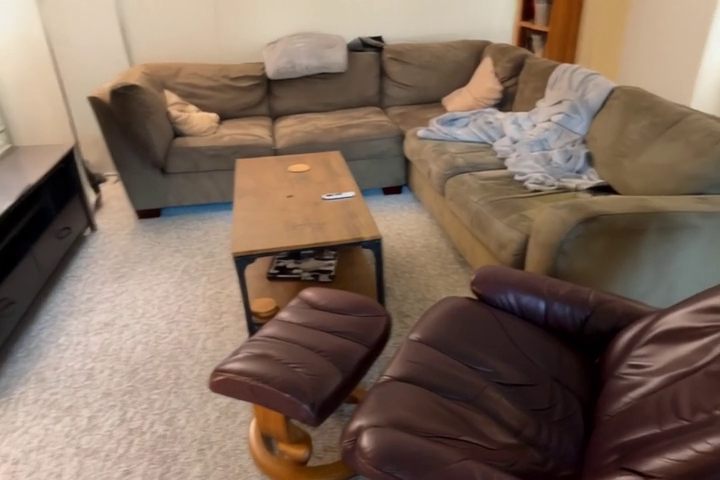}\\
    
    \raisebox{4ex}{\rotatebox{90}{\tiny $t{=}49$}} &
    \includegraphics[width=0.16\linewidth]{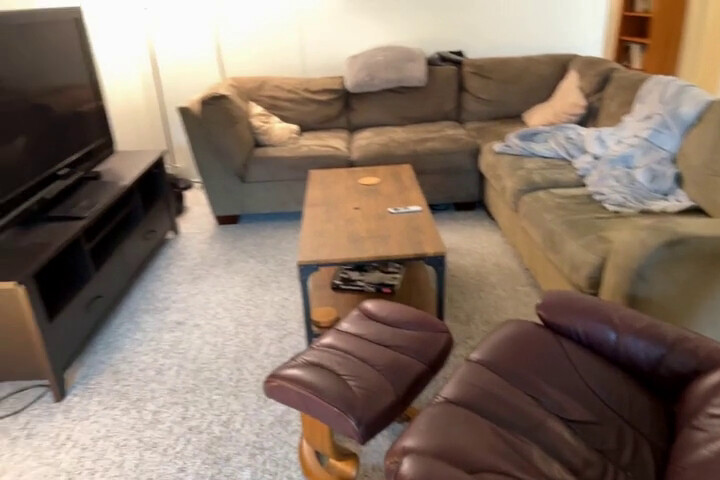} &
    \includegraphics[width=0.16\linewidth]{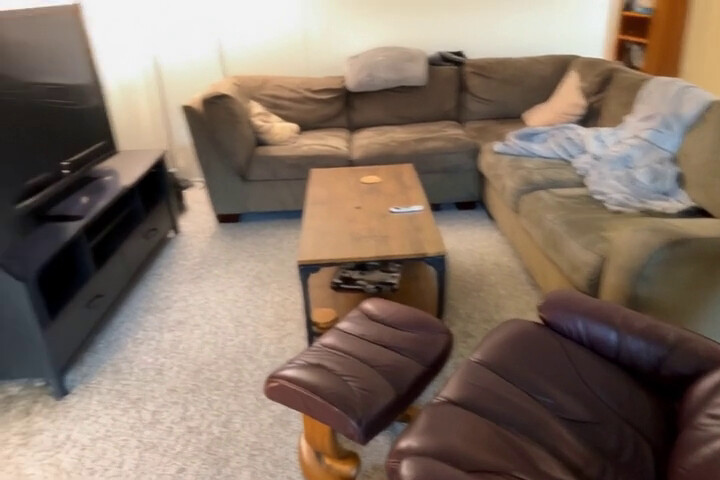} &
    \includegraphics[width=0.16\linewidth]{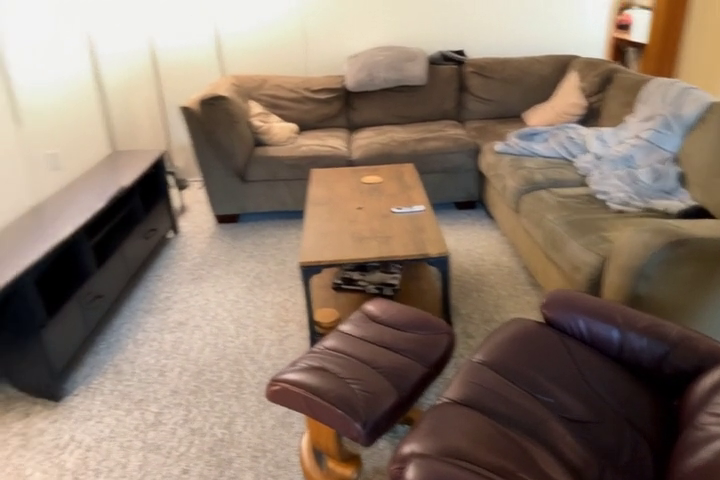} &
        \includegraphics[width=0.16\linewidth]{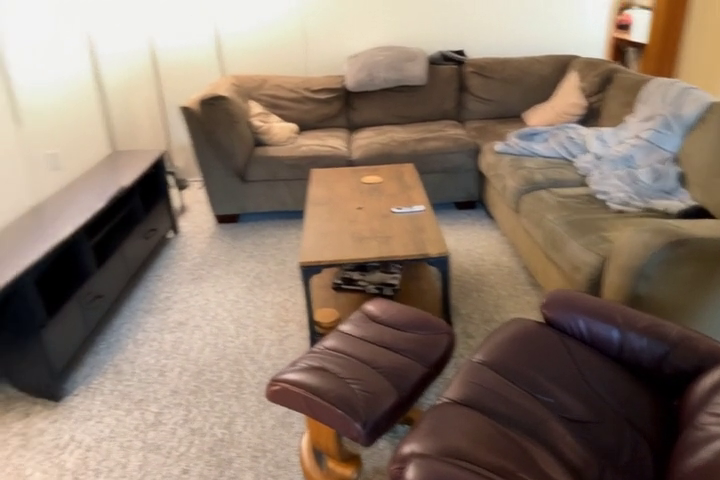}  &
    \includegraphics[width=0.16\linewidth]{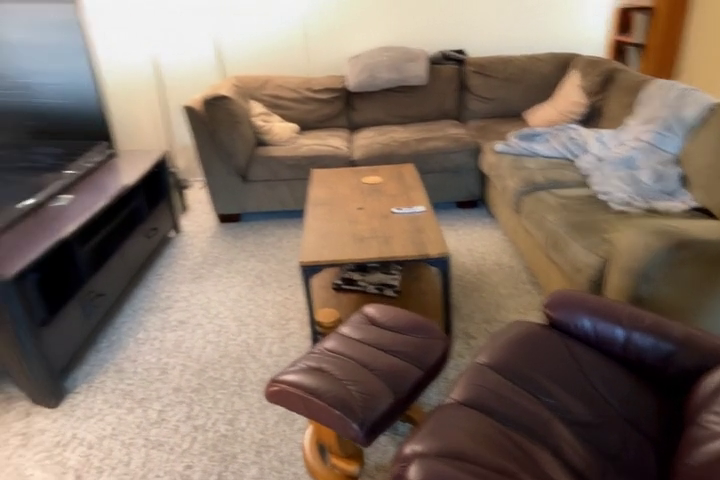} &
 \includegraphics[width=0.16\linewidth]{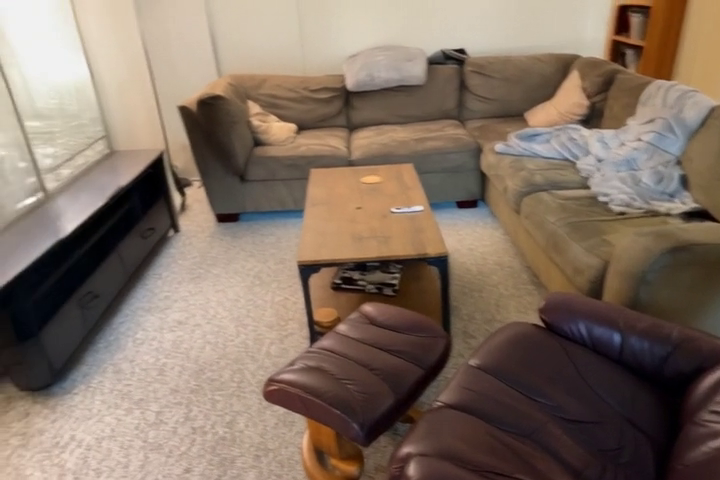}
\\
    
    \raisebox{5ex}{\rotatebox{90}{\tiny $t{=}0$}} &
    \includegraphics[width=0.16\linewidth]{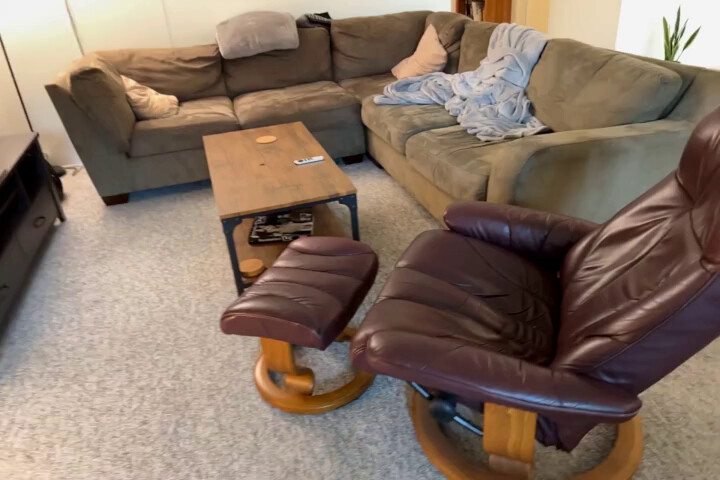} &
    \includegraphics[width=0.16\linewidth]{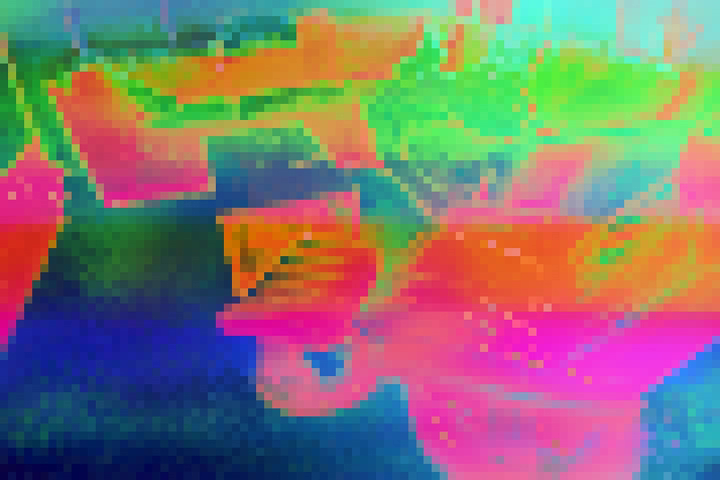} &
    \includegraphics[width=0.16\linewidth]{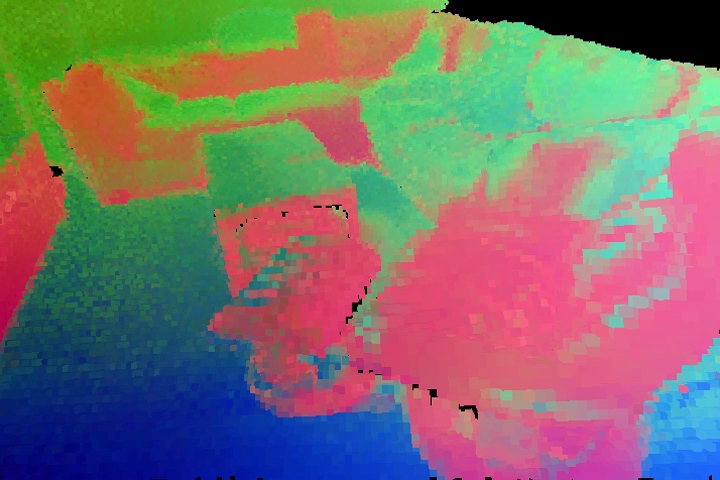} &
     \includegraphics[width=0.16\linewidth]{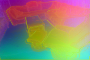} &
    \includegraphics[width=0.16\linewidth]{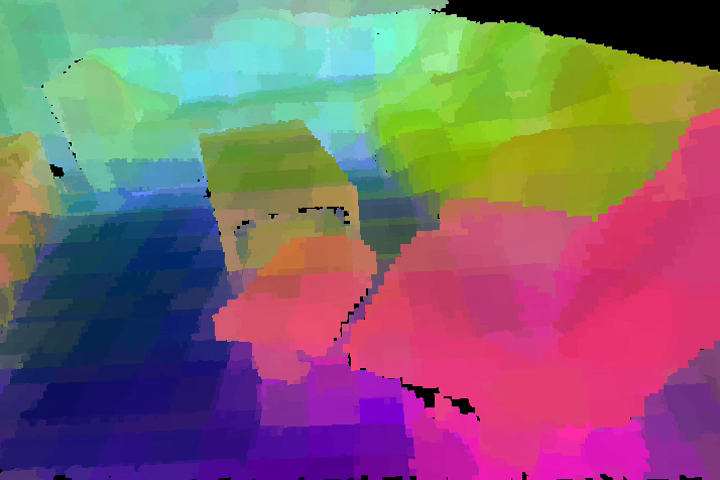} &
     \includegraphics[width=0.16\linewidth]{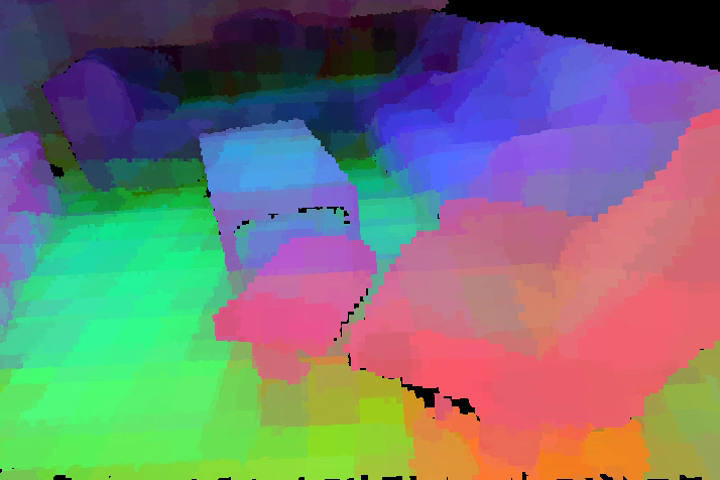}
 \\
  \end{tabular}

  \vspace{4pt}
   \textbf{ScanNet++ Scene} (0d2ee665be)\\[2pt]
  \begin{tabular}{@{}c c c c c c c @{}}
    & \tiny{GT} & \tiny{2D} & \tiny{Voxel} & \tiny{Mesh} & \tiny{Concerto} & \shortstack{\tiny{Concerto}\\\tiny{inpainted}}\\
    \raisebox{3ex}{\rotatebox{90}{\tiny $t{=}35$}} &
    \includegraphics[width=0.16\linewidth]{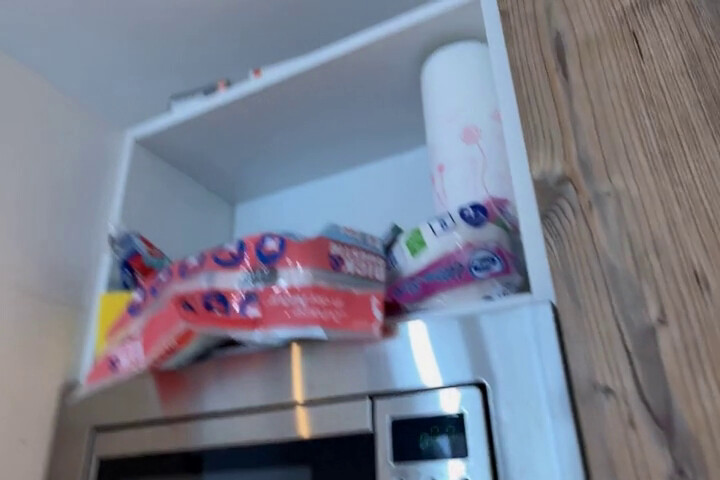} &
    \includegraphics[width=0.16\linewidth]{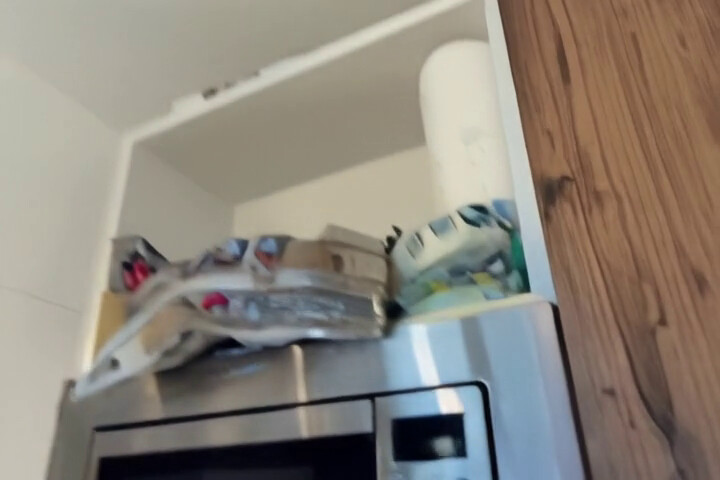} &
    \includegraphics[width=0.16\linewidth]{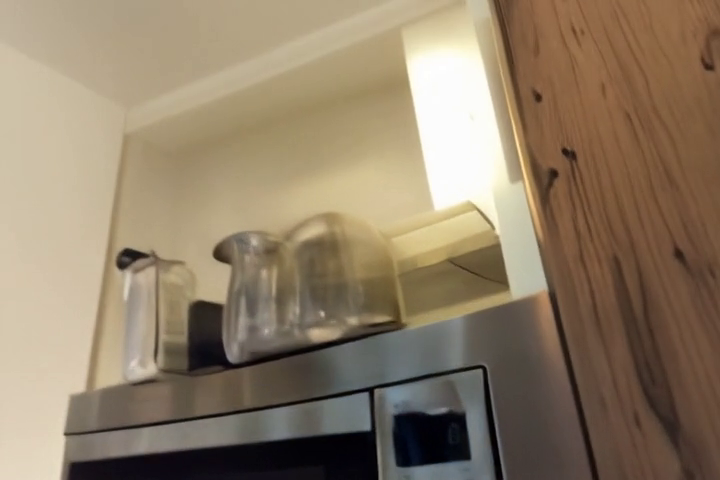} &
    \includegraphics[width=0.16\linewidth]{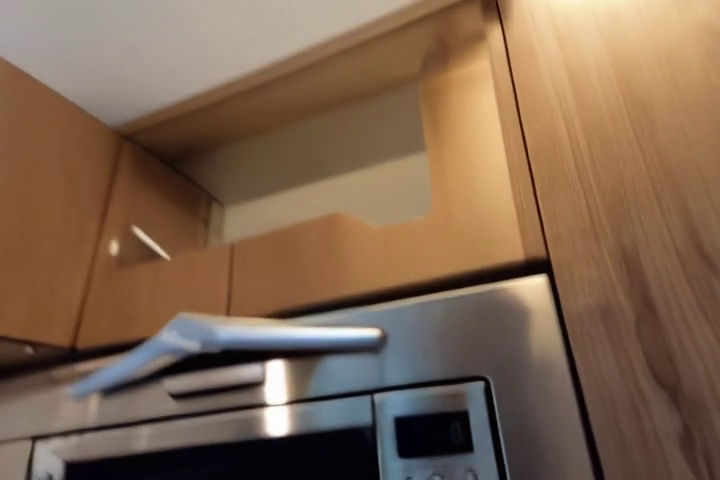} &
    \includegraphics[width=0.16\linewidth]{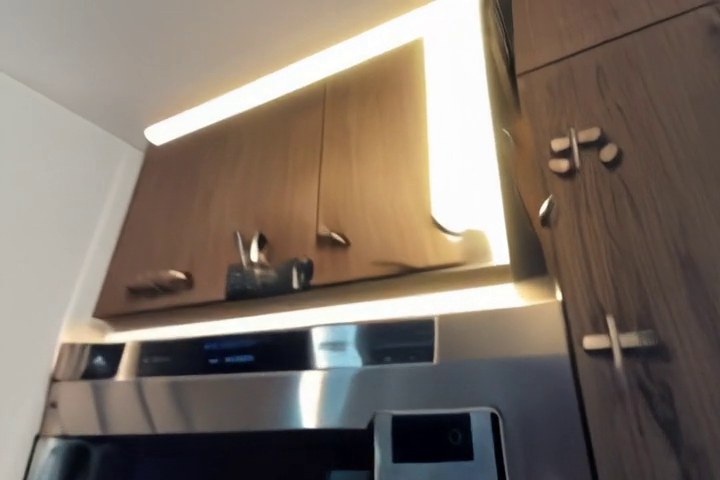} &
    \includegraphics[width=0.16\linewidth]{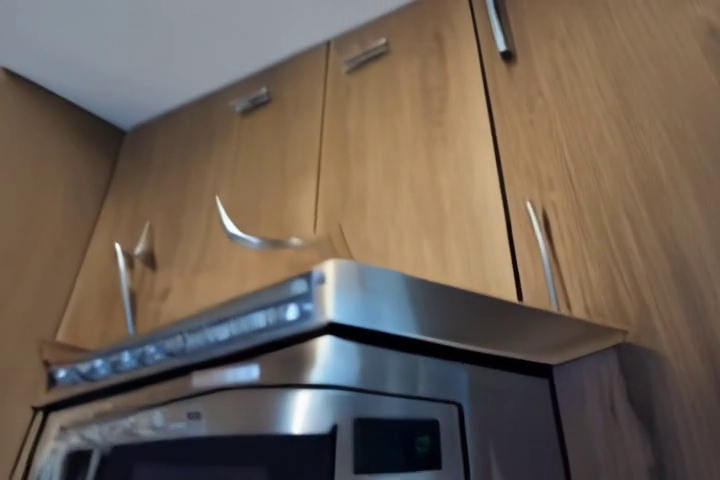} \\
    
    \raisebox{4ex}{\rotatebox{90}{\tiny $t{=}49$}} &
    \includegraphics[width=0.16\linewidth]{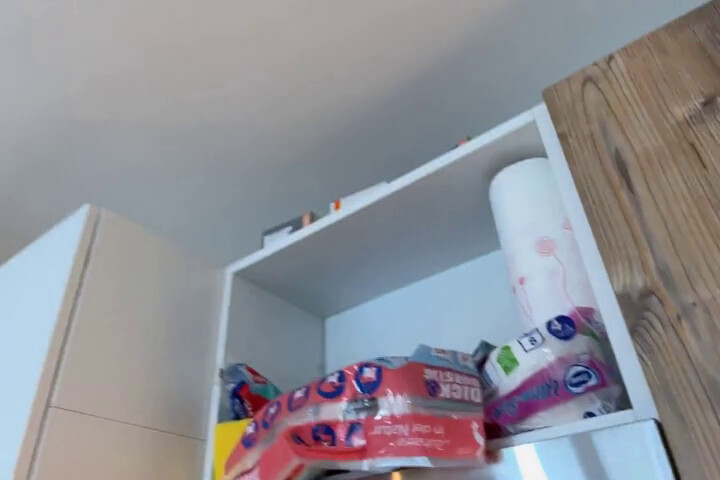} &
    \includegraphics[width=0.16\linewidth]{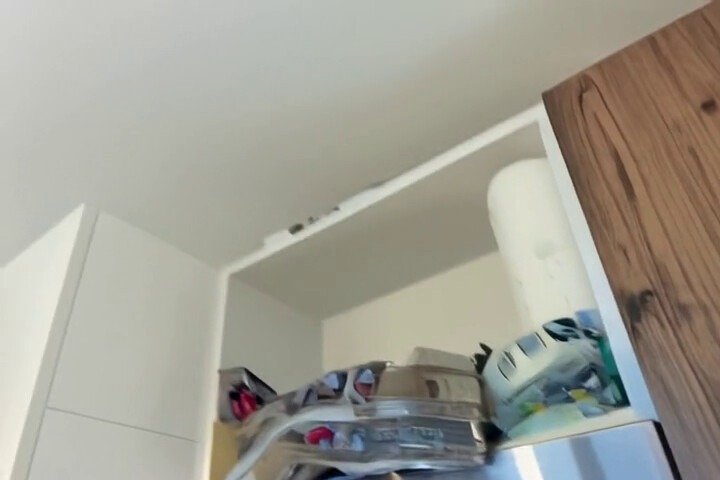} &
    \includegraphics[width=0.16\linewidth]{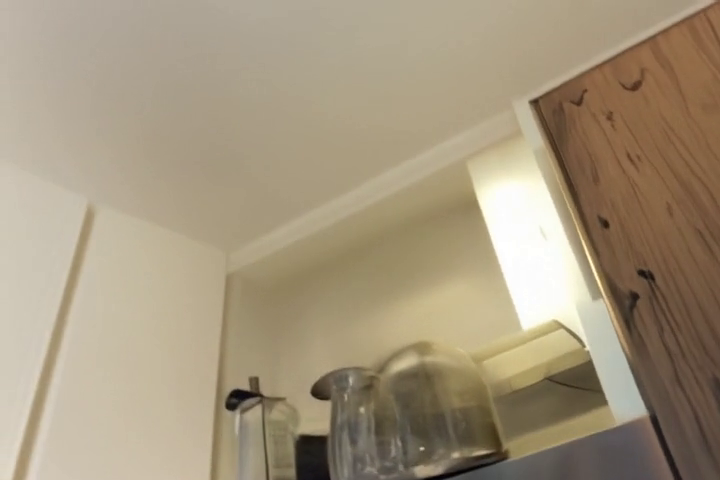} &
    \includegraphics[width=0.16\linewidth]{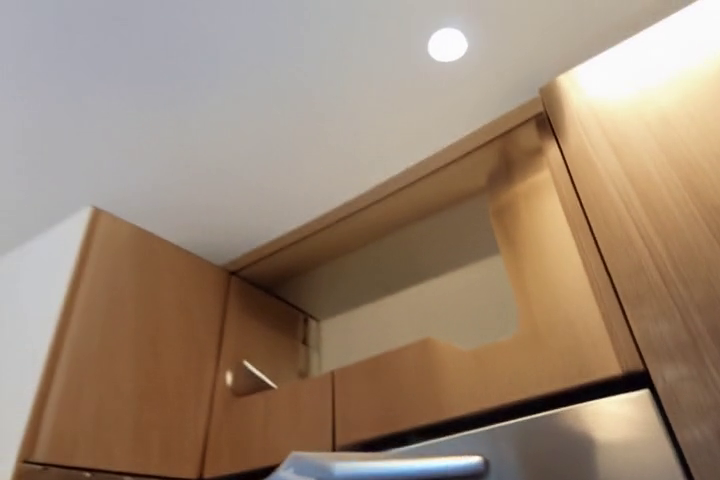}  &
    \includegraphics[width=0.16\linewidth]{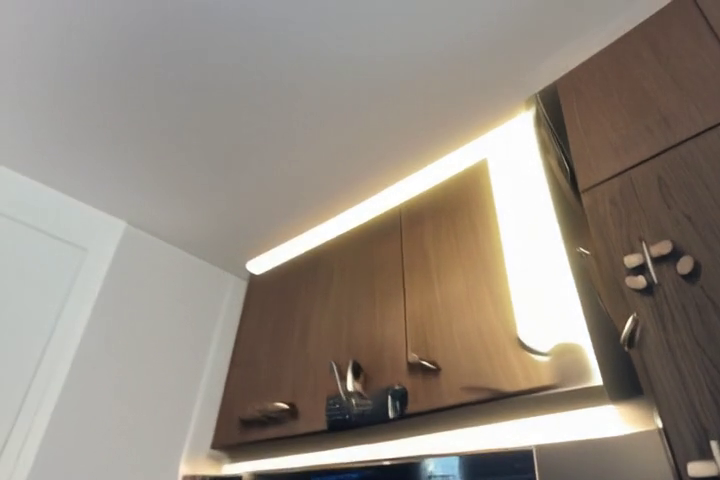} &
    \includegraphics[width=0.16\linewidth]{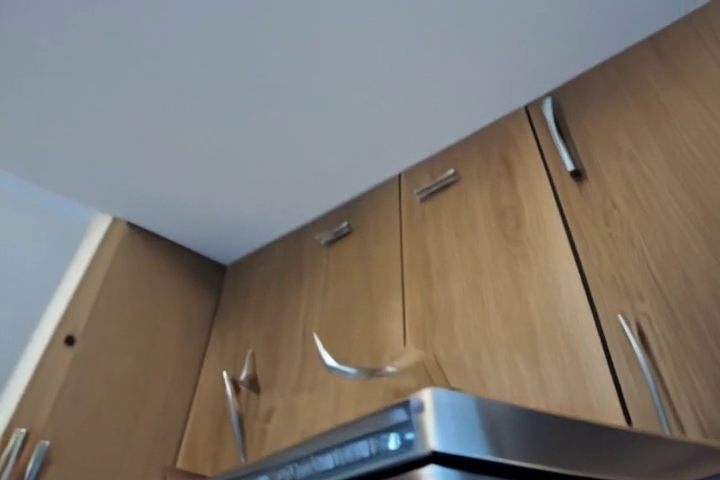}\\
    
    \raisebox{5ex}{\rotatebox{90}{\tiny $t{=}0$}} &
    \includegraphics[width=0.16\linewidth]{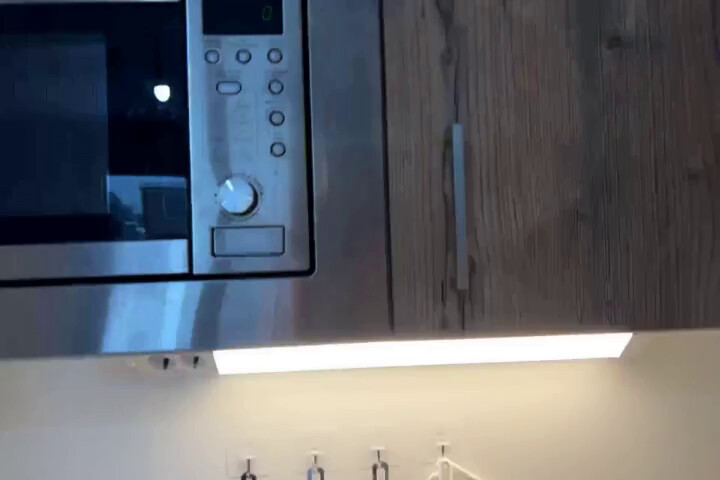} &
    \includegraphics[width=0.16\linewidth]{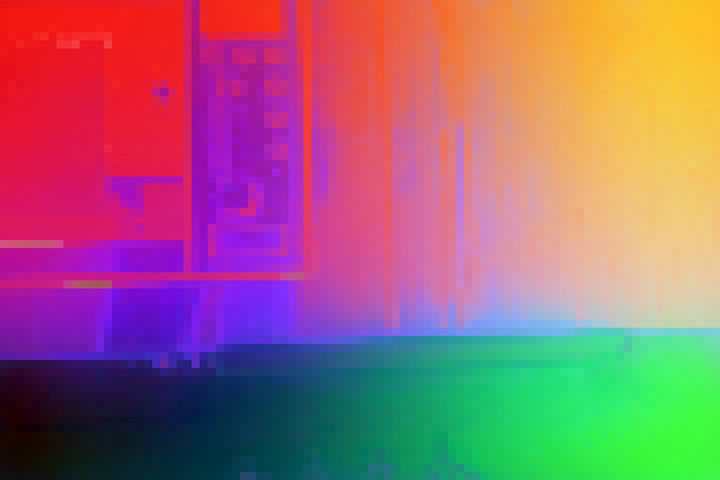} &
    \includegraphics[width=0.16\linewidth]{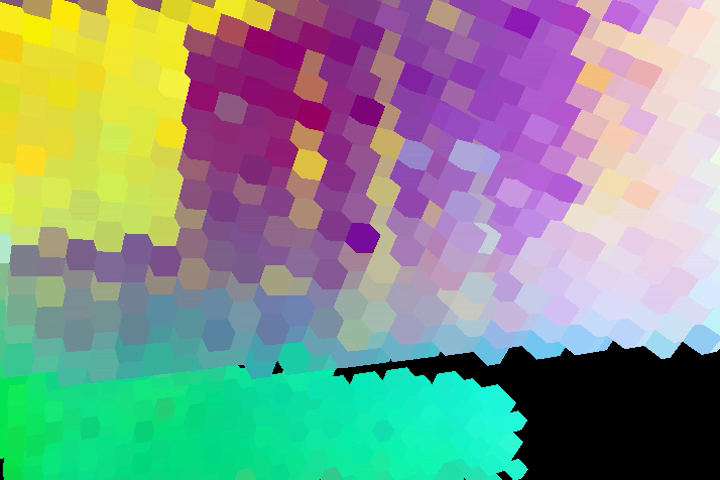} &
    \includegraphics[width=0.16\linewidth]{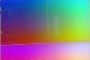} &
    \includegraphics[width=0.16\linewidth]{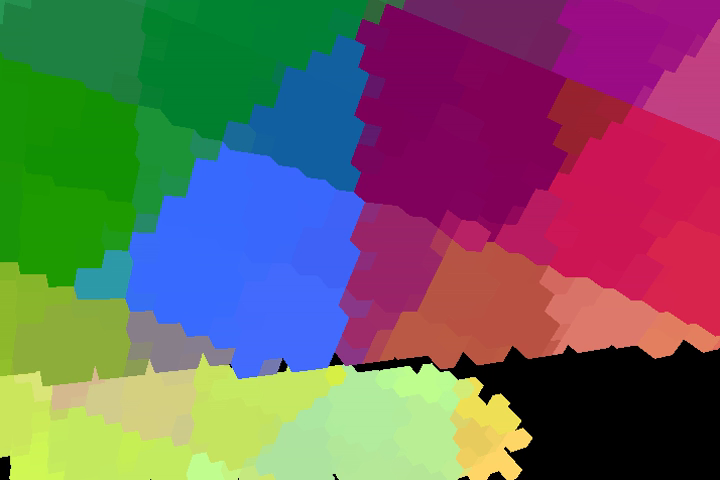} &
    \includegraphics[width=0.16\linewidth]{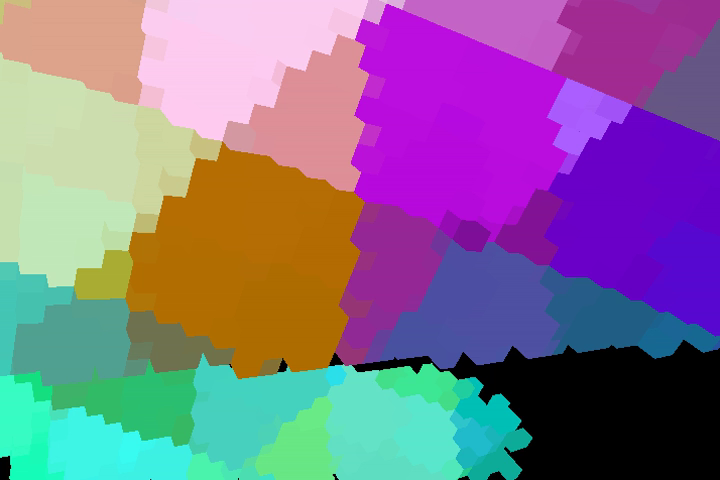}
     \\
  \end{tabular}

  \caption{Qualitative comparison of Control-DINO conditioning methods at matched time steps for ScanNet++. We show generated outputs at $t{=}35$ and $t{=}49$ for \textit{2D}, \textit{Voxel}, \textit{Concerto}, and \textit{Mesh} conditioning. In the bottom row $t{=}0$, we show the conditioning DINO feature signal for each method, while the GT frame at $t{=}0$ is the given starting frame.}
  \label{fig:conditioning_signal_comparison}
\end{figure}

\begin{table}[t]
\centering
\begin{minipage}[t]{0.49\linewidth}
\centering
\caption{We measure feature disentanglement of different projection strategies, measuring cosine similarity as agreement between real and stylized scene features after projection, and retained feature variance. The metrics are computed from samples of the training dataset. }
   \label{tab:pca_basis}
\textbf{Projection basis disentanglement}\\[2pt]
\resizebox{\linewidth}{!}{
\begin{tabular}{lcc}
\toprule
Method & Cosine Similarity\ $\uparrow$ & Explained Variance (\%) $\uparrow$ \\
\midrule
No projection (384-d) & 0.9194 & 100.0 \\
\midrule
Sim2Real reference (DwD~\cite{chen2026dwd}) & 0.9598 & -- \\
\midrule
Standard PCA & 0.9110 & 72.5 \\
Style-Invariant PCA & 0.9481 & 30.4 \\
Bottom Eigenvectors & 0.9929 & 2.9 \\
Random Orthogonal & 0.9226 & 16.7 \\
\bottomrule
\end{tabular}
}
\end{minipage}\hfill
\begin{minipage}[t]{0.49\linewidth}
\centering
\caption{We ablate our architectural design choices on DL3DV140. We observe that retaining the raw high-dimensional features outperform PCA-projected or tail-dropped variants, and that removing the spatial mixing step substantially degrades quality.}
\label{tab:pca_ablation}
\textbf{Architectural ablation on DL3DV140.}\\[2pt]
\resizebox{\linewidth}{!}{
\begin{tabular}{lccccc}
\toprule
Method & FID $\downarrow$ & Subj. Cons. $\uparrow$ & BG Cons. $\uparrow$ & Aesthetic $\uparrow$ & Imaging $\uparrow$ \\
\midrule
Control-DINO & 71.67 & 0.959 & 0.955 & 0.504 & 0.542 \\
w/o Spatial Mixing & 177.80 & 0.919 & 0.933 & 0.474 & 0.495 \\
w/ Style-Invariant PCA & 89.14 & 0.940 & 0.940 & 0.479 & 0.524 \\
w/ Tail Drop (k=64) & 108.87 & 0.918 & 0.920 & 0.459 & 0.494 \\
w/ Tail Drop (k=8) & 156.67 & 0.845 & 0.908 & 0.468 & 0.490 \\
\bottomrule
\end{tabular}
}
\end{minipage}
\end{table}

\subsubsection{Video-from-3D} We evaluate three 3D-to-video conditioning pipelines that all use a ControlNet branch driven by DINO-like feature maps. For the \textbf{mesh} setting, we render untextured meshes from target camera trajectories, upscale the rendered views, and extract DINOv3-S features as conditioning maps with shape $384\times60\times90$. Because mesh renderings contain holes and missing regions, we additionally encode a binary visibility mask with the CogVideoX VAE and provide it as an auxiliary conditioning signal, so the diffusion model can explicitly inpaint invalid regions instead of overfitting to artifacts.

For the \textbf{voxel} setting, we investigate whether low-resolution geometry with color cues is sufficient. Starting from an input video sequence, we reconstruct geometry from a dense point cloud and associate DINO-encoded image features to 3D points, then voxelize this feature-augmented point cloud. The resulting voxel grid is rendered back to the target views to produce DINO-like conditioning maps; as in the mesh case, we propagate hole masks so the model can ignore unobserved areas during denoising, visualized in \autoref{fig:visualisation_3d_structure}.

Regarding video-from-3D, for \textbf{Concerto-based} conditioning we target the realistic case where only sparse geometry is available and RGB coverage is limited to one input view. We extract Concerto point-cloud descriptors, voxelize them, and render Concerto feature maps as the control signal.
This setup allows generation from sparse, color-free 3D observations while preserving semantic consistency.
Quantitative comparisons are reported in \autoref{tab:video_from_3d}, and qualitative conditioning examples are shown in \autoref{fig:conditioning_signal_comparison}. 

For image-based conditioning, our 2D Control-DINO reference outperforms state-of-the-art depth- and edge-conditioned baselines (notably Wan2.2) by a large margin in reconstruction quality; our 3D-derived variants remain competitive with these baselines while operating in a different application setting. Control-DINO performs best when full color information is available, consistent with the 2D setting where appearance cues are strong.
Variants based on untextured geometry show lower reconstruction quality, as expected: missing appearance reduces feature fidelity and weakens structural disambiguation, and sparse 3D observations leave holes that carry no conditioning signal.

Concerto-based conditioning is reported in two regimes (\autoref{tab:video_from_3d}): with color (\emph{Concerto}) and without color (\emph{Concerto inpainted}, the uncolored point-cloud setting). Despite the stronger input sparsity, the color-free Concerto setup still provides usable appearance priors and enables plausible scene generation, highlighting its practical value for generation-focused use-cases rather than strict reconstruction.

\subsection{Ablations}
\label{subsec:ablations}
We first ablate the effect of training data composition (real-only vs. mixed real+augmented). As shown in \autoref{tab:benchmark_tt}, mixing data can slightly reduce generation quality, while improving transfer fidelity.

We ablate the feature conditioning against two PCA decomposition strategies (\autoref{tab:pca_ablation}): (1) Given paired real and stylized DINOv3 features $f_{\mathrm{real}}$ and $f_{\mathrm{style}}$, we compute difference vectors $d=f_{\mathrm{real}}-f_{\mathrm{style}}$, perform eigendecomposition on their covariance $S=(1/M)D^T D$ to obtain the top-$K$ style directions $V_k$, project them out via $P=I-V_kV_k^T$, then run standard PCA on the cleaned features $Pf_{\mathrm{real}}$ to extract the top-$D$ structure directions as our final projection basis; (2) A tail dropping strategy described in~\cite{Huang2025DIVE} where we first compute a 64-component PCA basis during training, then randomly retain only the first $k$ components with $k\in\{8,16,32,64\}$. At inference, we ablate against $k=8$ and $k=64$.
As can be seen in \autoref{tab:pca_basis}, PCA decomposition, in part because of the already small DINO model we are using, is quite disruptive of the information signal. Even when computing a style invariant decomposition with paired data the amount of information explained remains below $40\%$, which indicates substantial information loss.

We consider the necessity of spatially mixing the signal with the hidden state in a transformer block by replacing it with a simple linear projection directly into the residual (similar to~\cite{Huang2025DIVE}), we have found that the quality of the images remains quite high but the model has difficulty following the conditioning under large camera motion.

\section{Discussion}
We introduced a DINO-based conditioning mechanism for video diffusion.
Dense self-supervised features provide robust structural control while allowing
flexible appearance variation and feature-space resampling at generation time.
The same interface spans 2D transfer (stylization, relighting) and 3D-guided
generation from meshes, voxels, and point clouds.
This is not without limitations, as appearance and semantics are not always perfectly disentangled: the model
occasionally leaks source appearance, particularly for styles far from the
training distribution. Lower guidance mitigates this, but we report all results
at a fixed guidance scale for fairness; \autoref{tab:benchmark} reports the
control to diversity behavior across scales.
More broadly, using high-dimensional features as conditioning entails a trade-off
with generative flexibility: stronger control restricts the diversity of
plausible outputs.
For 3D-derived conditioning, generation quality additionally depends on the
fidelity of the underlying reconstruction and on how well the projected features
approximate the distribution of 2D DINO features the model was trained on.

We believe that learning spatial weighting of the conditioning signal, as well as extending to other foundation feature spaces,
  could further improve the balance between control and generation quality. We release source code and weights for the community to
  build upon.


{\small
  \bibliographystyle{splncs04}
  \bibliography{main}
}

\clearpage
\section{Supplemental}

\subsection{Additional Evaluation Details}
For Tanks and Temples we sample $3$ videos from $10$ scenes for each one of the $3$ times $5$ transfer modes for a total of approximately $20$ thousand unique frames, including some crashes and failures.
For DL3DV140, we sample one video per scene with stride 1 and a random starting point, for a single style and a total of $6.8$ thousand unique frames.

\begin{figure}[t]
    \centering
    \includegraphics[width=\linewidth]{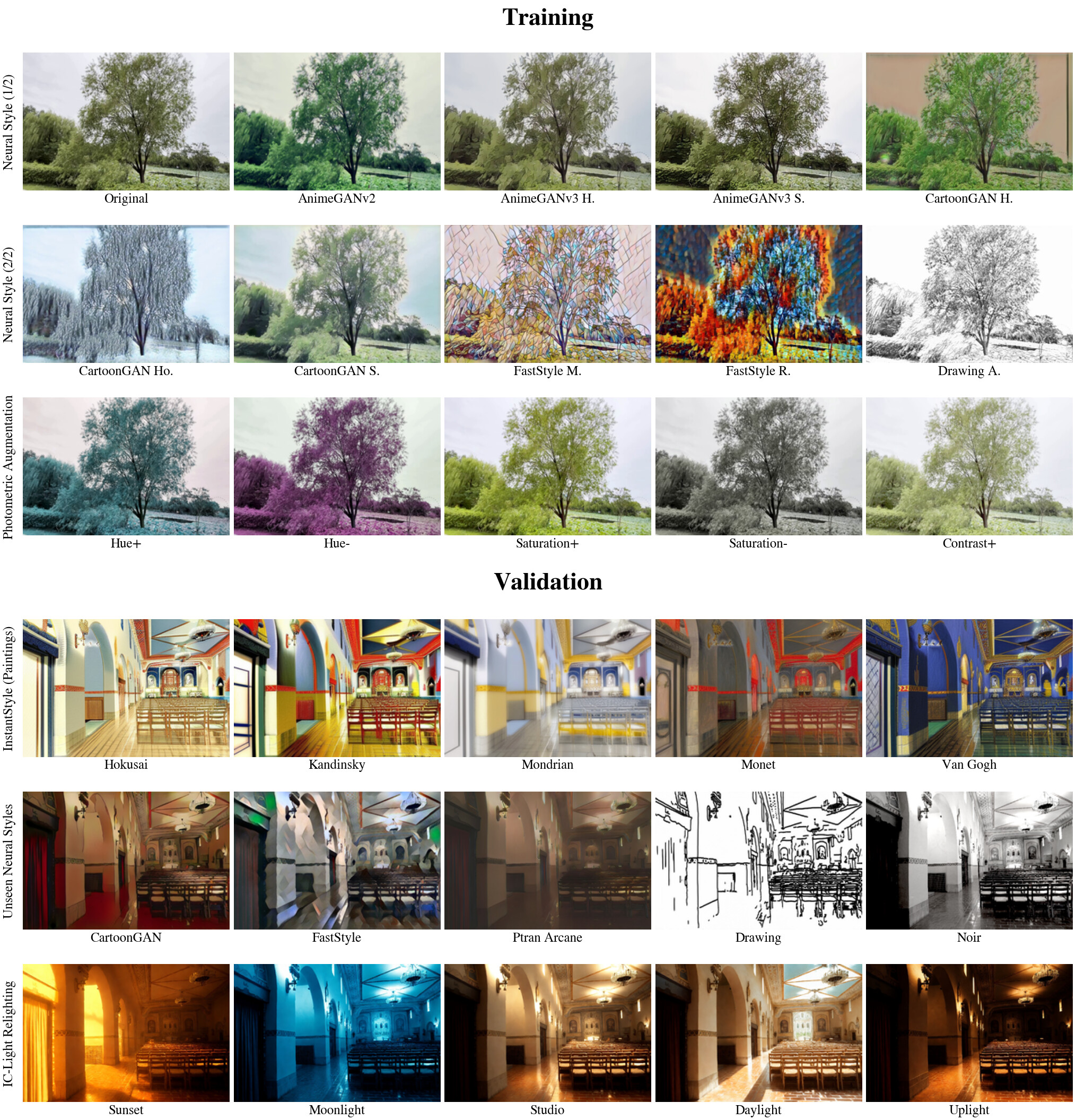}
    \caption{Overview of appearance augmentations used during training and validation. \textbf{Training:} We apply neural style transfer (AnimeGAN, CartoonGAN, FastStyle, Drawing) and photometric augmentations (color temperature, contrast, saturation shifts) to the training videos. \textbf{Validation:} We evaluate generalization using held-out styles including InstantStyle with famous paintings, unseen neural style transfer models, and IC-Light relighting with varying lighting conditions.}
    \label{fig:augmentations}
\end{figure}

The styles used for training and validation are illustrated in \autoref{fig:augmentations}. The model appears to generalize from simple photometric augmentations to more dramatic lighting variations and has learned the neural style transformations effectively. However, under the most extreme InstantStyle transfers, it occasionally struggles to remain within the training data distribution. We hypothesize that incorporating stronger stylistic augmentations during training could improve robustness in such cases.

\begin{figure*}[t]
    \centering
    \includegraphics[width=\textwidth]{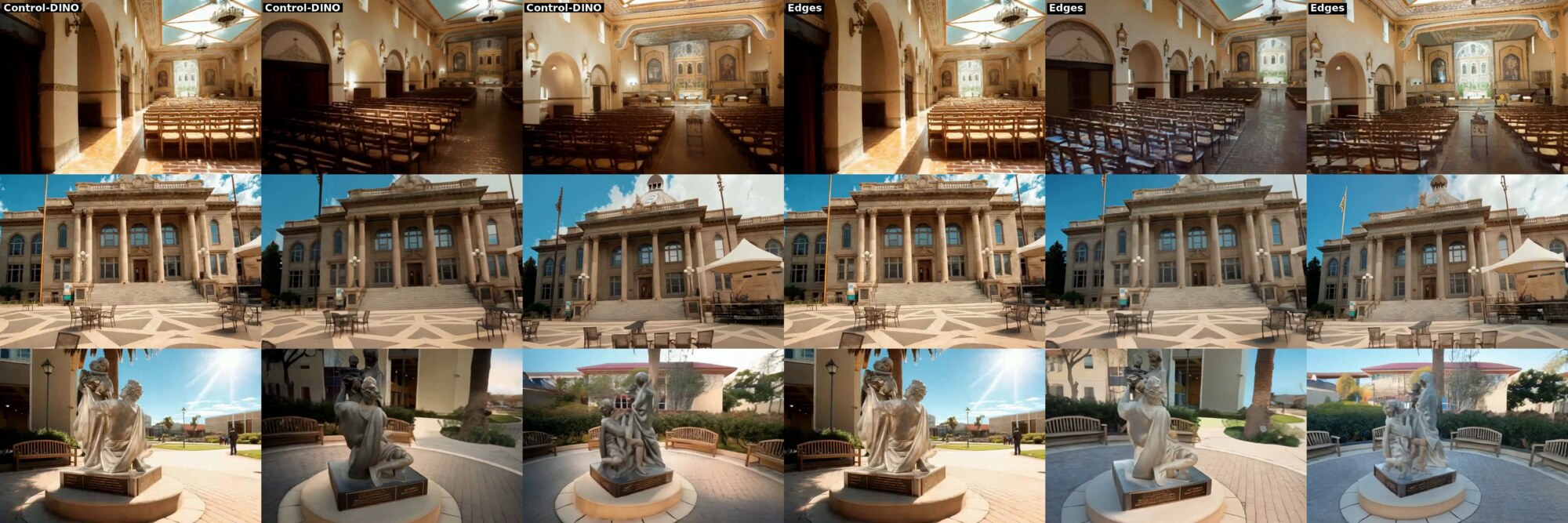}
    \caption{Control-DINO (first 3 columns) and edges (first 3 columns) conditioning results for Tanks and Temples transfer on "Bright Daylight".}
    \label{fig:bright_daylight}
     \vspace{-9ex}
\end{figure*}

\begin{figure*}[t]
    \centering
    \includegraphics[width=\textwidth]{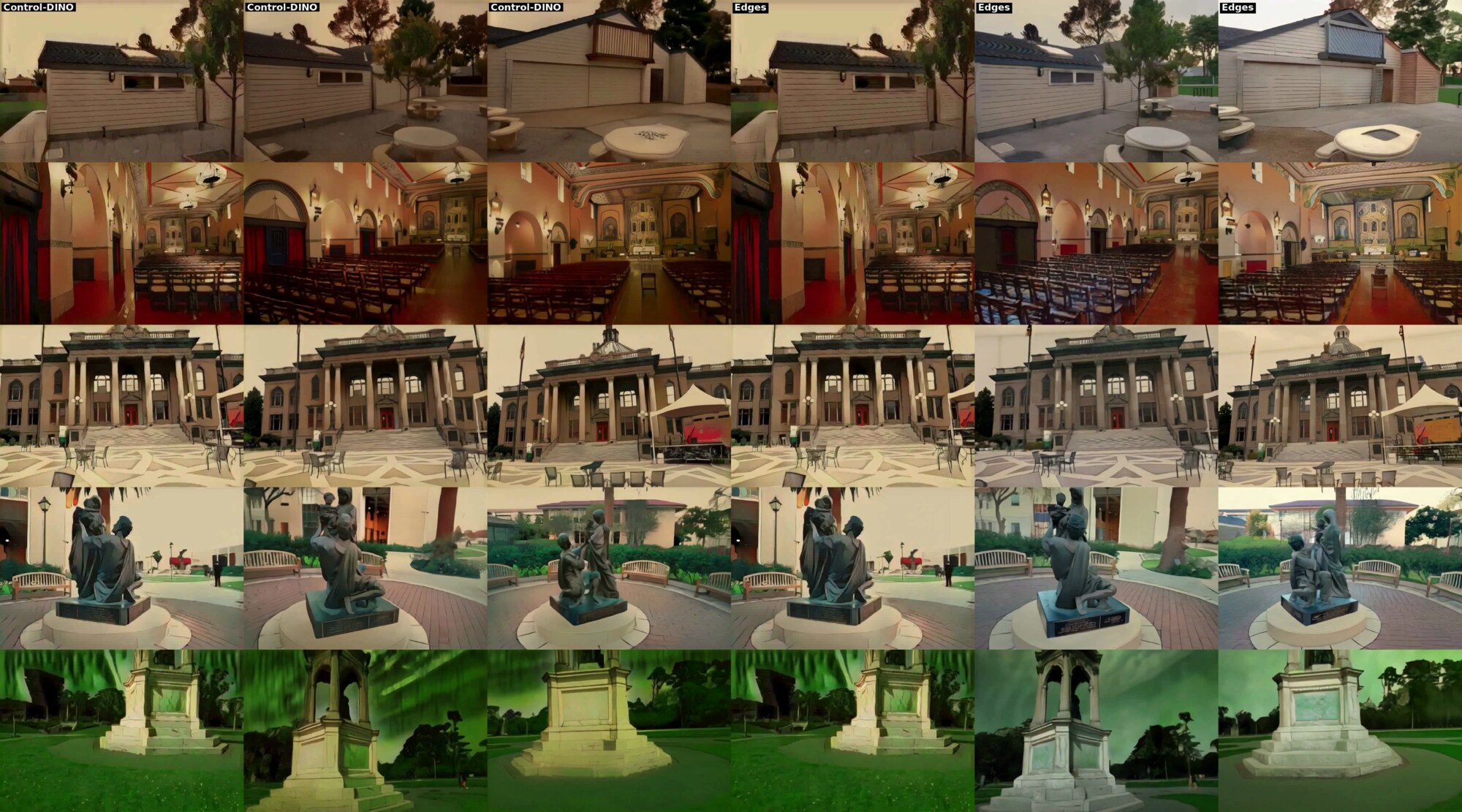}
    \caption{Control-DINO (first 3 columns) and edges (first 3 columns) conditioning results for Tanks and Temples transfer on "CartoonGAN Paprika".}
    \label{fig:cartoongan_paprika}
\end{figure*}

\begin{figure*}[t]
    \centering
    \includegraphics[width=\textwidth]{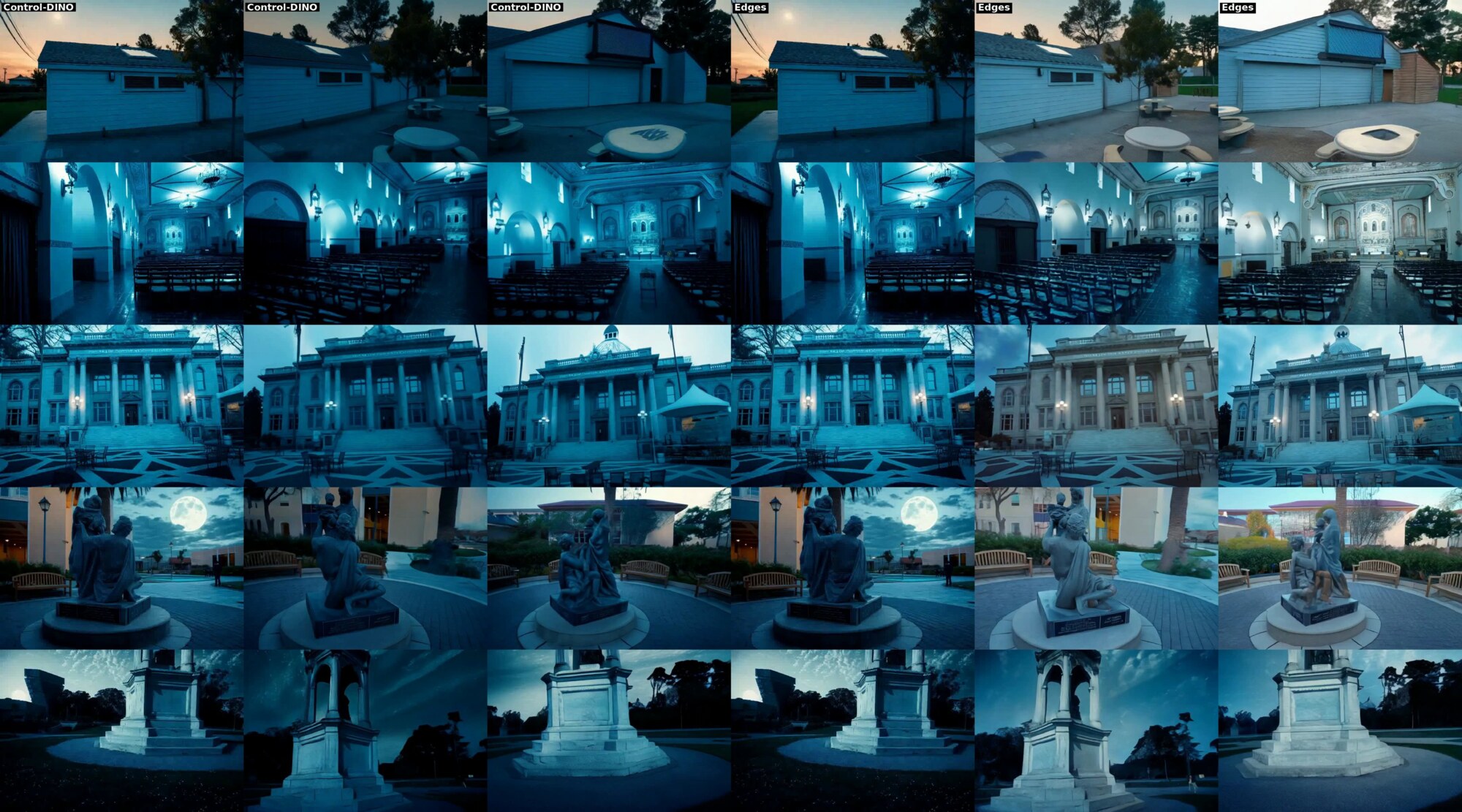}
    \caption{Control-DINO (first 3 columns) and edges (first 3 columns) conditioning results for Tanks and Temples transfer on "Cool Moonlight".}
    \label{fig:cool_moonlight}
\end{figure*}

\begin{figure*}[t]
    \centering
    \includegraphics[width=\textwidth]{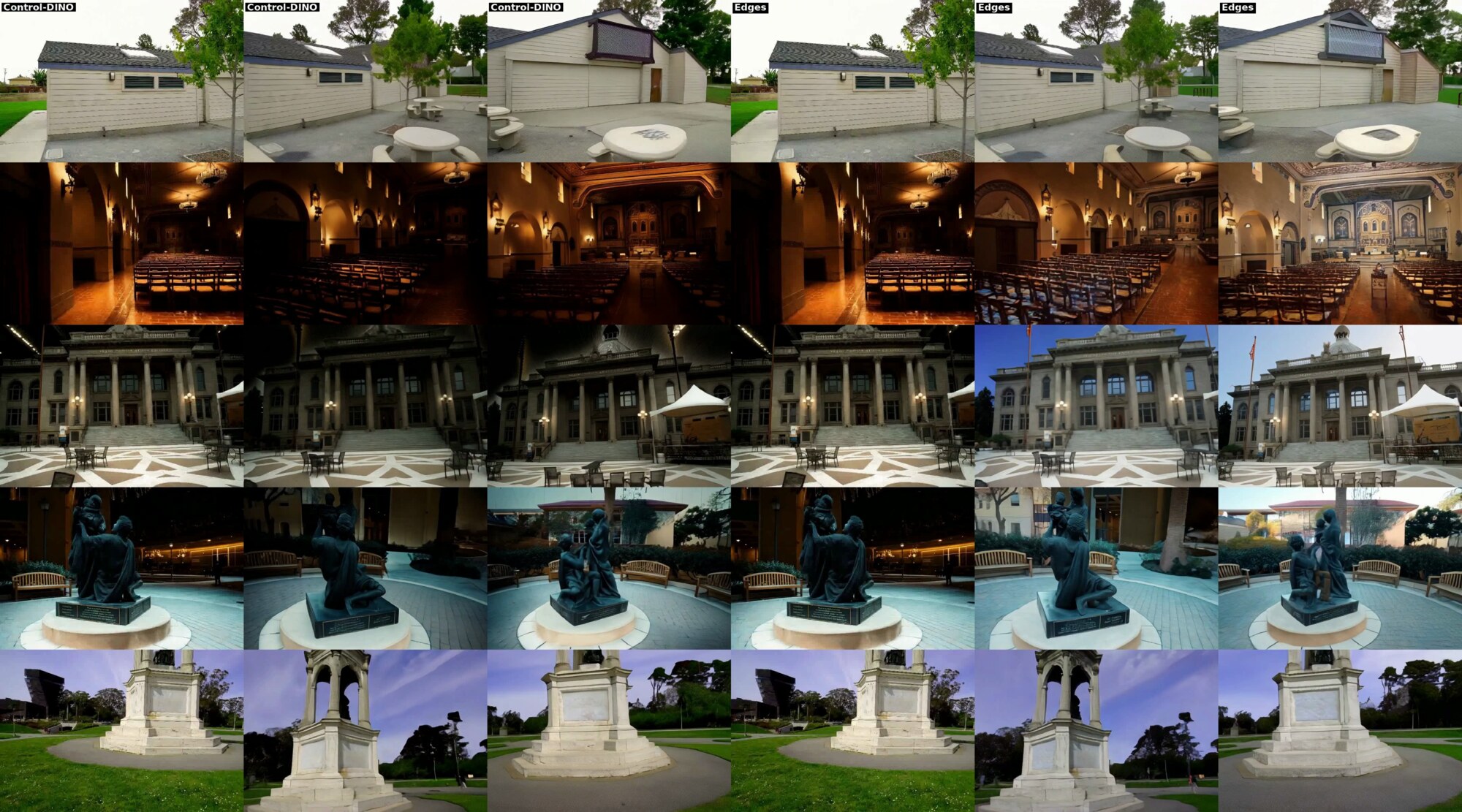}
    \caption{Control-DINO (first 3 columns) and edges (last 3 columns) conditioning results for Tanks and Temples transfer on "Dramatic Uplight".}
    \label{fig:dramatic_uplight}
\end{figure*}

\begin{figure*}[t]
    \centering
    \includegraphics[width=\textwidth]{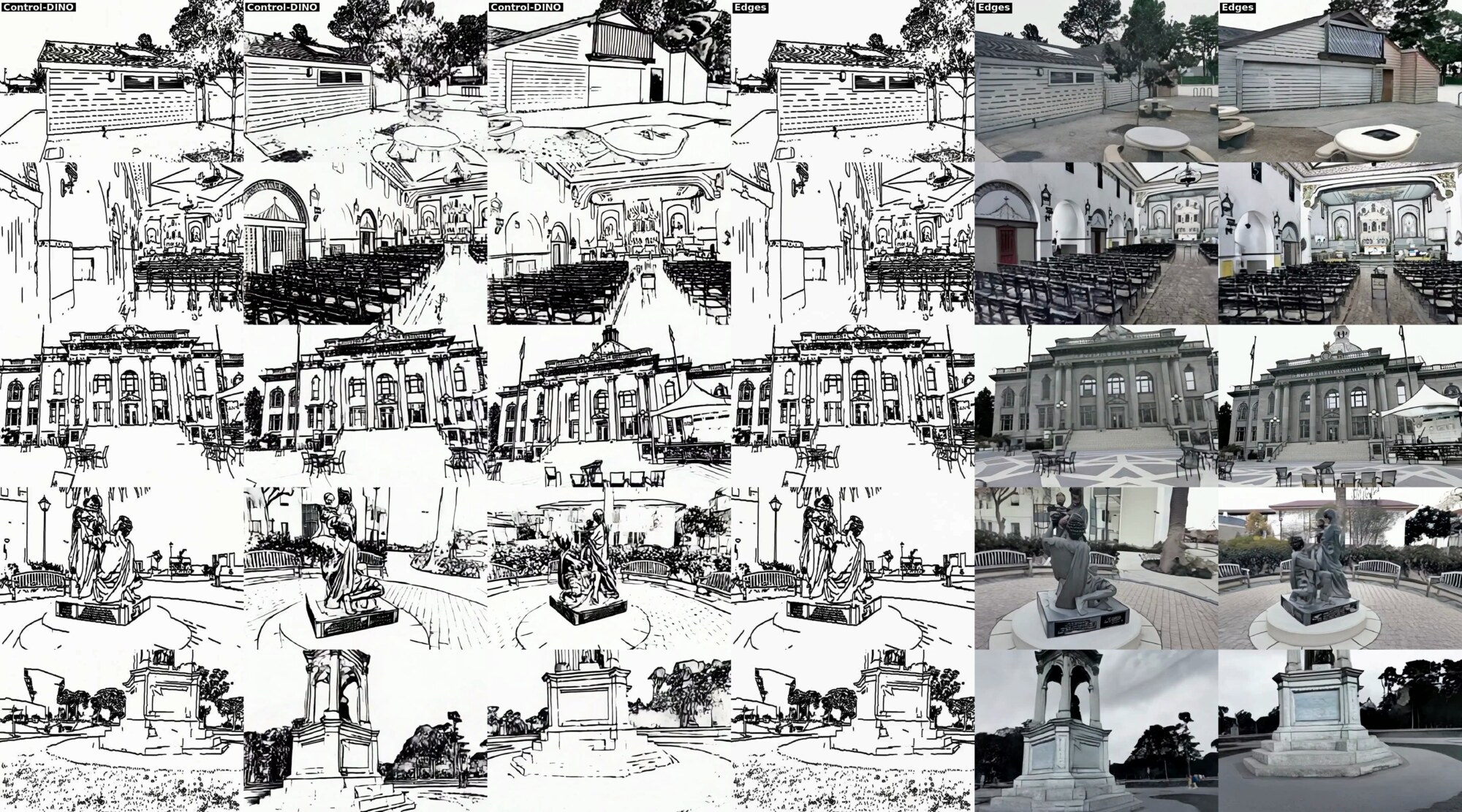}
    \caption{Control-DINO (first 3 columns) and edges (last 3 columns) conditioning results for Tanks and Temples transfer on "Drawing Contour Style".}
    \label{fig:drawing_contour}
\end{figure*}

\begin{figure*}[t]
    \centering
    \includegraphics[width=\textwidth]{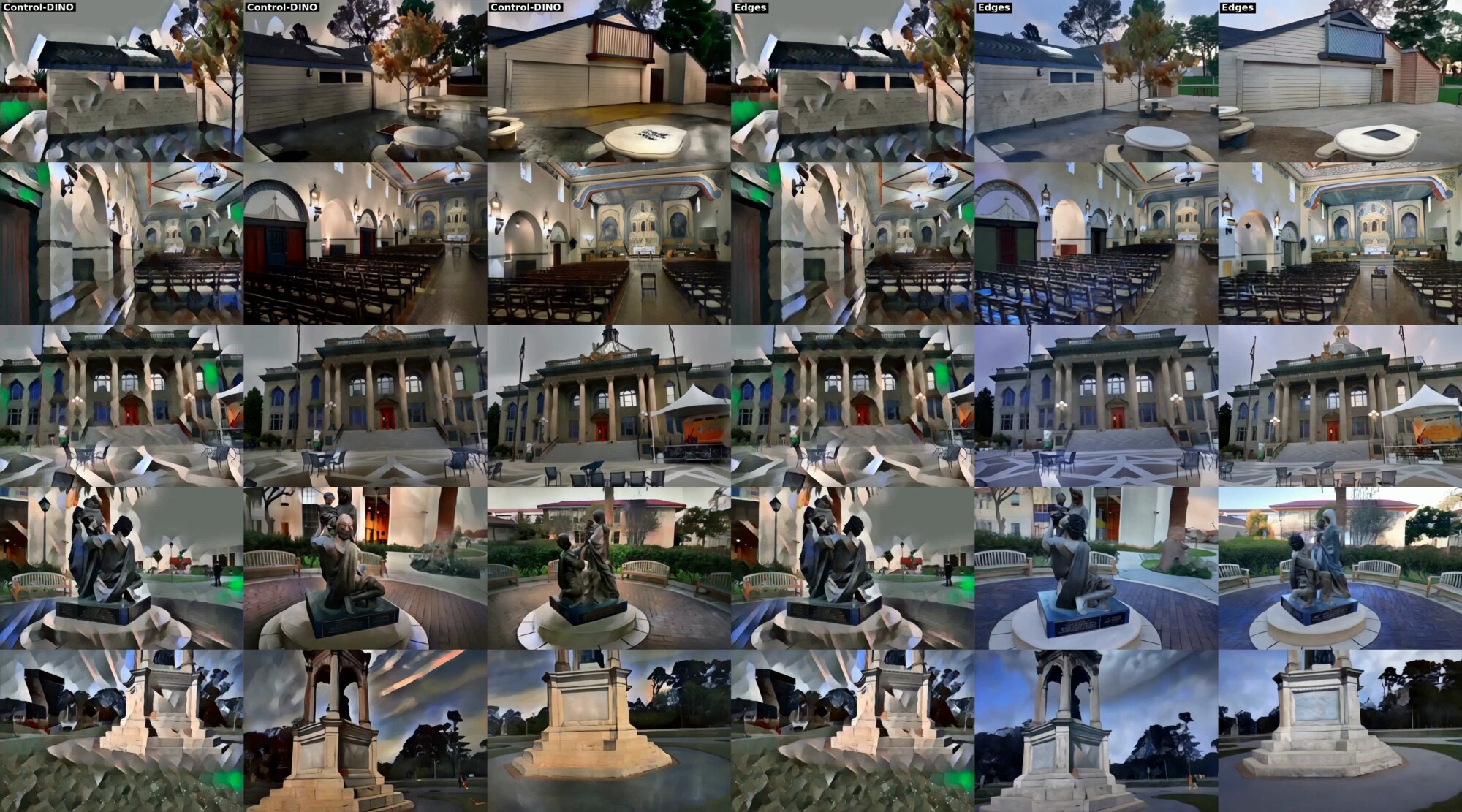}
    \caption{Control-DINO (first 3 columns) and edges (last 3 columns) conditioning results for Tanks and Temples transfer on "FastStyle Udnie".}
    \label{fig:faststyle_udnie}
\end{figure*}

\begin{figure*}[t]
    \centering
    \includegraphics[width=\textwidth]{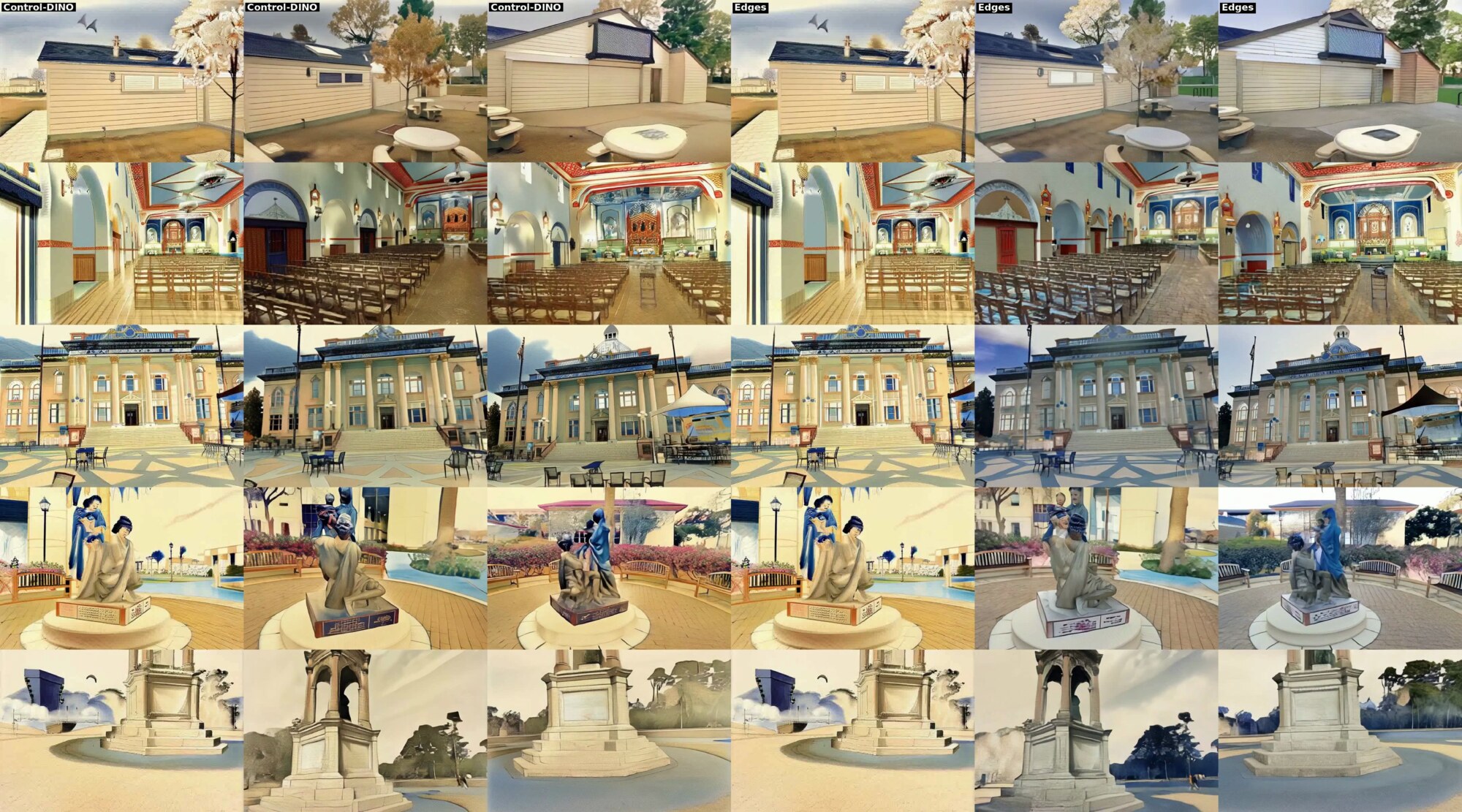}
    \caption{Control-DINO (first 3 columns) and edges (last 3 columns) conditioning results for Tanks and Temples transfer on "Hokusai Great Wave".}
    \label{fig:hokusai}
\end{figure*}

\begin{figure*}[t]
    \centering
    \includegraphics[width=\textwidth]{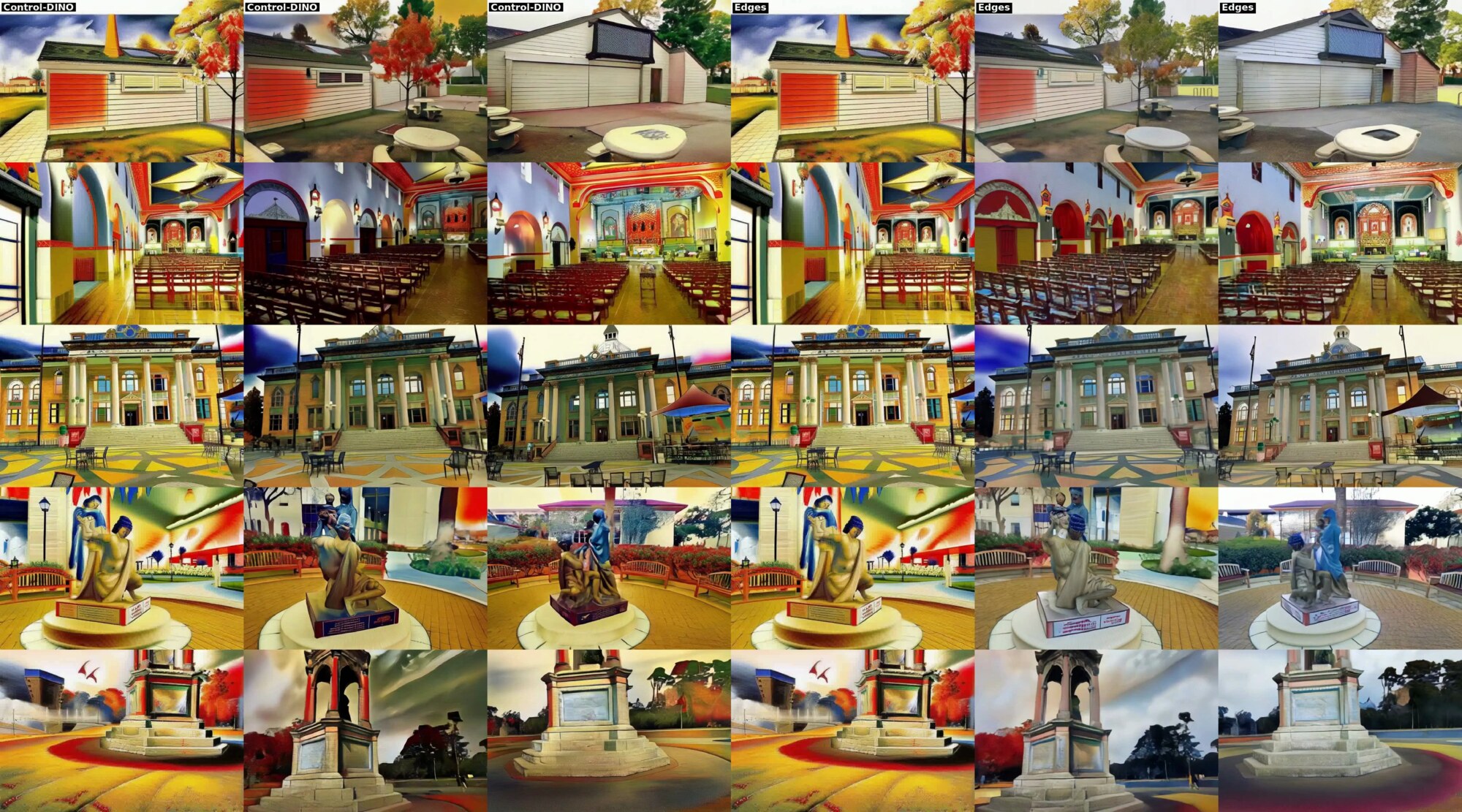}
    \caption{Control-DINO (first 3 columns) and edges (last 3 columns) conditioning results for Tanks and Temples transfer on "Kandinsky Composition VII".}
    \label{fig:kandinsky}
\end{figure*}

\begin{figure*}[t]
    \centering
    \includegraphics[width=\textwidth]{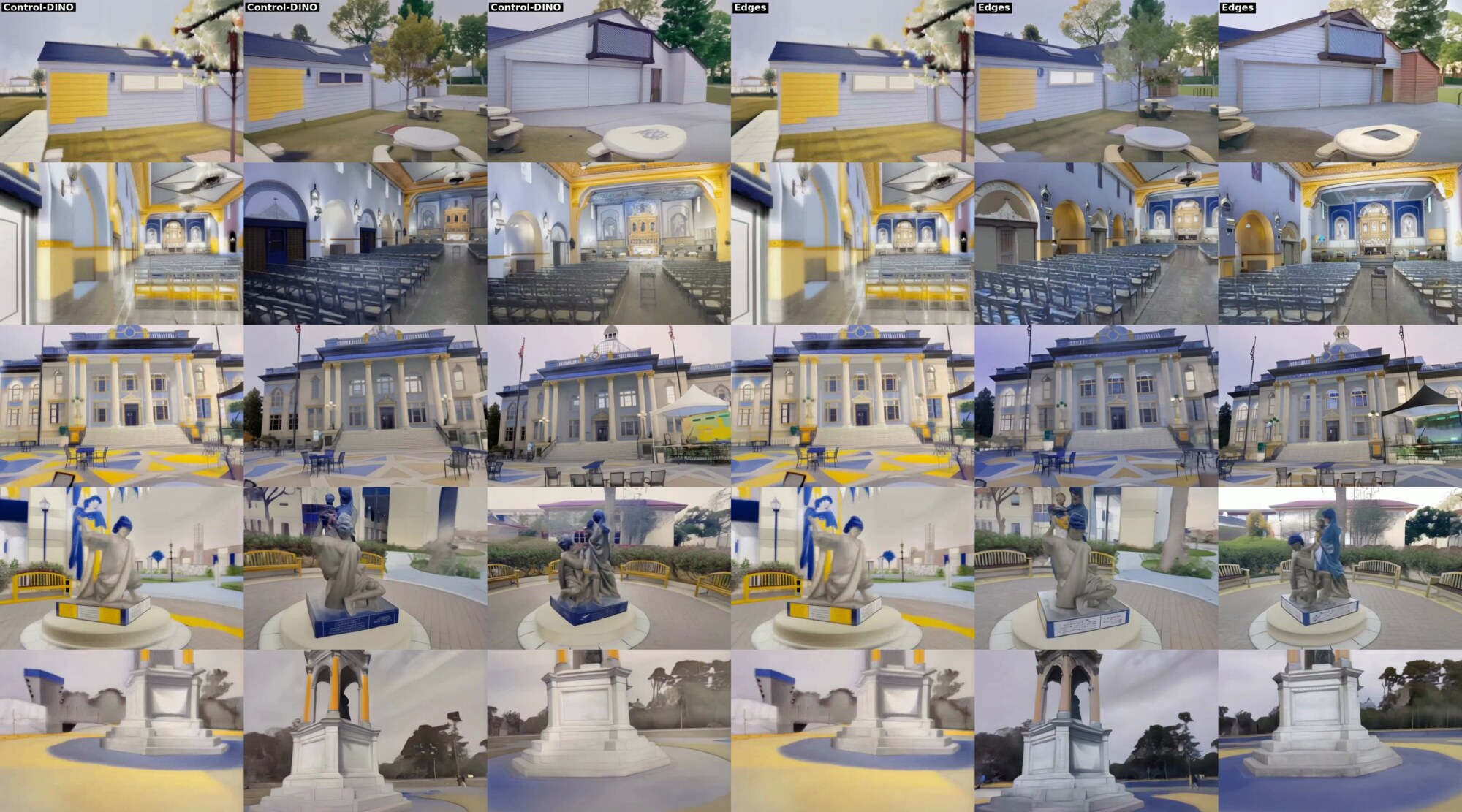}
    \caption{Control-DINO (first 3 columns) and edges (last 3 columns) conditioning results for Tanks and Temples transfer on "Mondrian Composition".}
    \label{fig:mondrian}
\end{figure*}

\begin{figure*}[t]
    \centering
    \includegraphics[width=\textwidth]{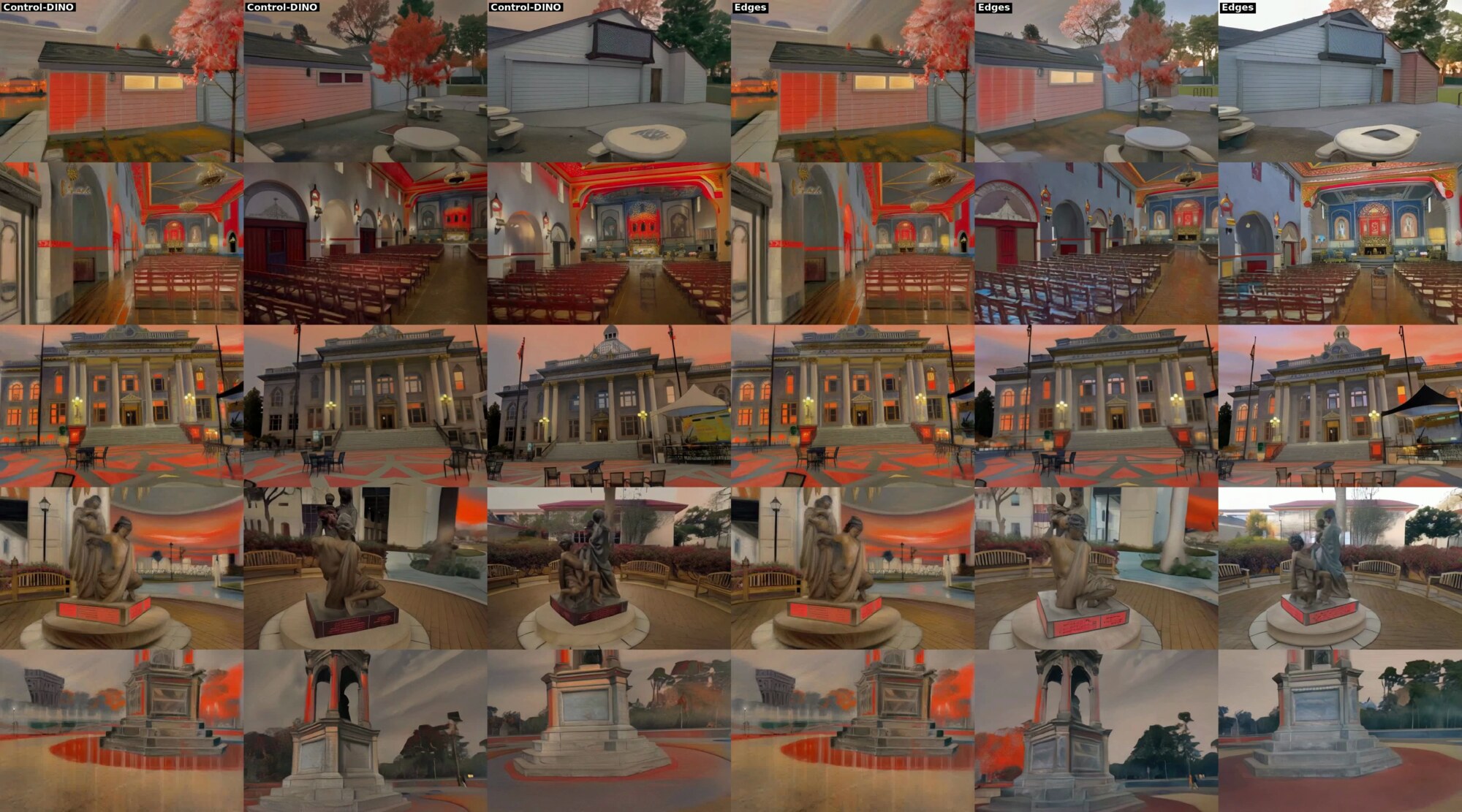}
    \caption{Control-DINO (first 3 columns) and edges (last 3 columns) conditioning results for Tanks and Temples transfer on "Monet Impression".}
    \label{fig:monet}
\end{figure*}

\begin{figure*}[t]
    \centering
    \includegraphics[width=\textwidth]{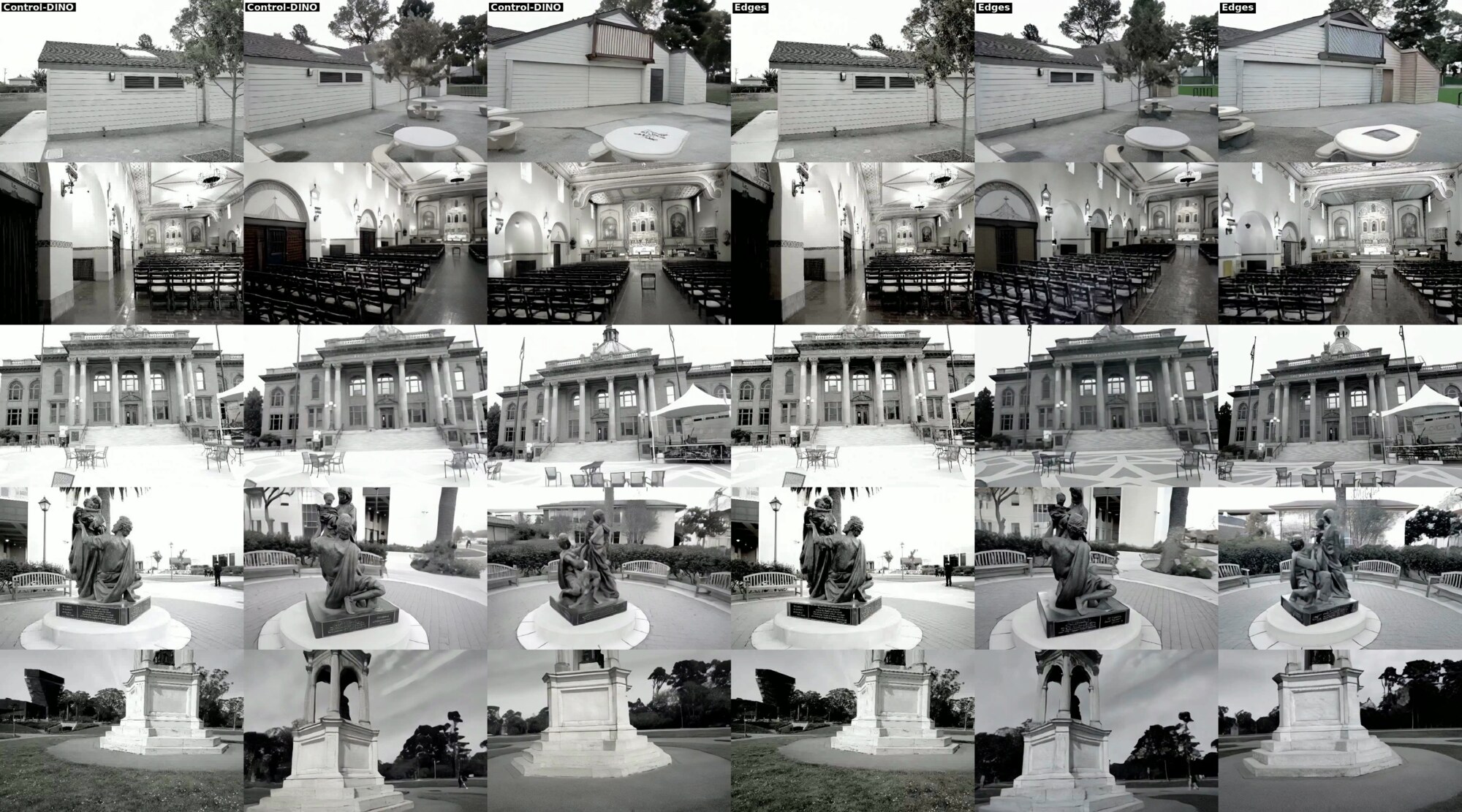}
    \caption{Control-DINO (first 3 columns) and edges (last 3 columns) conditioning results for Tanks and Temples transfer on "Photometric Noir".}
    \label{fig:photometric_noir}
\end{figure*}

\begin{figure*}[t]
    \centering
    \includegraphics[width=\textwidth]{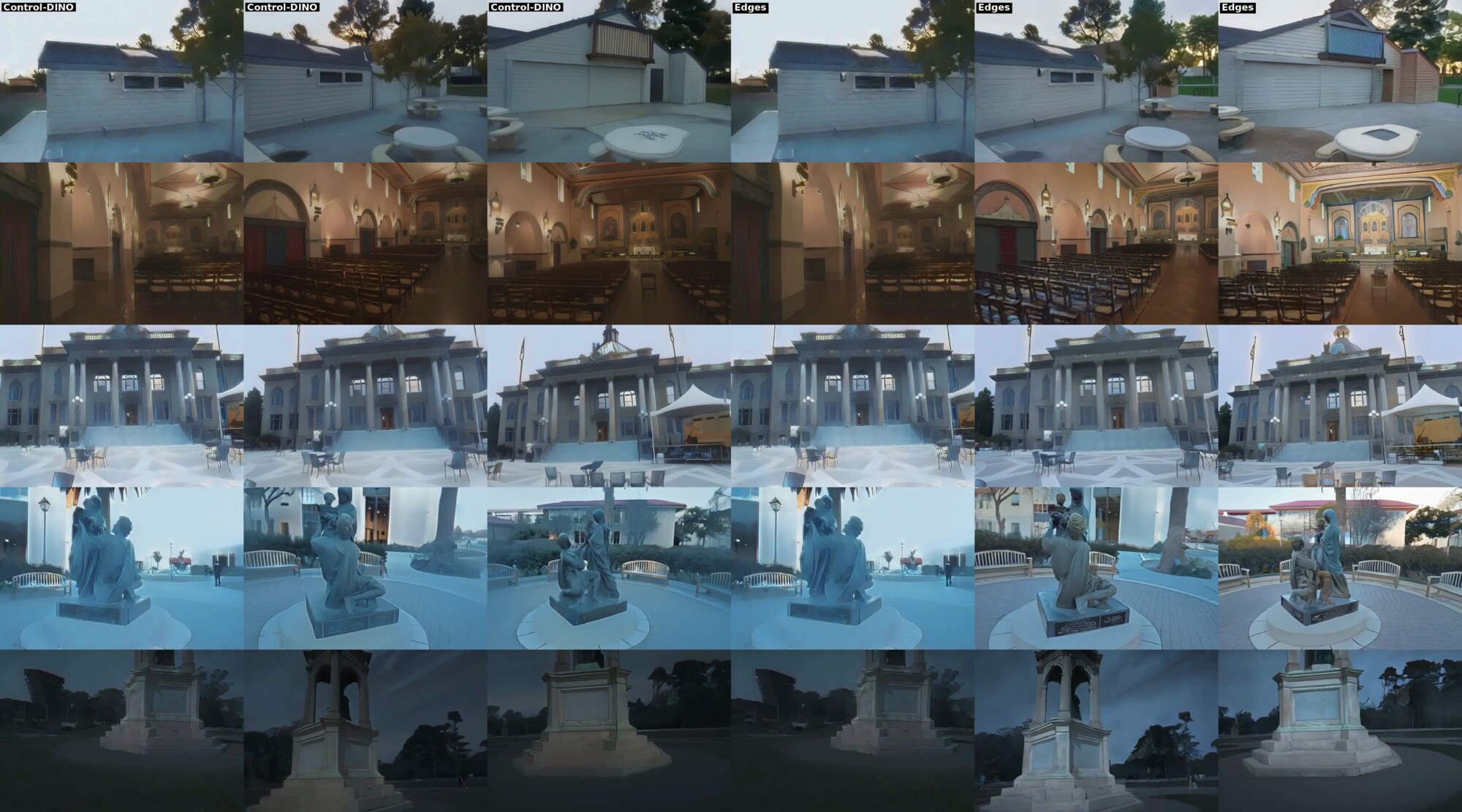}
    \caption{Control-DINO (first 3 columns) and edges (last 3 columns) conditioning results for Tanks and Temples transfer on "PTran Arcane v2".}
    \label{fig:ptran_arcane}
\end{figure*}

\begin{figure*}[t]
    \centering
    \includegraphics[width=\textwidth]{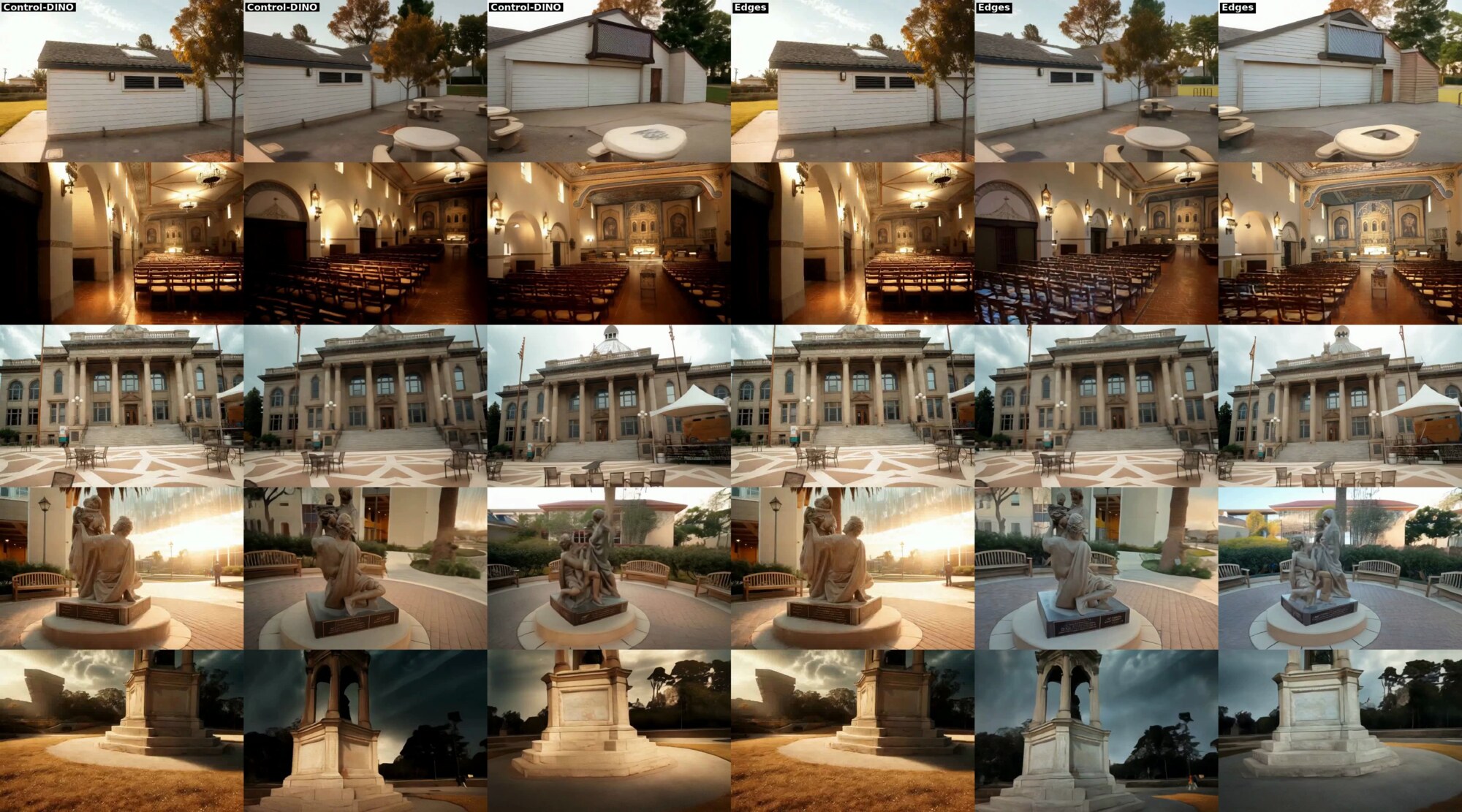}
    \caption{Control-DINO (first 3 columns) and edges (last 3 columns) conditioning results for Tanks and Temples transfer on "Soft Studio".}
    \label{fig:soft_studio}
\end{figure*}

\begin{figure*}[t]
    \centering
    \includegraphics[width=\textwidth]{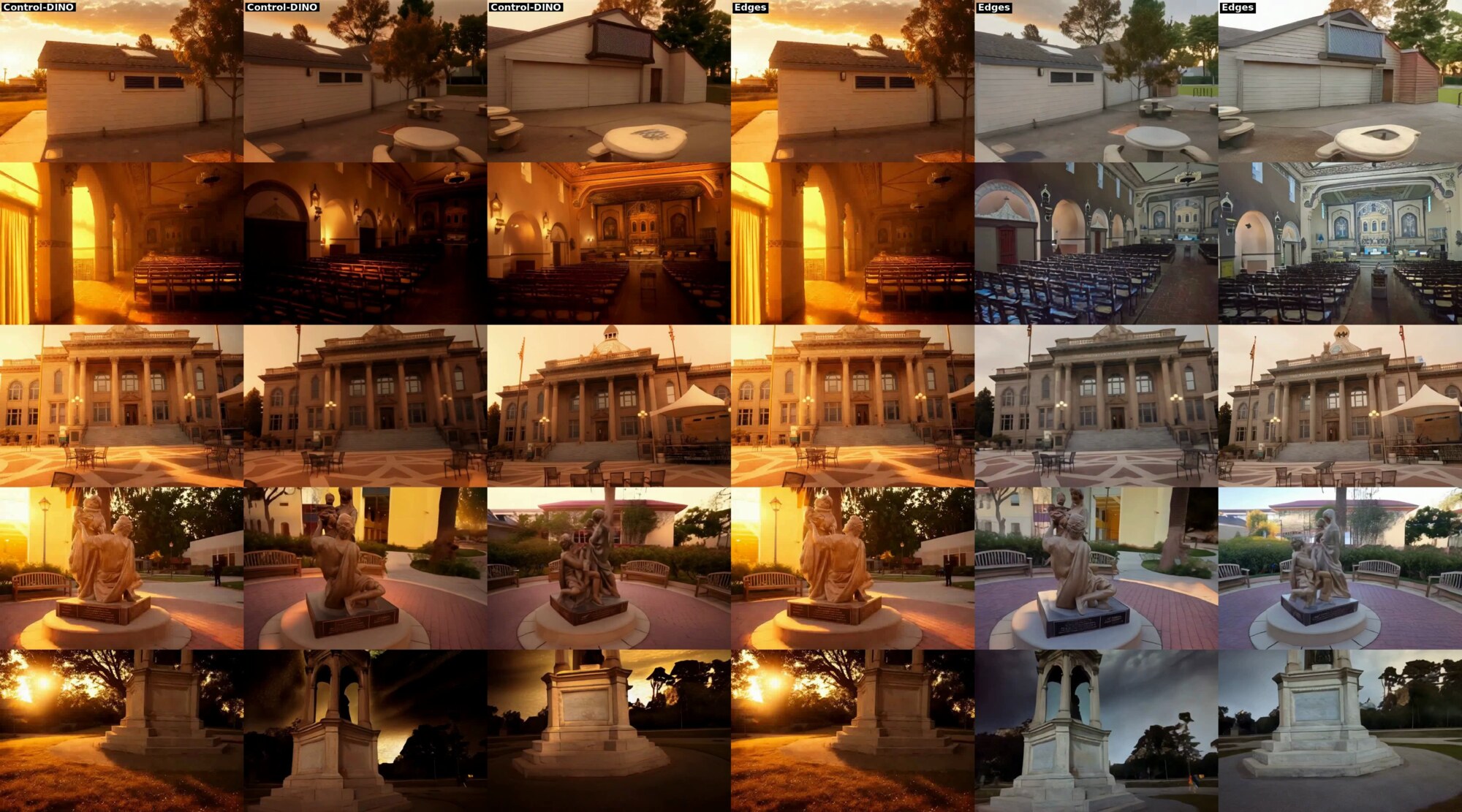}
    \caption{Control-DINO (first 3 columns) and edges (last 3 columns) conditioning results for Tanks and Temples transfer on "Sunset Left".}
    \label{fig:sunset_left}
\end{figure*}

\begin{figure*}[t]
    \centering
    \includegraphics[width=\textwidth]{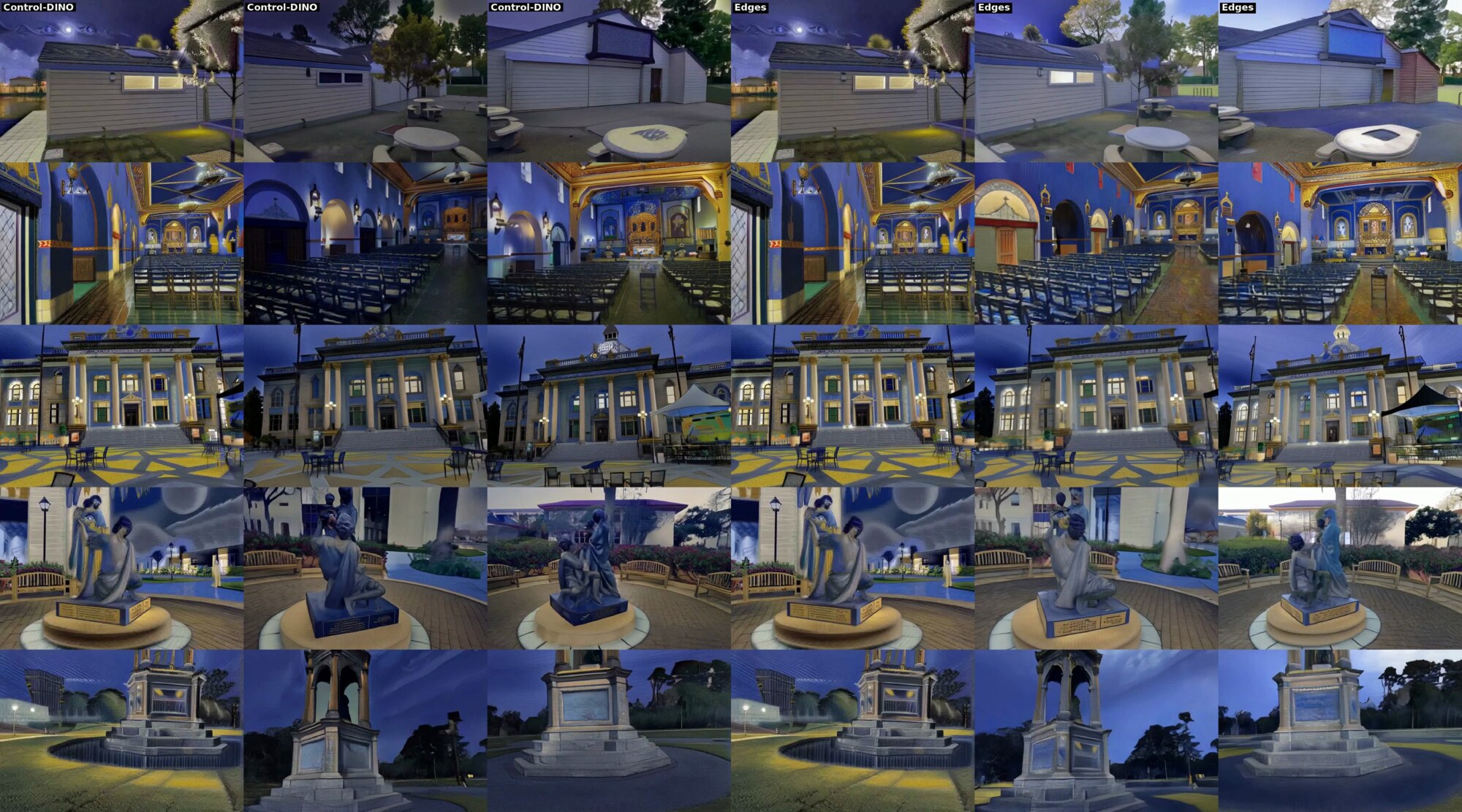}
    \caption{Control-DINO (first 3 columns) and edges (last 3 columns) conditioning results for Tanks and Temples transfer on "Van Gogh Starry Night".}
    \label{fig:vangogh}
\end{figure*}

Additional qualitative results on Tanks and Temples are shown in \autoref{fig:bright_daylight}, \ref{fig:cartoongan_paprika}, \ref{fig:cool_moonlight}, \ref{fig:dramatic_uplight}, \ref{fig:drawing_contour}, \ref{fig:faststyle_udnie}, \ref{fig:hokusai}, \ref{fig:kandinsky}, \ref{fig:mondrian}, \ref{fig:monet}, \ref{fig:photometric_noir}, \ref{fig:ptran_arcane}, \ref{fig:soft_studio}, \ref{fig:sunset_left}, and~\ref{fig:vangogh}.

We evaluate geometric consistency by running COLMAP on generated video frames using \texttt{pycolmap}, reporting the fraction of frames successfully registered.
Fréchet Inception Distance (FID) is measured with \texttt{clean-fid}~\cite{parmar2021cleanfid} using the default Inception-V3 features, with frames sampled every 4 steps from both generated and reference videos.
Style alignment is assessed via cosine similarity between \texttt{CLIP ViT-L/14}~\cite{radford2021learning} embeddings of the style reference image and the mean embedding across all generated frames.

\begin{figure}[htbp]
  \centering
  \setlength{\tabcolsep}{1pt}
  \renewcommand{\arraystretch}{1.0}
  \scriptsize
  \textbf{Scene 1}\\[2pt]
  \begin{tabular}{@{}c c c c c @{}}
     & \textbf{2D} & \textbf{Voxel} & \textbf{Mesh} & \textbf{Concerto} \\
    \raisebox{6ex}{\rotatebox{90}{\tiny 720p}} &
    \includegraphics[width=0.24\linewidth]{figures/cond_signal_comparison/scene0/2d/2d_cond/debug_3d_features_scale2_2d_frame_000001.png} &
    \includegraphics[width=0.24\linewidth]{figures/cond_signal_comparison/scene0/voxel/voxel_cond/debug_3d_features_scale2_voxel_vs0.02_frame_000001.png} &
     \includegraphics[width=0.24\linewidth]{figures/cond_signal_comparison/scene0/mesh/mesh_cond/meshframe0_pca.png} &
    \includegraphics[width=0.24\linewidth]{figures/cond_signal_comparison/scene0/concerto/concerto_cond/debug_3d_features_scale2_color_concerto_voxel_vs0.02_frame_000001.png} 
 \\
     \raisebox{6ex}{\rotatebox{90}{\tiny 60x90}} &
    \includegraphics[width=0.24\linewidth]{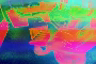} &
    \includegraphics[width=0.24\linewidth]{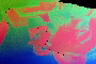} &
     \includegraphics[width=0.24\linewidth]{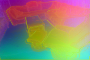} &
    \includegraphics[width=0.24\linewidth]{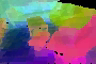}
 \\
  \end{tabular}
\caption{PCA visualization of the high-resolution and true resolution conditioning feature maps (60$\times$90).}  \label{fig:cond_signal}
\end{figure}

\begin{figure}[htbp]
  \centering
  \setlength{\tabcolsep}{0.5pt}
  \renewcommand{\arraystretch}{1.0}
  \scriptsize
  \begin{tabular}{@{} c c c c c c c c @{}}
   &  t=1 & t=35 & t=49 &  & t=1 & t=35 & t=49 \\
  \raisebox{3ex}{\rotatebox{90}{\tiny $GT$}} &
      \includegraphics[width=0.16\linewidth]{figures/cond_signal_comparison/scene0/gt/00000_1_reference_frame_000001.jpg} &
    \includegraphics[width=0.16\linewidth]{figures/cond_signal_comparison/scene0/gt/00000_1_reference_frame_000035.jpg} &
    \includegraphics[width=0.16\linewidth]{figures/cond_signal_comparison/scene0/gt/00000_1_reference_frame_000049.jpg} &
    \raisebox{3ex}{\rotatebox{90}{\tiny $GT$}} &
      \includegraphics[width=0.16\linewidth]{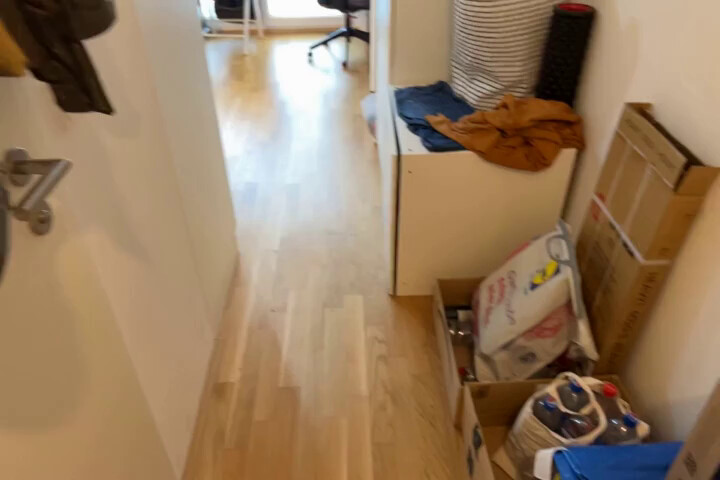} &
    \includegraphics[width=0.16\linewidth]{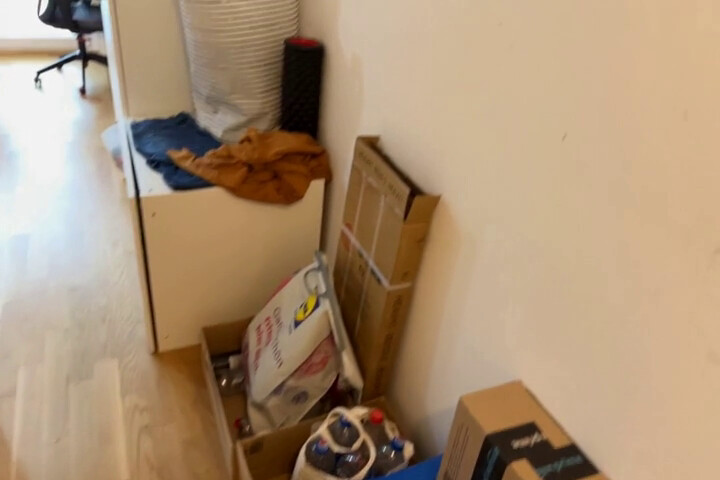} &
    \includegraphics[width=0.16\linewidth]{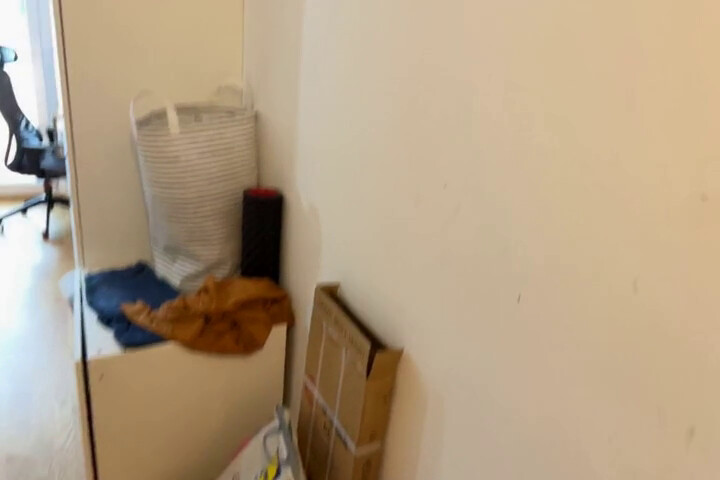}\\

  \raisebox{3ex}{\rotatebox{90}{\tiny $Ours$}} &
      \includegraphics[width=0.16\linewidth]{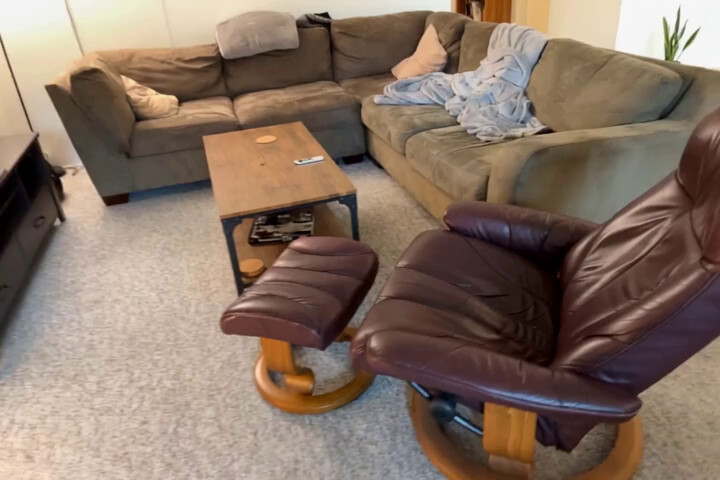} &
    \includegraphics[width=0.16\linewidth]{figures/cond_signal_comparison/scene0/2d/2d_out/00000_1_out_frame_000035.jpg} &
    \includegraphics[width=0.16\linewidth]{figures/cond_signal_comparison/scene0/2d/2d_out/00000_1_out_frame_000049.jpg} &
    \raisebox{3ex}{\rotatebox{90}{\tiny $Ours$}} &
    \includegraphics[width=0.16\linewidth]{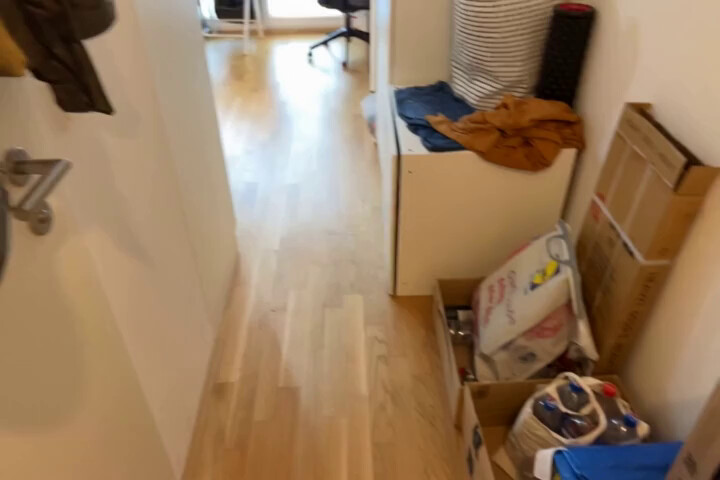} &
    \includegraphics[width=0.16\linewidth]{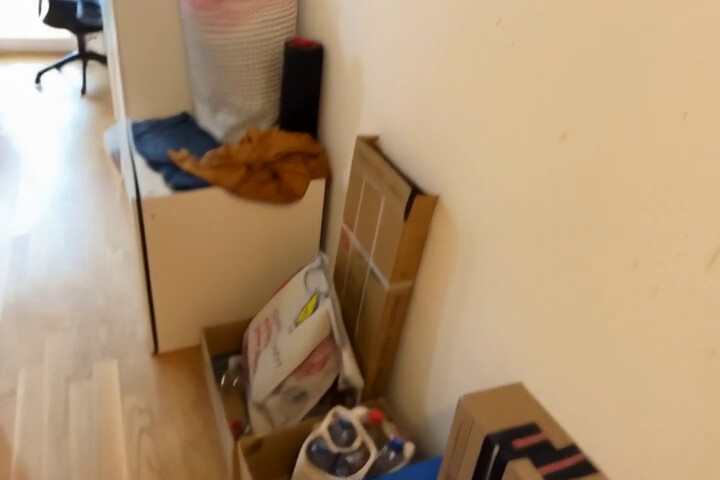} & 
    \includegraphics[width=0.16\linewidth]{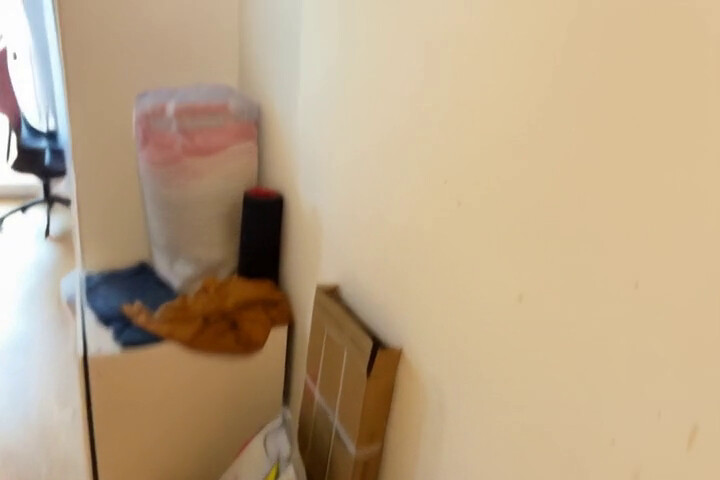}\\

      \raisebox{3ex}{\rotatebox{90}{\tiny $Edge$}} &
      \includegraphics[width=0.16\linewidth]{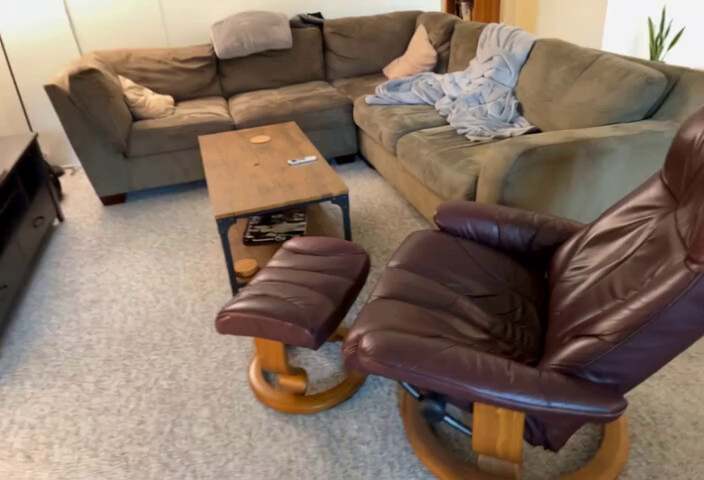} &
    \includegraphics[width=0.16\linewidth]{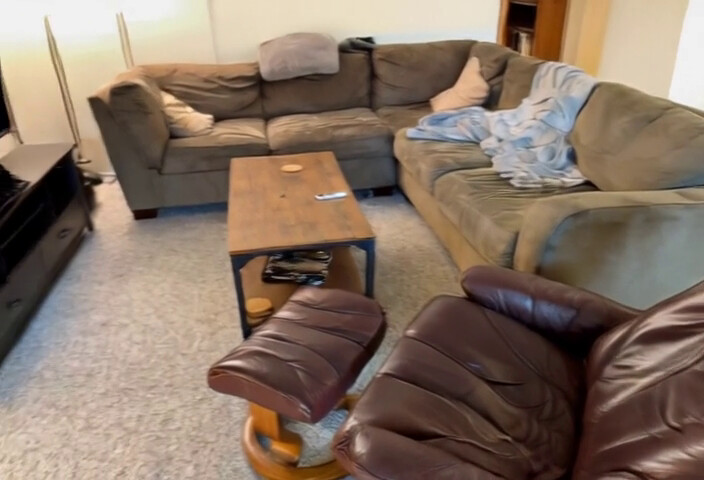} &
    \includegraphics[width=0.16\linewidth]{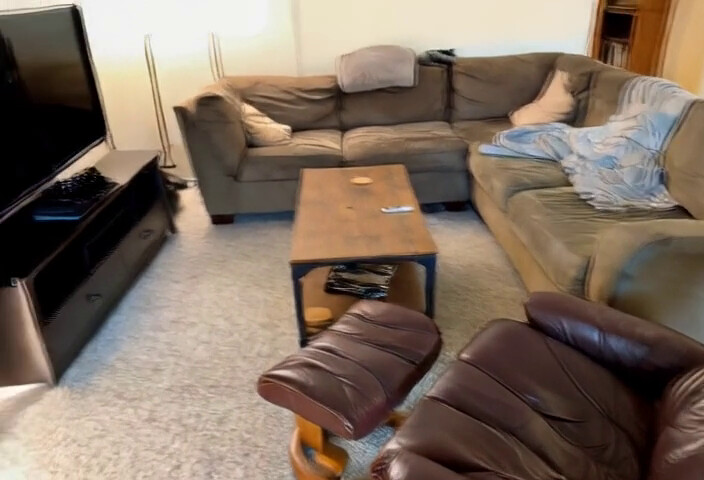} &
    \raisebox{3ex}{\rotatebox{90}{\tiny $Edge$}} &
    \includegraphics[width=0.16\linewidth]{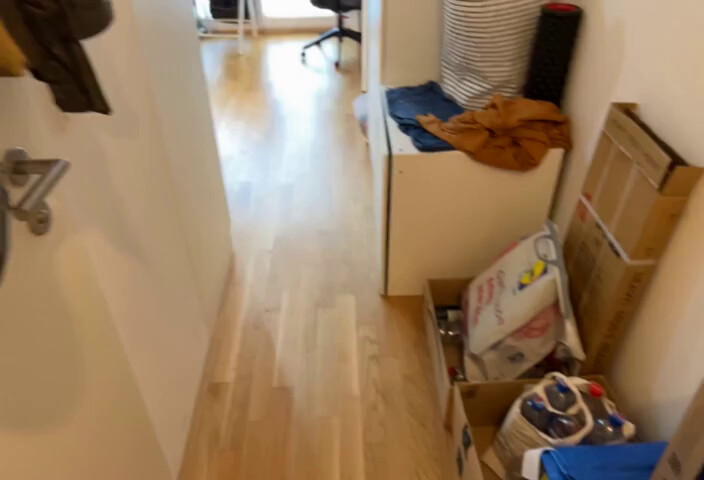} &
    \includegraphics[width=0.16\linewidth]{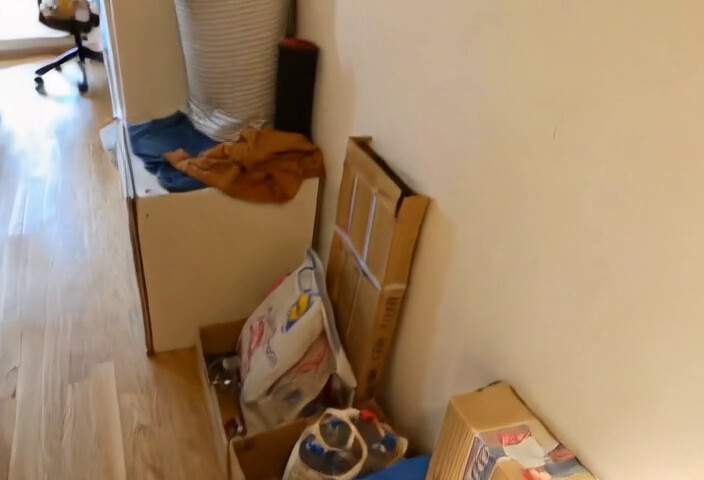} & 
    \includegraphics[width=0.16\linewidth]{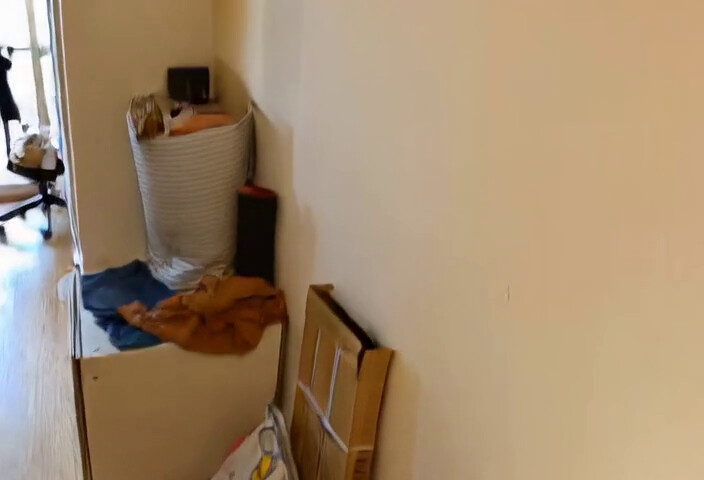}\\

      \raisebox{3ex}{\rotatebox{90}{\tiny $Depth$}} &
      \includegraphics[width=0.16\linewidth]{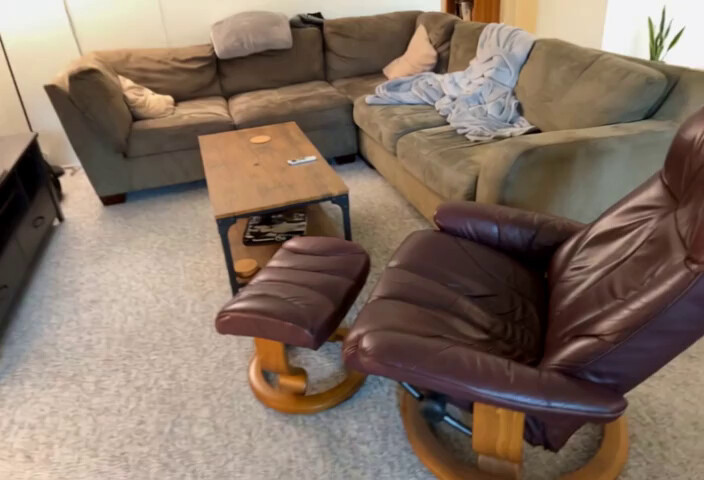} &
    \includegraphics[width=0.16\linewidth]{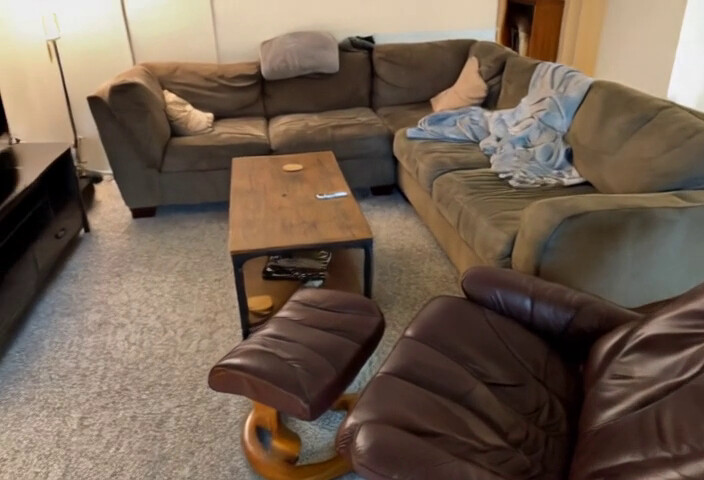} &
    \includegraphics[width=0.16\linewidth]{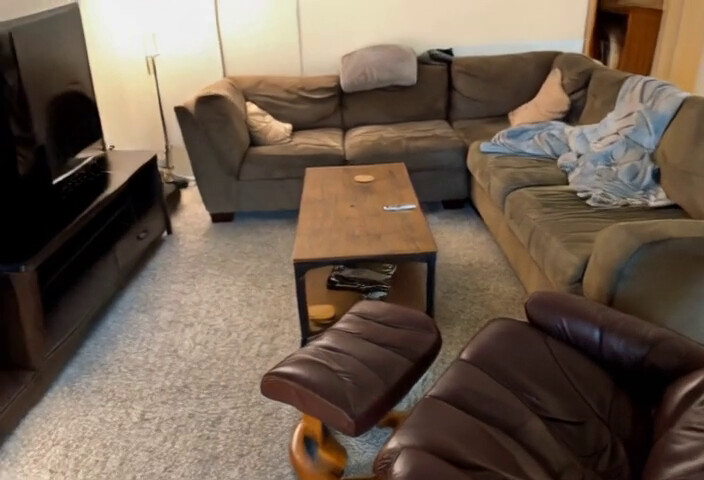} &
    \raisebox{3ex}{\rotatebox{90}{\tiny $Depth$}} &
    \includegraphics[width=0.16\linewidth]{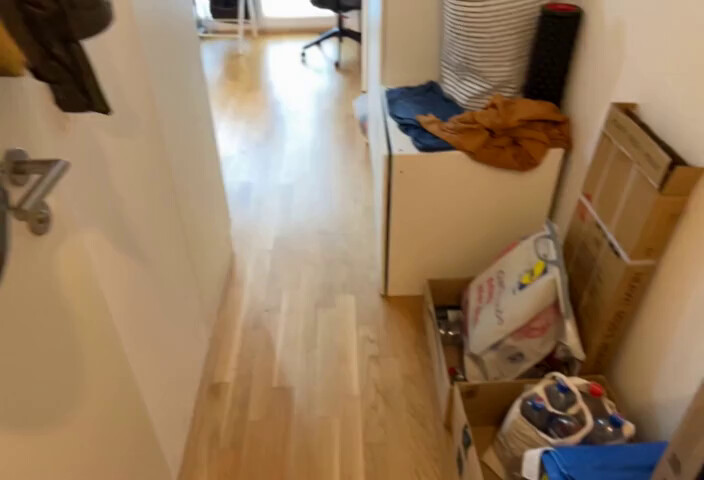} &
    \includegraphics[width=0.16\linewidth]{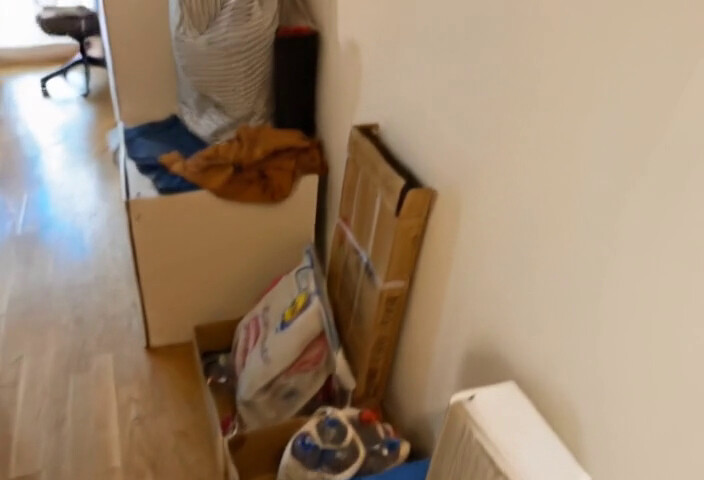} & 
    \includegraphics[width=0.16\linewidth]{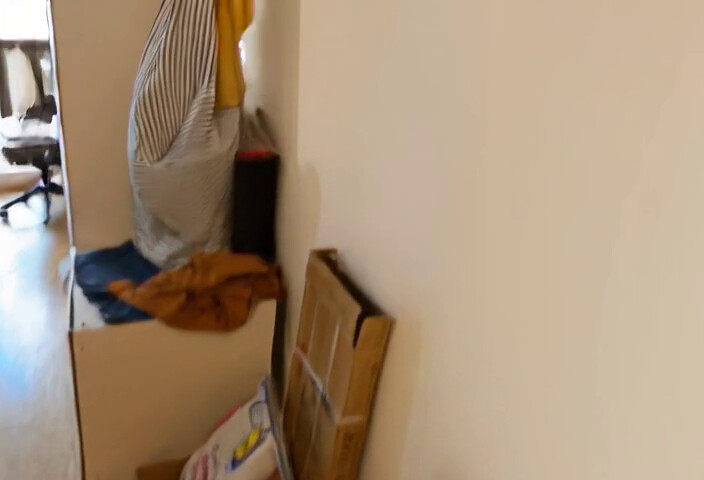}\\
  \end{tabular}

  \begin{tabular}{@{} c c c c c c c c @{}}
   &  t=1 & t=35 & t=49 &  & t=1 & t=35 & t=49 \\
  \raisebox{3ex}{\rotatebox{90}{\tiny $GT$}} &
      \includegraphics[width=0.16\linewidth]{figures/cond_signal_comparison/scene8/gt/00008_1_reference_frame_000001.jpg} &
    \includegraphics[width=0.16\linewidth]{figures/cond_signal_comparison/scene8/gt/00008_1_reference_frame_000035.jpg} &
    \includegraphics[width=0.16\linewidth]{figures/cond_signal_comparison/scene8/gt/00008_1_reference_frame_000049.jpg} &
    \raisebox{3ex}{\rotatebox{90}{\tiny $GT$}} &
      \includegraphics[width=0.16\linewidth]{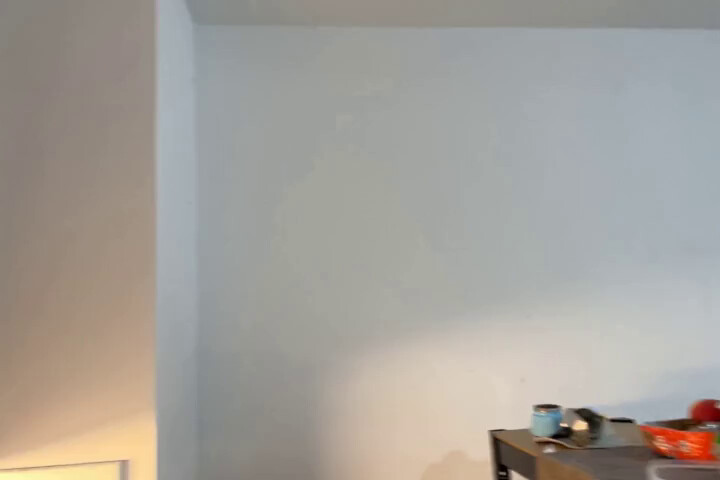} &
    \includegraphics[width=0.16\linewidth]{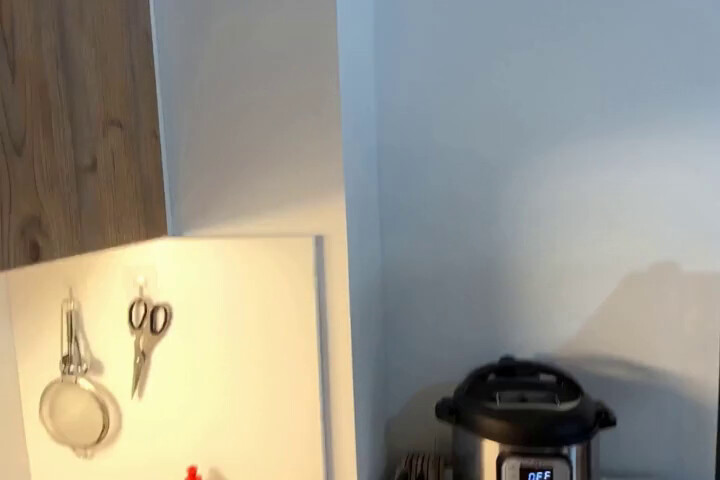} &
    \includegraphics[width=0.16\linewidth]{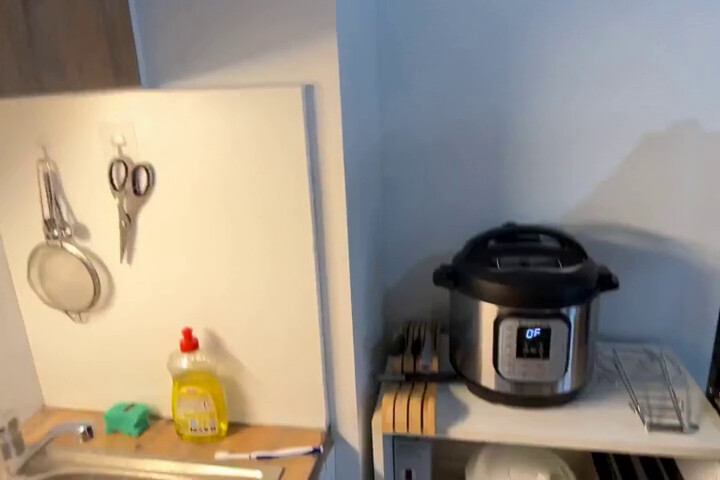}\\

  \raisebox{3ex}{\rotatebox{90}{\tiny $Ours$}} &
      \includegraphics[width=0.16\linewidth]{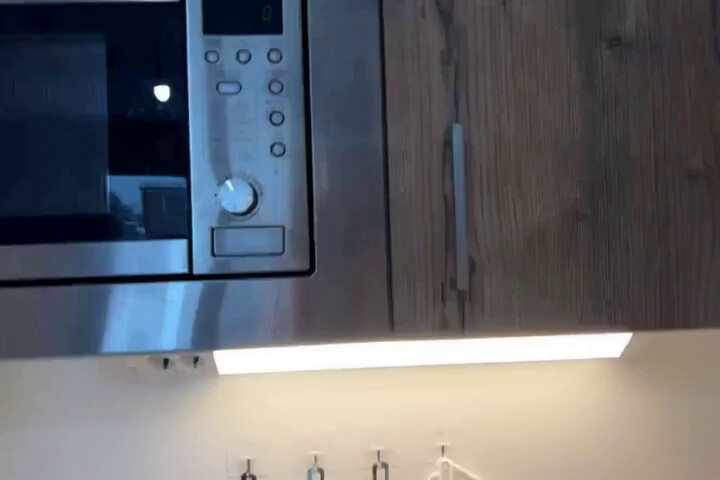} &
    \includegraphics[width=0.16\linewidth]{figures/cond_signal_comparison/scene8/2d/2d_out/00008_1_out_frame_000035.jpg} &
    \includegraphics[width=0.16\linewidth]{figures/cond_signal_comparison/scene8/2d/2d_out/00008_1_out_frame_000049.jpg} &
    \raisebox{3ex}{\rotatebox{90}{\tiny $Ours$}} &
    \includegraphics[width=0.16\linewidth]{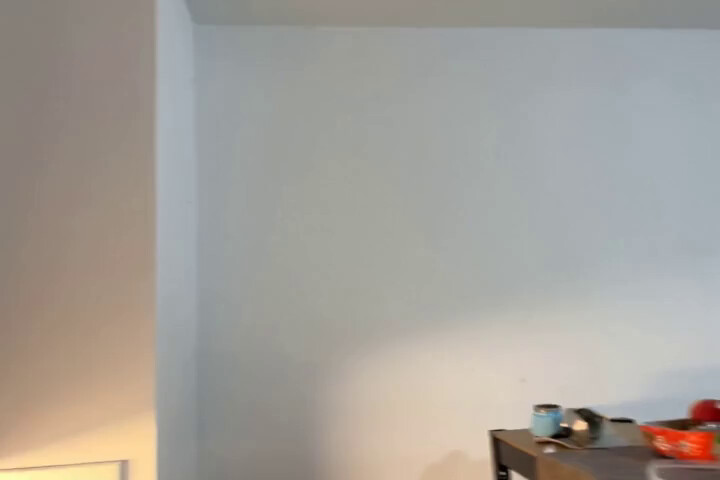} &
    \includegraphics[width=0.16\linewidth]{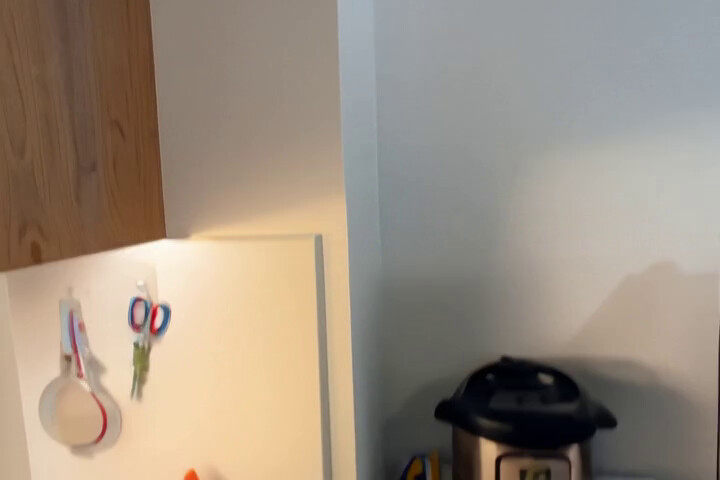} & 
    \includegraphics[width=0.16\linewidth]{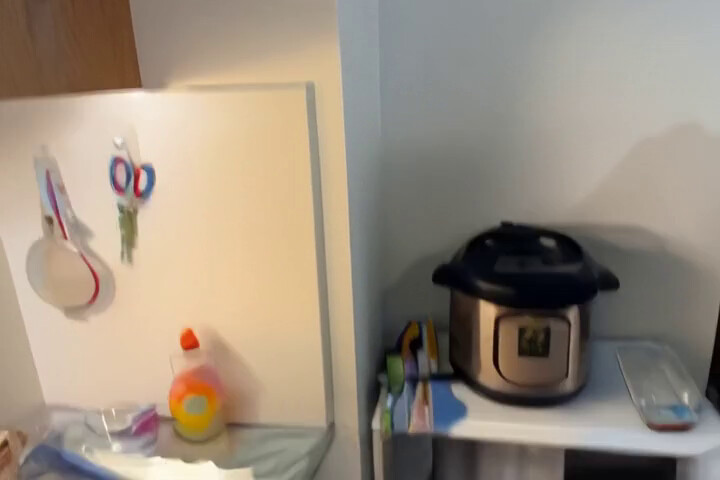}\\

      \raisebox{3ex}{\rotatebox{90}{\tiny $Edge$}} &
      \includegraphics[width=0.16\linewidth]{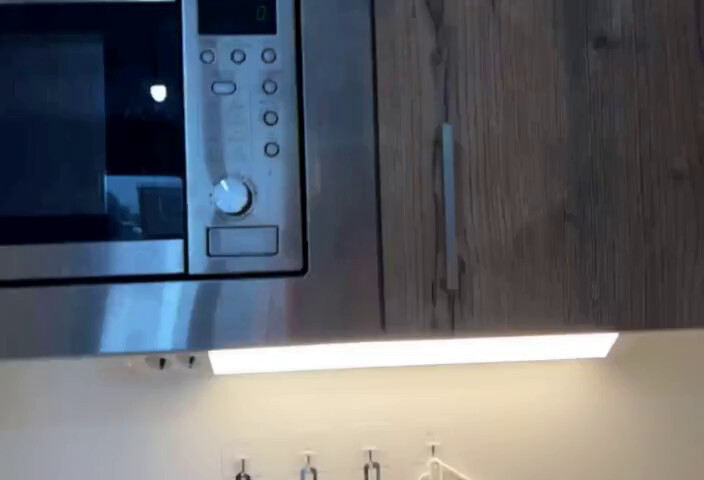} &
    \includegraphics[width=0.16\linewidth]{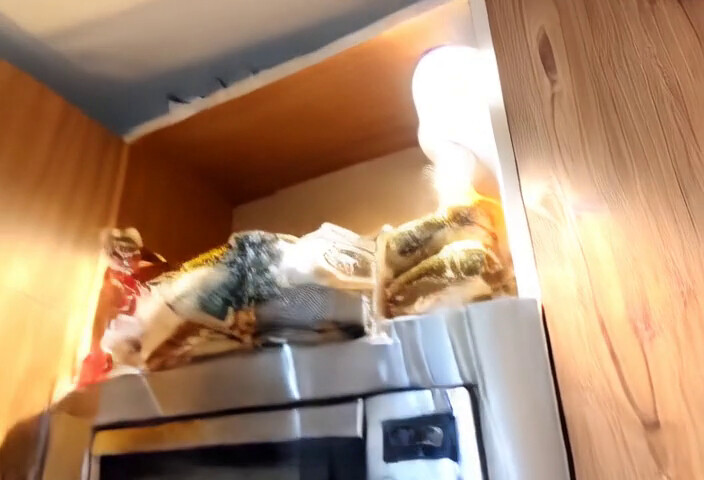} &
    \includegraphics[width=0.16\linewidth]{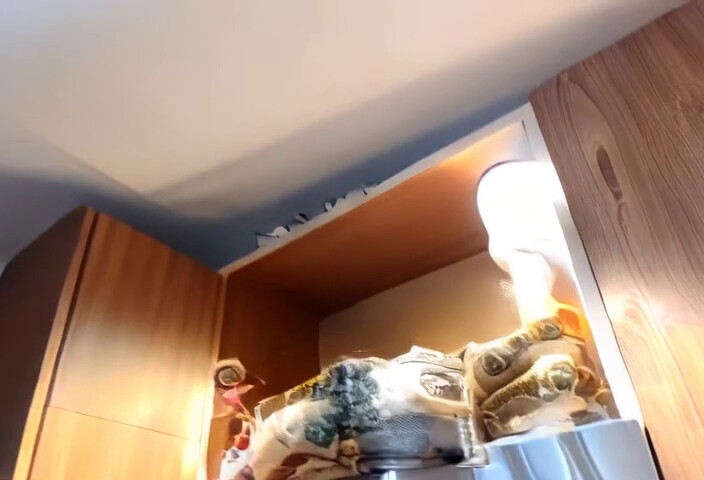} &
    \raisebox{3ex}{\rotatebox{90}{\tiny $Edge$}} &
    \includegraphics[width=0.16\linewidth]{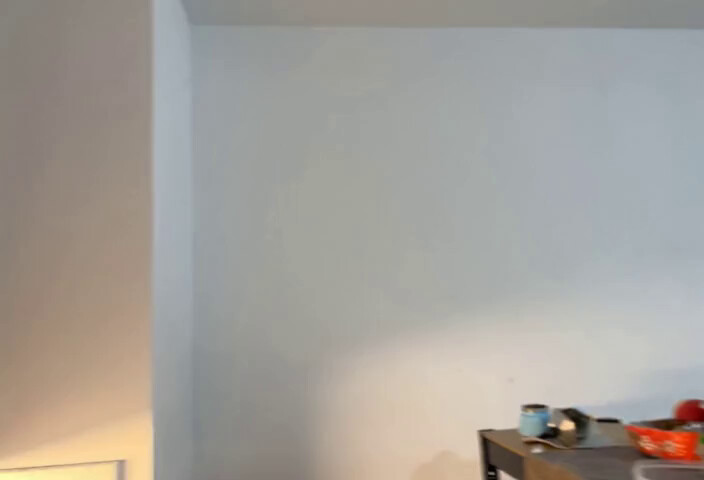} &
    \includegraphics[width=0.16\linewidth]{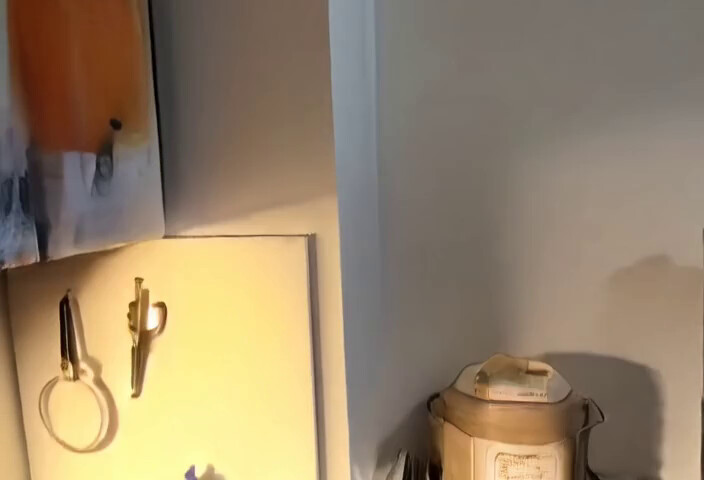} & 
    \includegraphics[width=0.16\linewidth]{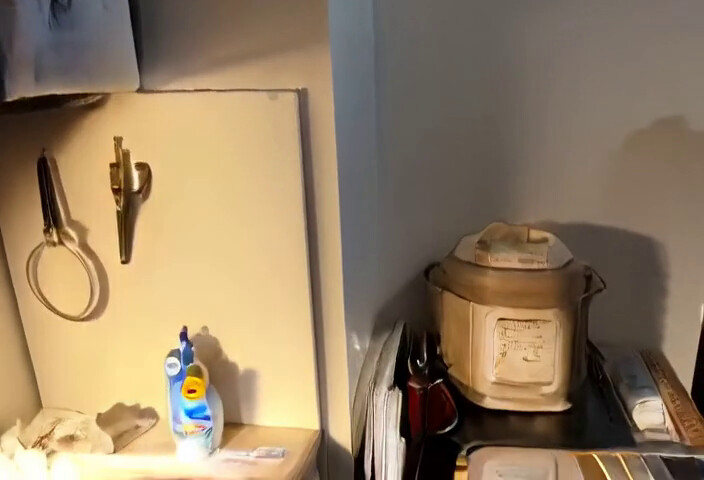}\\

      \raisebox{3ex}{\rotatebox{90}{\tiny $Depth$}} &
      \includegraphics[width=0.16\linewidth]{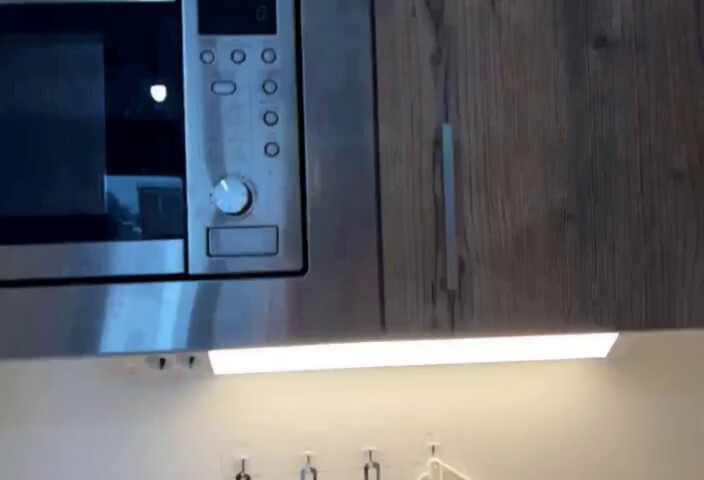} &
    \includegraphics[width=0.16\linewidth]{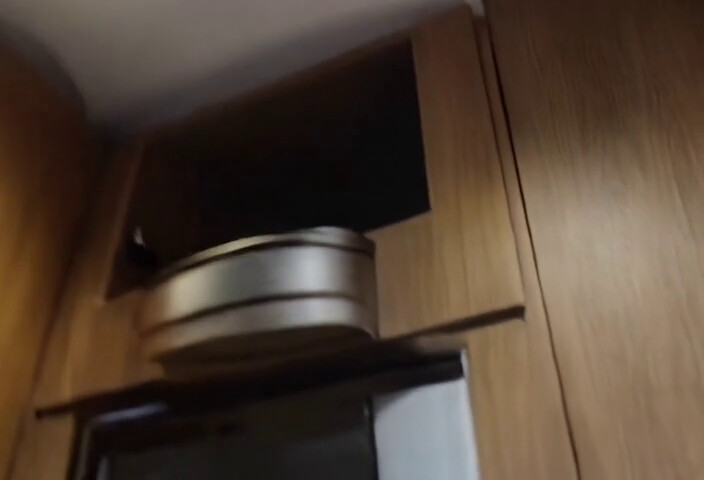} &
    \includegraphics[width=0.16\linewidth]{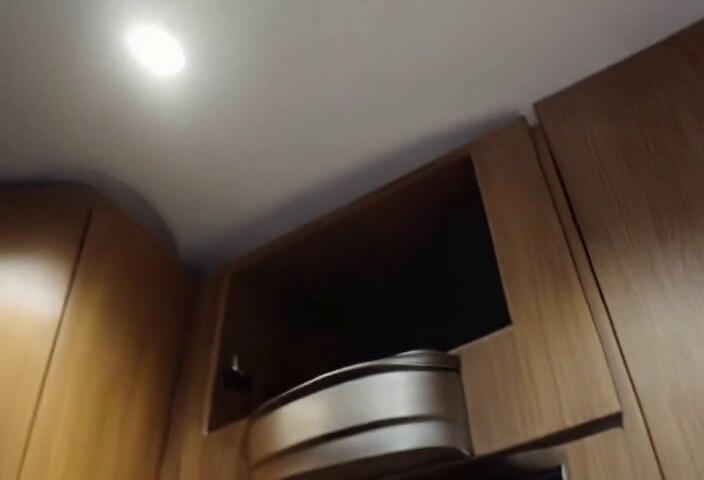} &
    \raisebox{3ex}{\rotatebox{90}{\tiny $Depth$}} &
    \includegraphics[width=0.16\linewidth]{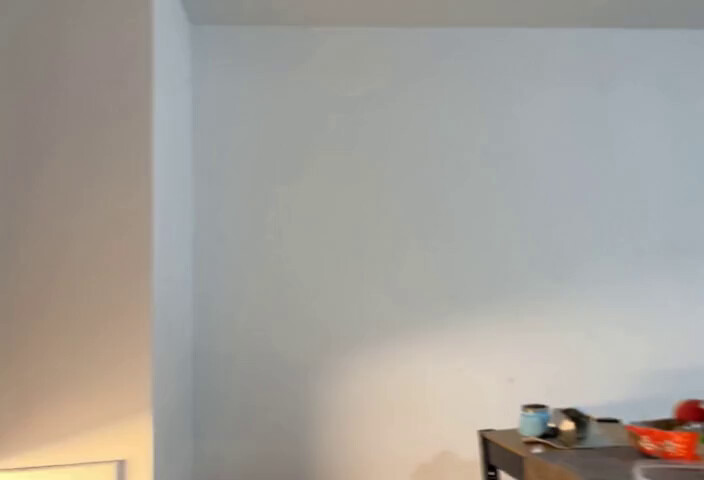} &
    \includegraphics[width=0.16\linewidth]{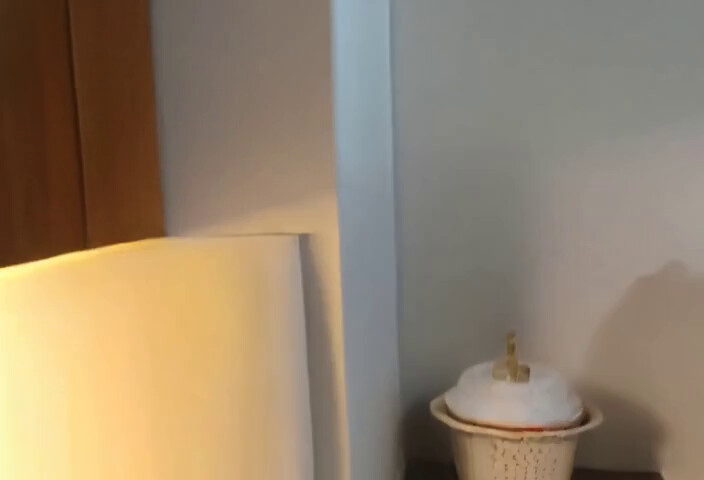} & 
    \includegraphics[width=0.16\linewidth]{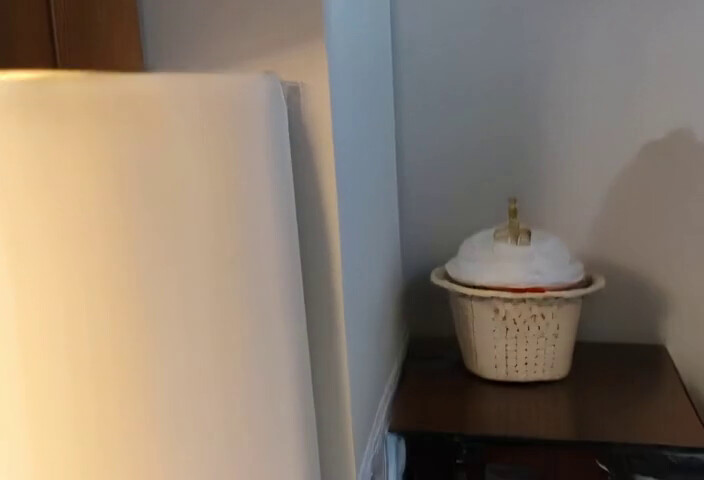}\\
  \end{tabular}
  \caption{Qualitative comparison on the ScanNet++ dataset between Control-DINO 2D and Wan2.2 under Canny-edge and depth conditioning.}
  \label{fig:visualisation_snpp_wan_ours}
\end{figure}

Furthermore, we present supplementary visualizations on the ScanNet++ dataset. \autoref{fig:cond_signal} shows the conditioning signals used by our approach at their native resolution of 60x90. \autoref{fig:visualisation_snpp_wan_ours} shows qualitative examples corresponding to the results reported in Table 4 in the main paper. We further demonstrate results on longer video sequences generated autoregressively with our model. \autoref{fig:temporal_stab} illustrates the temporal stability of DINO features by revisiting the same frames at different points in time. We additionally show the effect of different augmentations during conditioning signal generation (\autoref{fig:snpp_data_augment}).

\subsection{Dynamic Video} 
\begin{figure}[t]
    \centering
    \includegraphics[width=\linewidth, trim={0 0 0 3cm}]{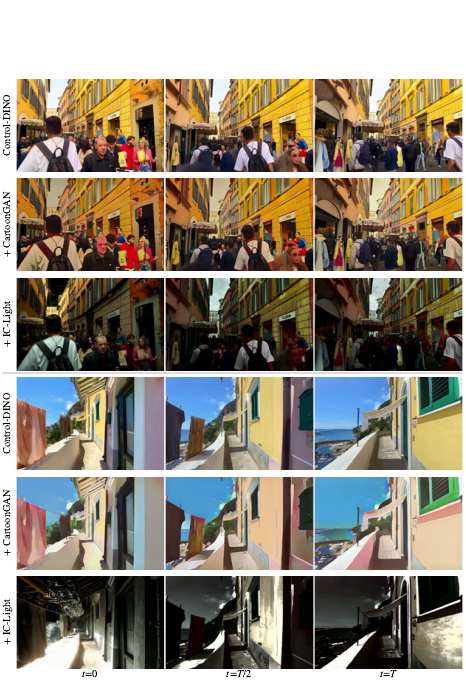}
    \caption{Control-DINO applied to a dynamic video with moving pedestrians, demonstrating compatibility with neural style transfer (CartoonGAN) and photometric augmentation (Noir).}
    \label{fig:dynamic_style}
    \vspace{-0.5ex}
\end{figure}

As shown in \autoref{fig:dynamic_style}, although the model was trained only with camera motion, it is able to generalize to scenes containing object motion.

\subsection{Causal Temporal Adapter}
To compress the per-frame DINO features from $T{=}49$ frames to match the VAE latent temporal dimension of $13$, we employ a causal temporal encoder that mirrors the design of the CogVideoX 3D Causal VAE~\cite{yang2025cogvideox}.
Like the VAE encoder, we use temporally causal 3D convolutions that pad only from past frames, ensuring each output depends solely on current and preceding inputs without access to future information. The encoder consists of two stages, each performing temporal downsampling with stride 2:
\begin{align}
    \mathbf{h}^{(1)} &= SiLU(\text{GN}(\text{CausalConv3D}(\mathbf{F}))), \quad &T: 49 \rightarrow 25 \\
    c &= SiLU(\text{GN}(\text{CausalConv3D}(\mathbf{h}^{(1)}))), \quad &T: 25 \rightarrow 13
\end{align}

Then $c$ is concatenated with the noisy state and passed to the 16 transformer blocks to obtain the residual for each of the blocks $h_l^c$.

\subsection{Video Diffusion Model}
\begin{figure}[t]
  \centering
  \setlength{\tabcolsep}{0pt}
  \renewcommand{\arraystretch}{0}
  \resizebox{0.9\linewidth}{!}{%
  \begin{tabular}{@{}>{\centering\arraybackslash}m{0.02\linewidth} @{}
                  >{\centering\arraybackslash}m{0.33\linewidth} @{}
                  >{\centering\arraybackslash}m{0.33\linewidth} @{}
                  >{\centering\arraybackslash}m{0.33\linewidth}@{}}
    & {\scriptsize \textbf{IC-Light Sunset}}
    & {\scriptsize \textbf{Fast Style Rain Princess}}
    & {\scriptsize \textbf{CartoonGAN Hayao}} \\[0pt]
    \rotatebox{90}{\scriptsize Wan 2.2}
      & \includegraphics[width=\linewidth]{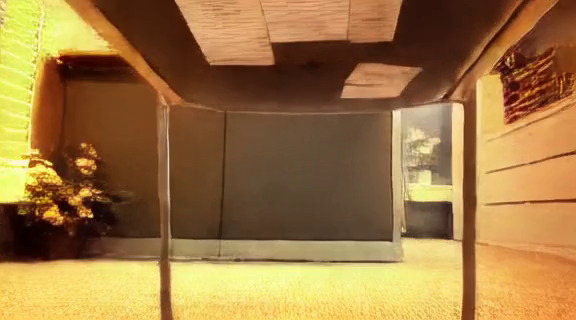}
      & \includegraphics[width=\linewidth]{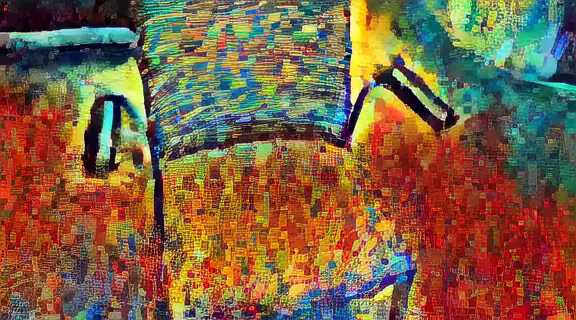}
      & \includegraphics[width=\linewidth]{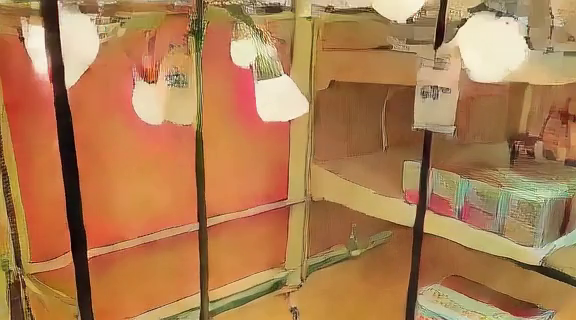} \\
    \rotatebox{90}{\scriptsize CogVideoX 1.1}
      & \includegraphics[width=\linewidth]{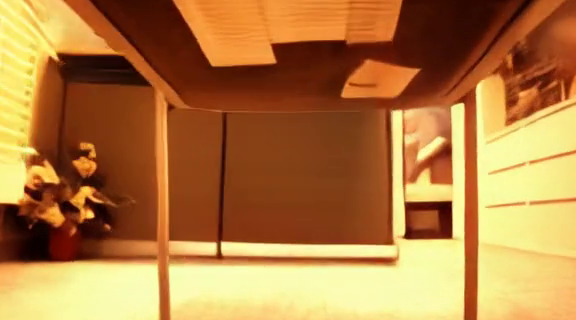}
      & \includegraphics[width=\linewidth]{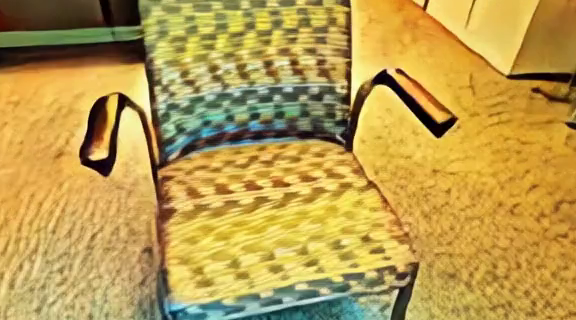}
      & \includegraphics[width=\linewidth]{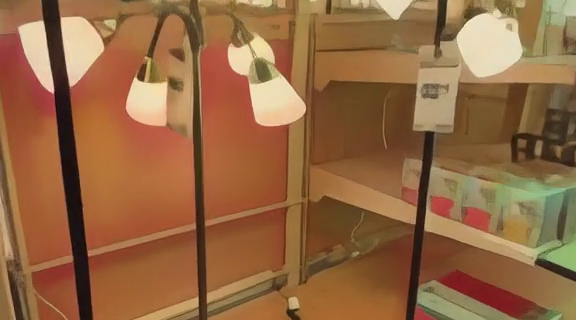} \\
  \end{tabular}%
  }
      \vspace{-0.2cm}
  \caption{Wan2.2 with Control-DINO ControlNet style transfer.}
  \label{fig:wan_vs_cogvideox_styles}
    \vspace{-0.6cm}
\end{figure}

We have additionally trained the same Control-DINO model on top of Wan2.2-I2V-5B (L=14) and we have found the training strategy and feature conditioning to work robustly and obtain results visually on par with our CogVideoX based backbone \autoref{fig:wan_vs_cogvideox_styles}.

\subsection{Prompt Conditioning}
For the transfer experiments, during both training and inference, we append style-specific keywords to the text prompt before passing it to the text encoder and diffusion model. For example we add "paprika", "drawing contour" or "soft light", this has a much weaker conditioning effect compared to the first frame image, but it helps to align the distributions.


\subsection{Long Video Sequences}

\begin{figure}[htbp]
  \centering
  \setlength{\tabcolsep}{0.3pt}
  \renewcommand{\arraystretch}{1.0}
  \scriptsize
  \begin{tabular}{@{} c c c c c c c c c c @{}}
    t=1 & t=25 & t=49 & t=74 & t=98 & t=123 & t=147 & t=172 & t=196 \\
      \includegraphics[width=0.11\linewidth]{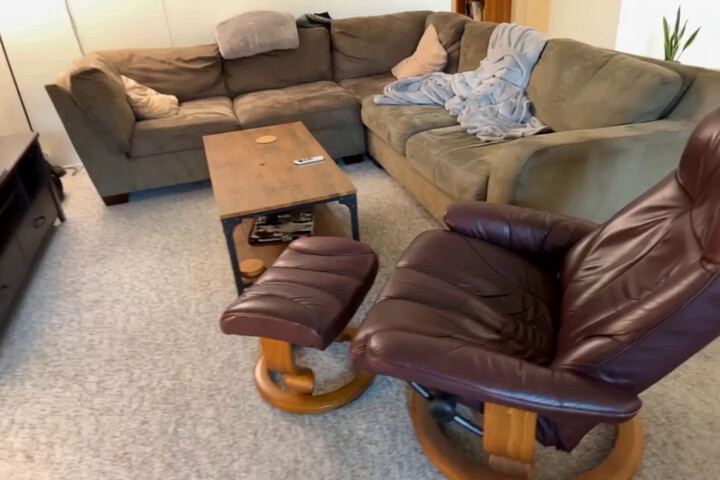} &
      \includegraphics[width=0.11\linewidth]{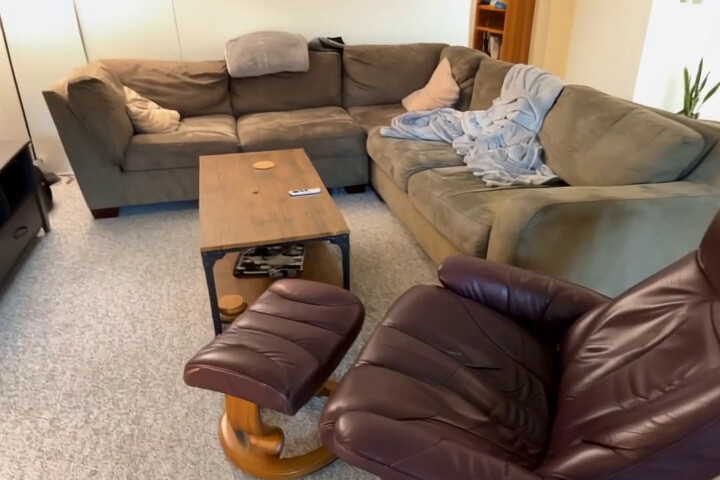} &
      \includegraphics[width=0.11\linewidth]{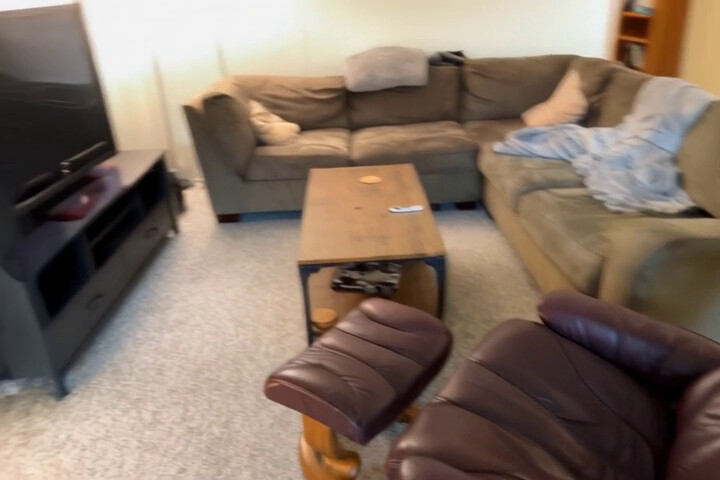} &
      \includegraphics[width=0.11\linewidth]{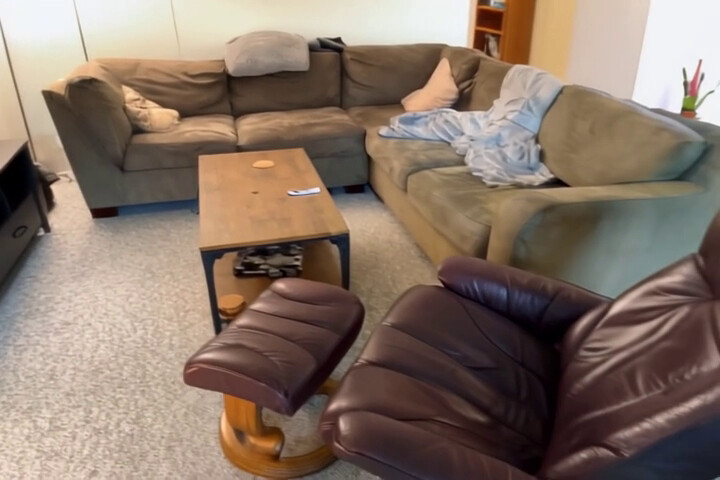} &
      \includegraphics[width=0.11\linewidth]{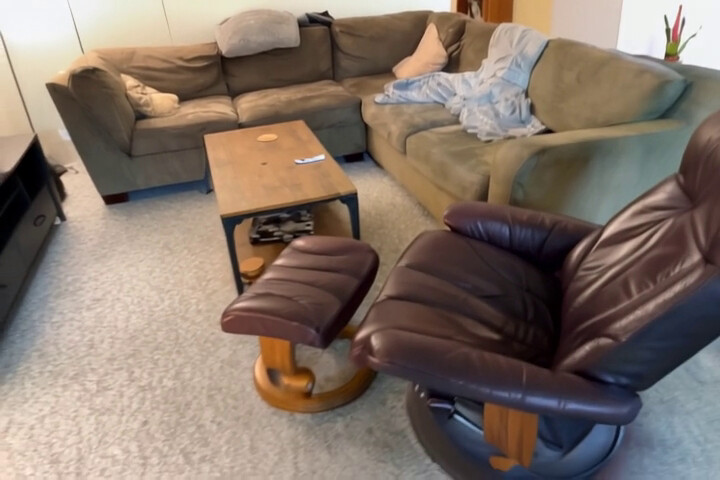} &
      \includegraphics[width=0.11\linewidth]{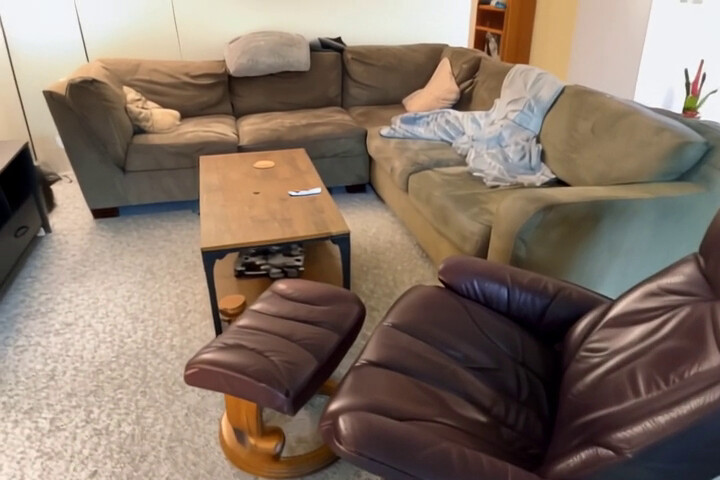} &
      \includegraphics[width=0.11\linewidth]{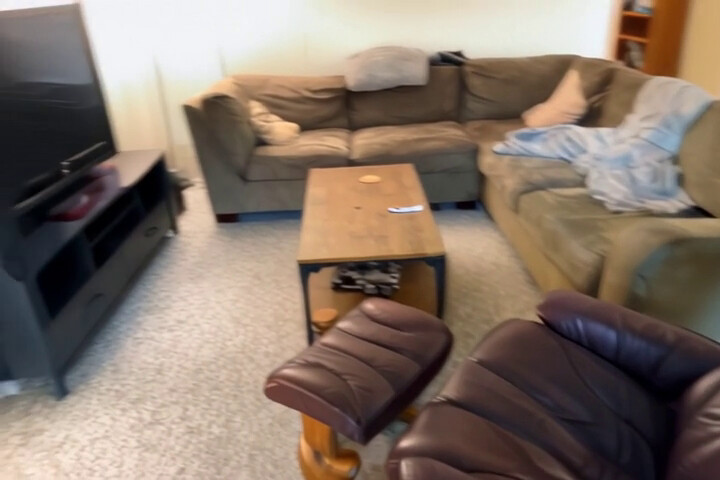} &
      \includegraphics[width=0.11\linewidth]{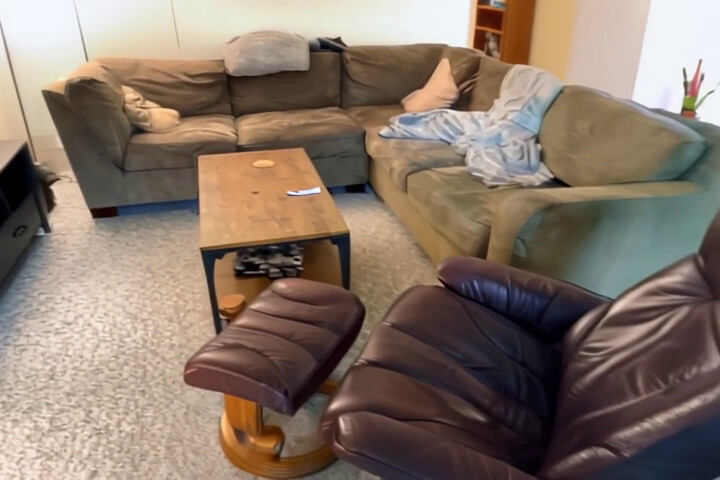} &
      \includegraphics[width=0.11\linewidth]{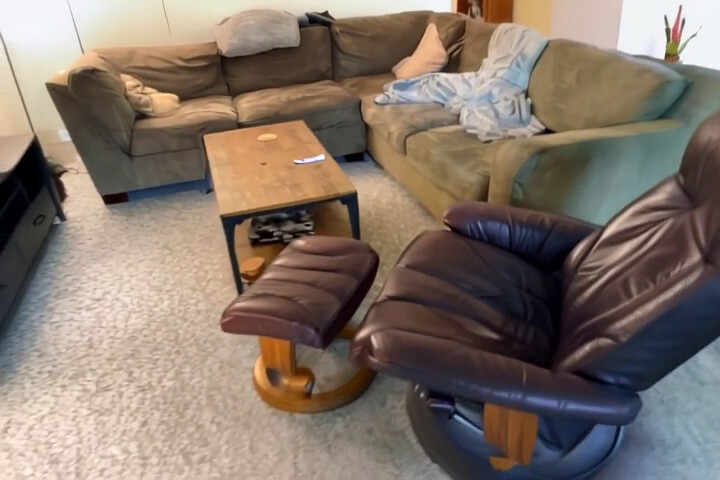} \\
      \includegraphics[width=0.11\linewidth]{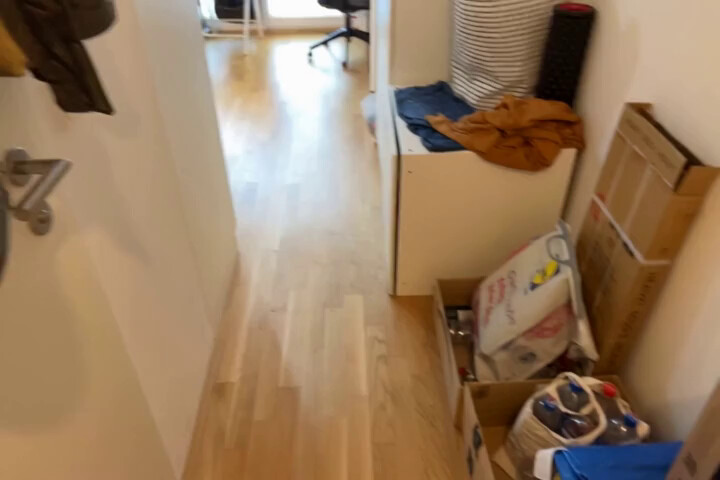} &
      \includegraphics[width=0.11\linewidth]{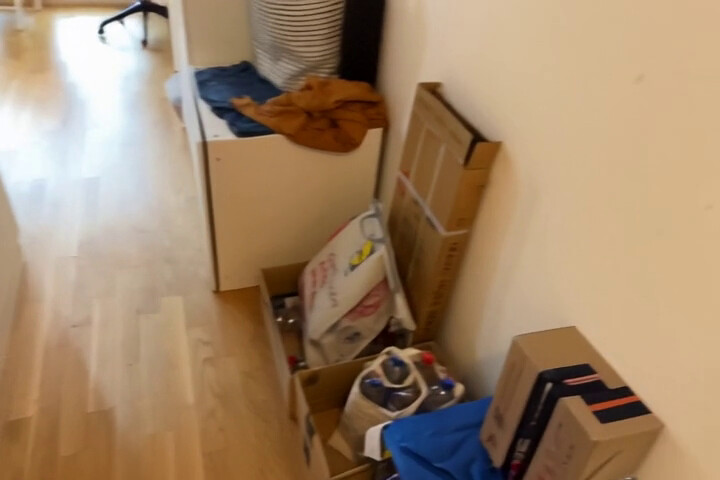} &
      \includegraphics[width=0.11\linewidth]{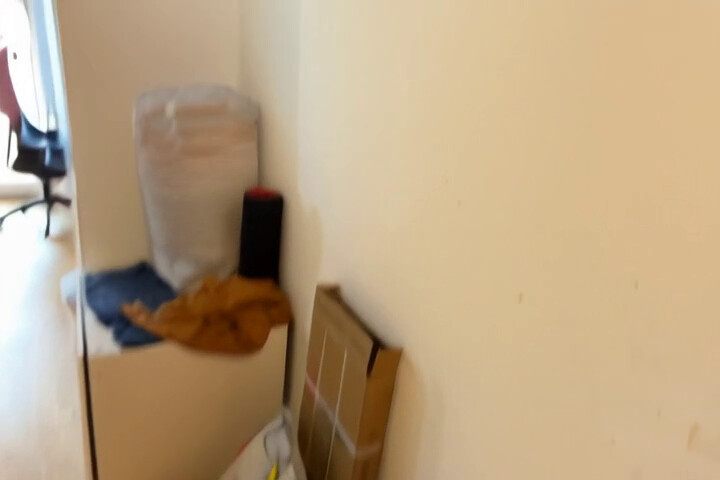} &
      \includegraphics[width=0.11\linewidth]{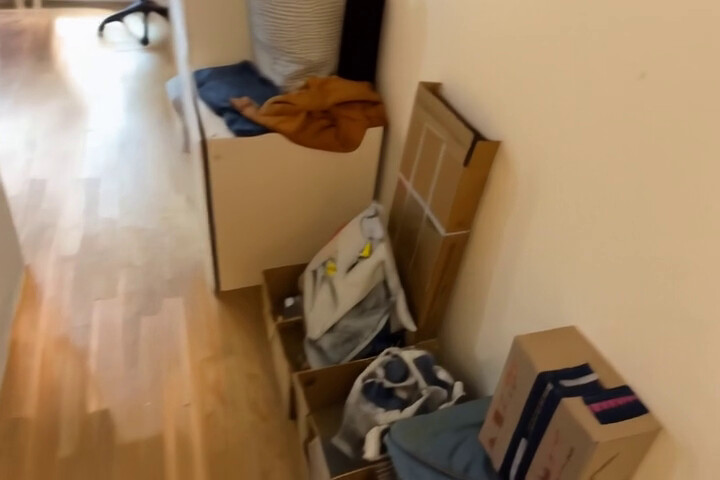} &
      \includegraphics[width=0.11\linewidth]{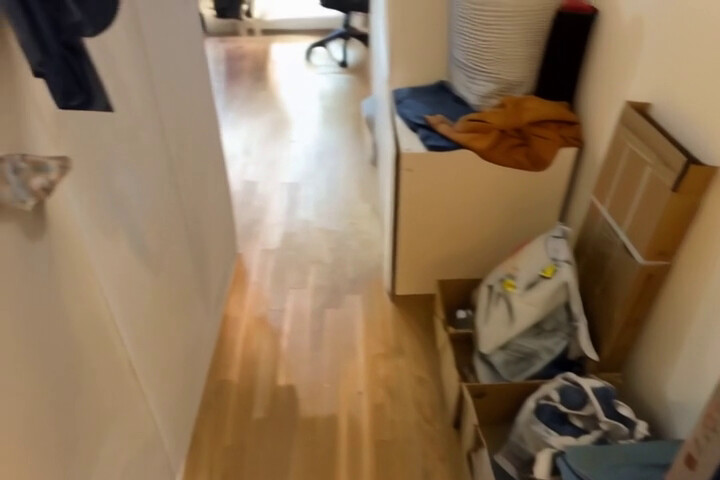} &
      \includegraphics[width=0.11\linewidth]{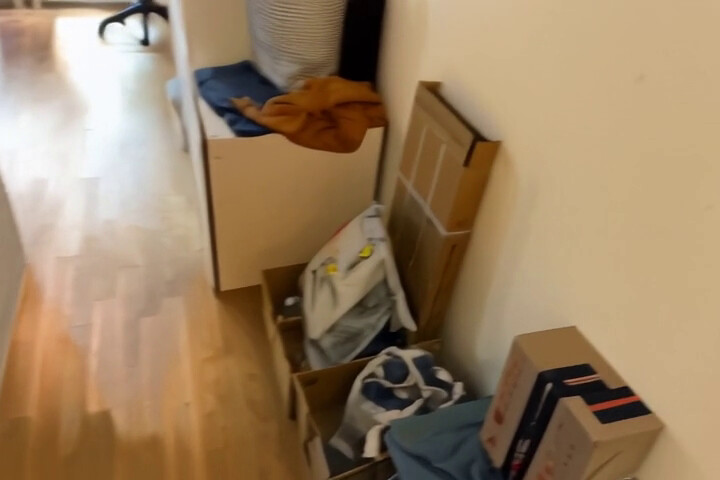} &
      \includegraphics[width=0.11\linewidth]{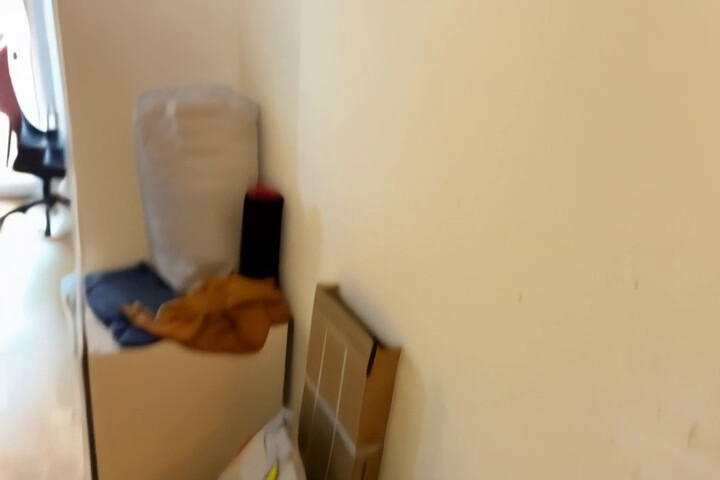} &
      \includegraphics[width=0.11\linewidth]{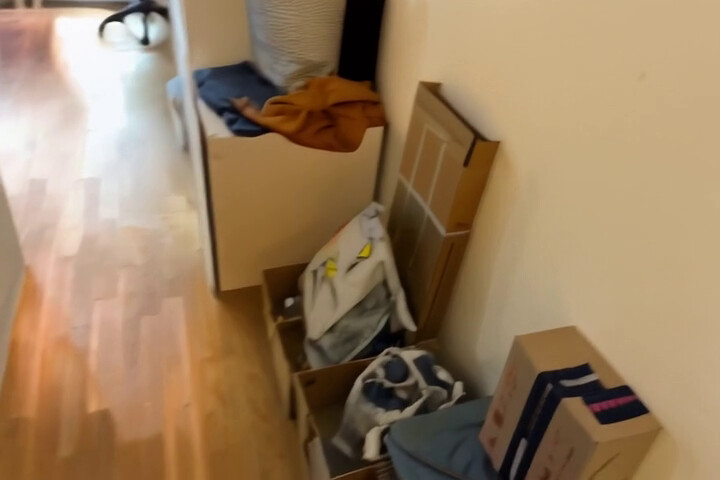} &
      \includegraphics[width=0.11\linewidth]{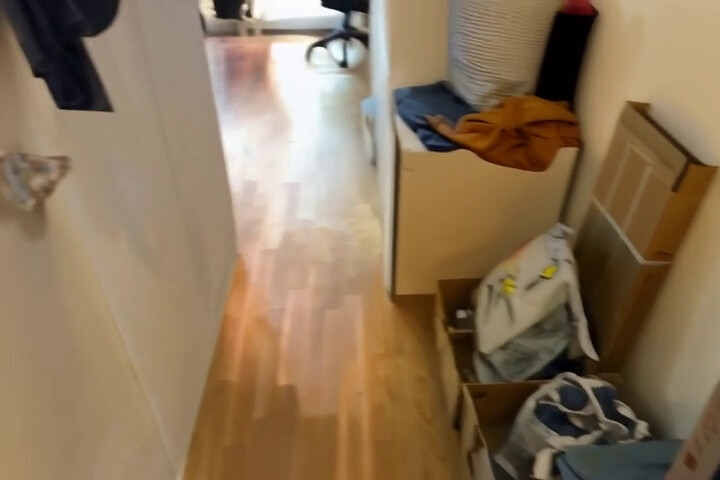} \\
      \includegraphics[width=0.11\linewidth]{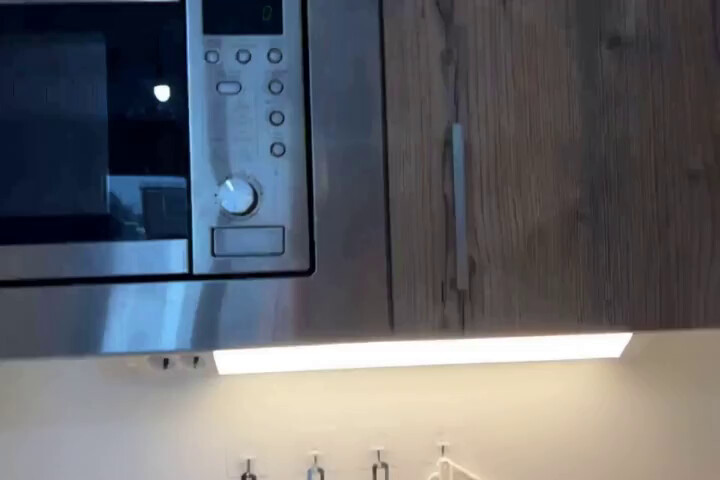} &
      \includegraphics[width=0.11\linewidth]{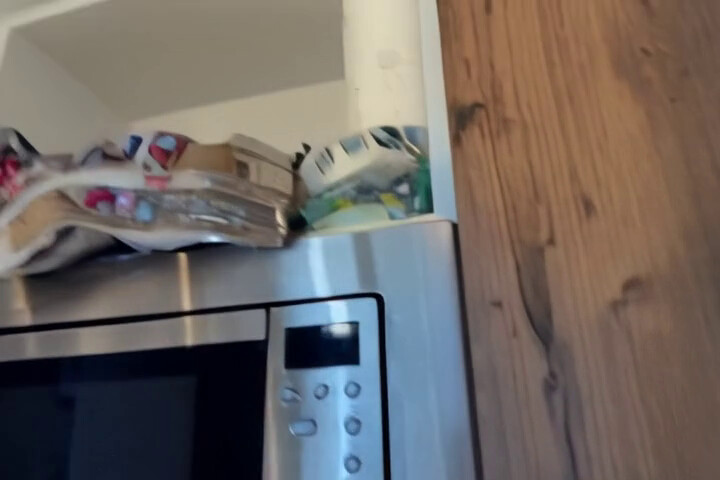} &
      \includegraphics[width=0.11\linewidth]{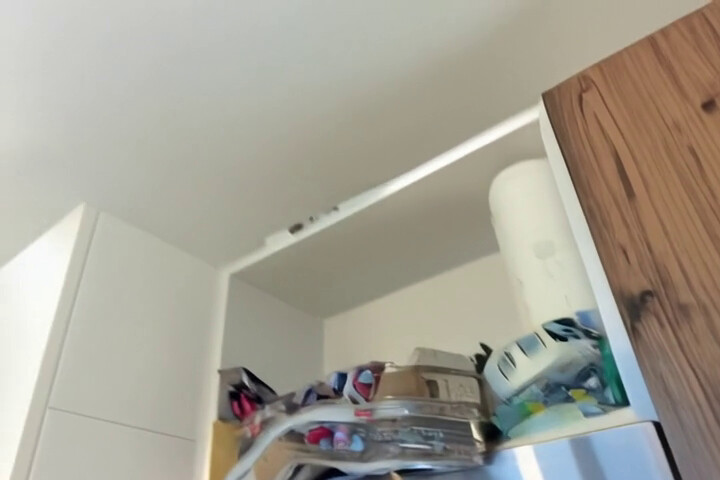} &
      \includegraphics[width=0.11\linewidth]{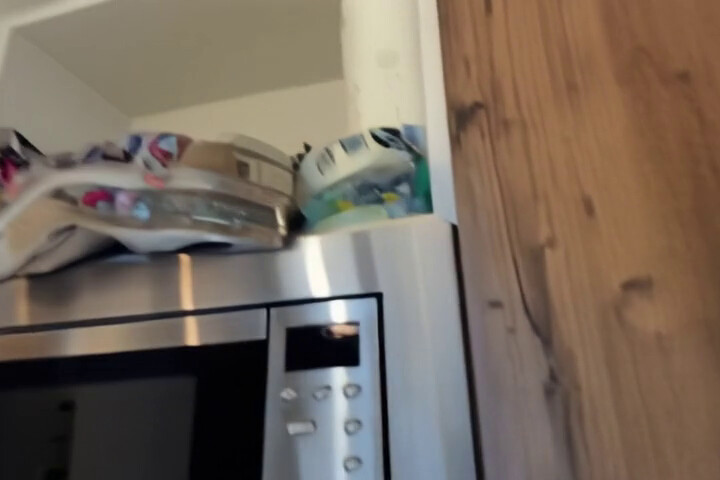} &
      \includegraphics[width=0.11\linewidth]{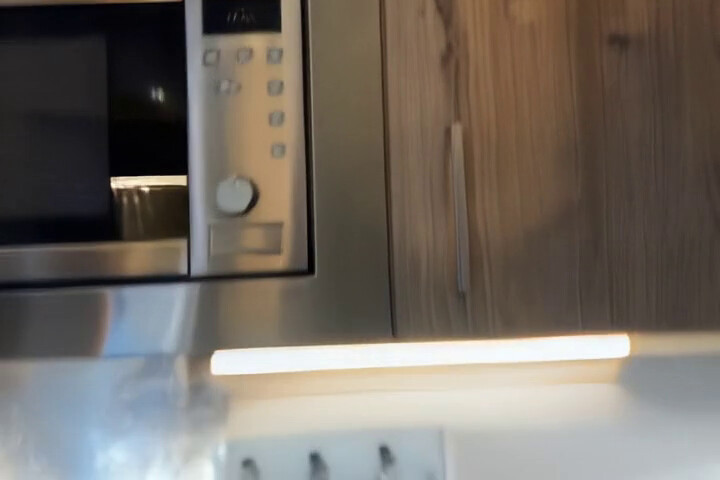} &
      \includegraphics[width=0.11\linewidth]{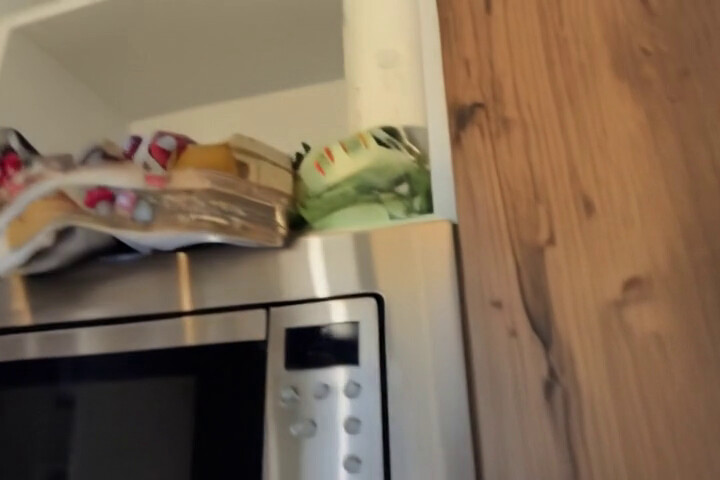} &
      \includegraphics[width=0.11\linewidth]{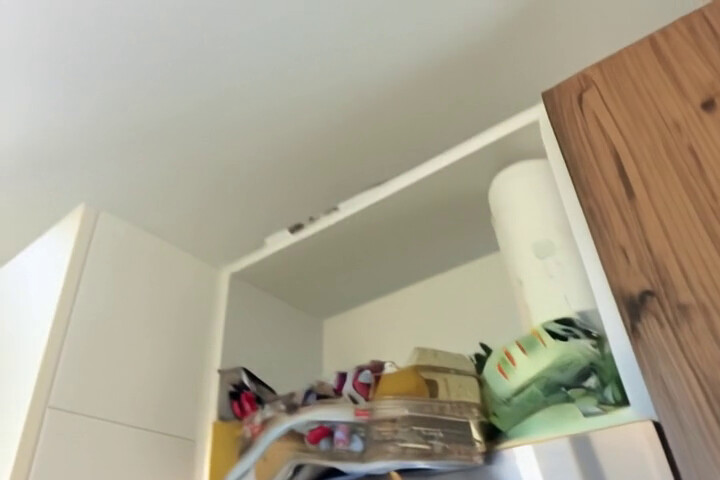} &
      \includegraphics[width=0.11\linewidth]{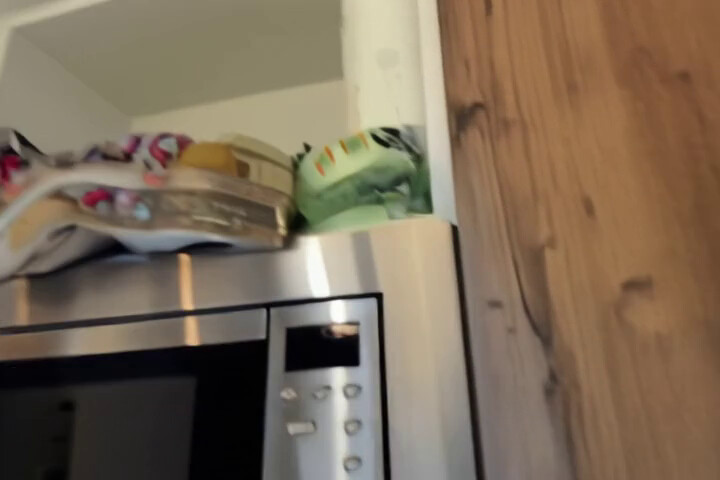} &
      \includegraphics[width=0.11\linewidth]{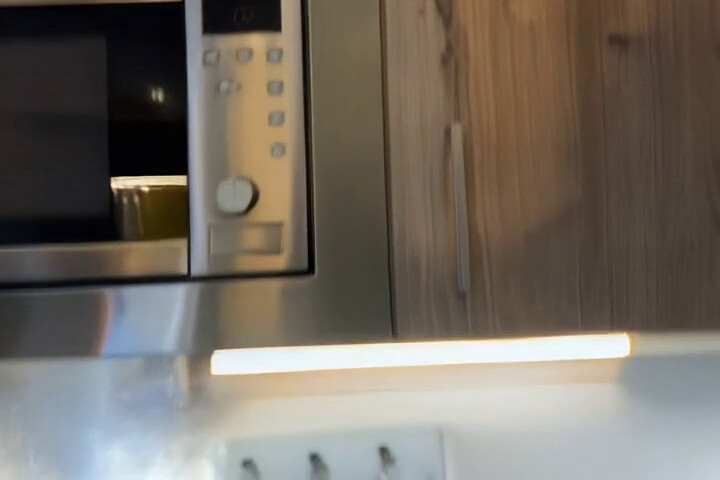} \\
      \includegraphics[width=0.11\linewidth]{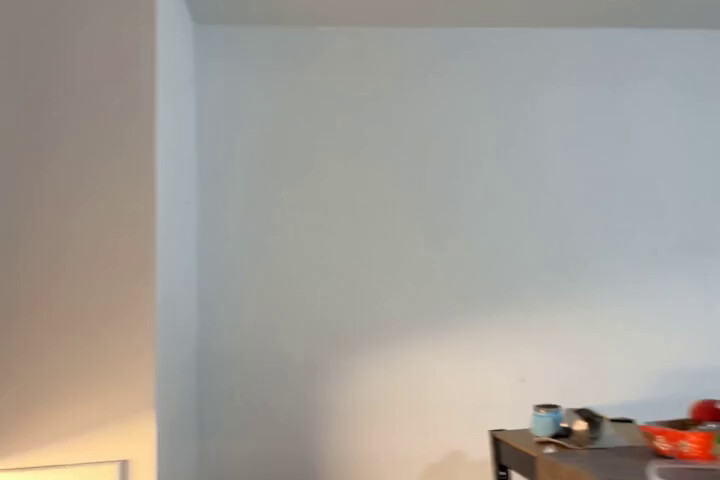} &
      \includegraphics[width=0.11\linewidth]{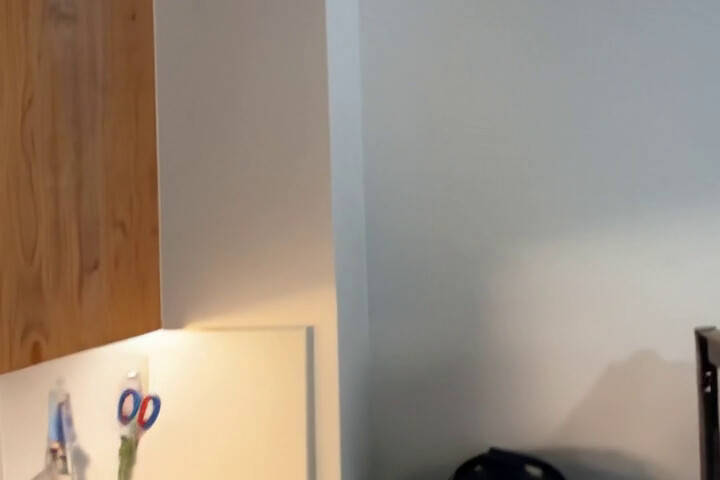} &
      \includegraphics[width=0.11\linewidth]{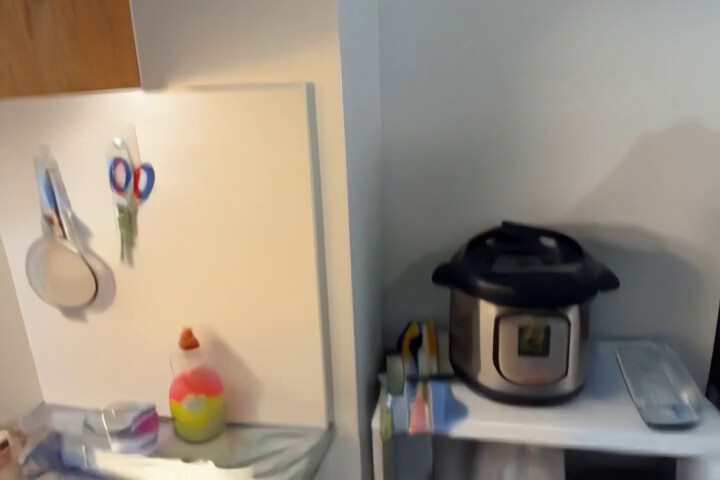} &
      \includegraphics[width=0.11\linewidth]{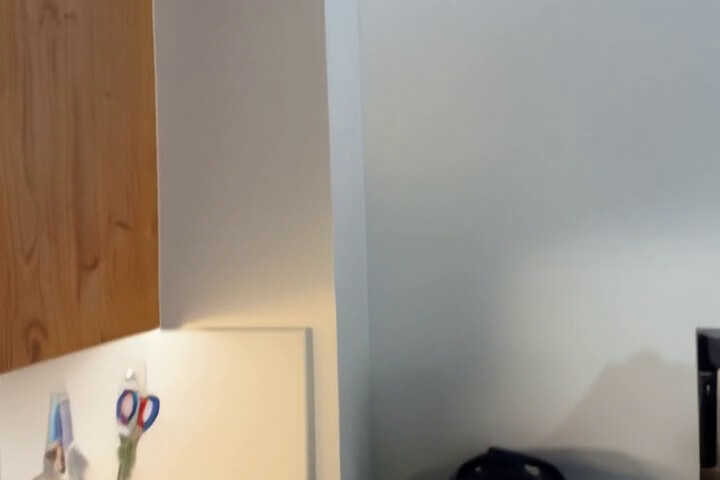} &
      \includegraphics[width=0.11\linewidth]{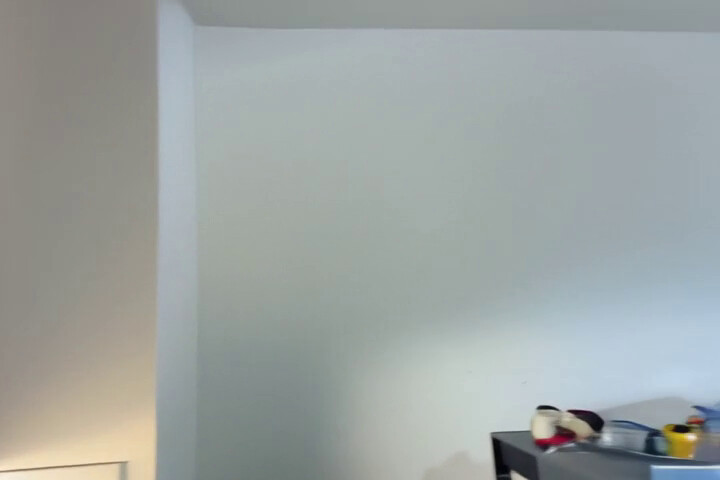} &
      \includegraphics[width=0.11\linewidth]{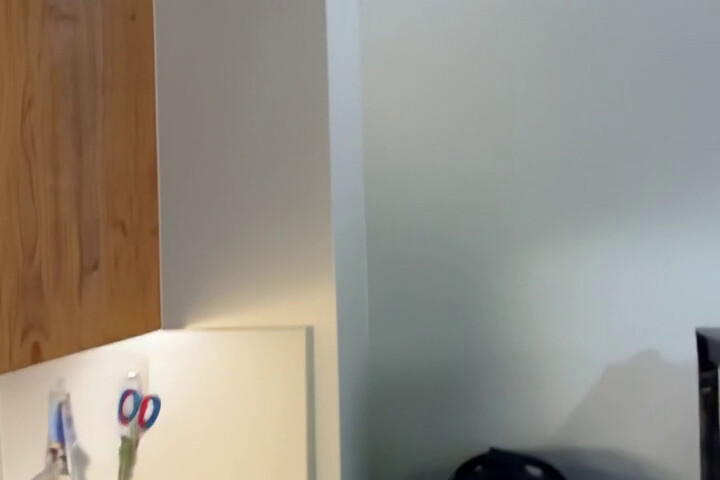} &
      \includegraphics[width=0.11\linewidth]{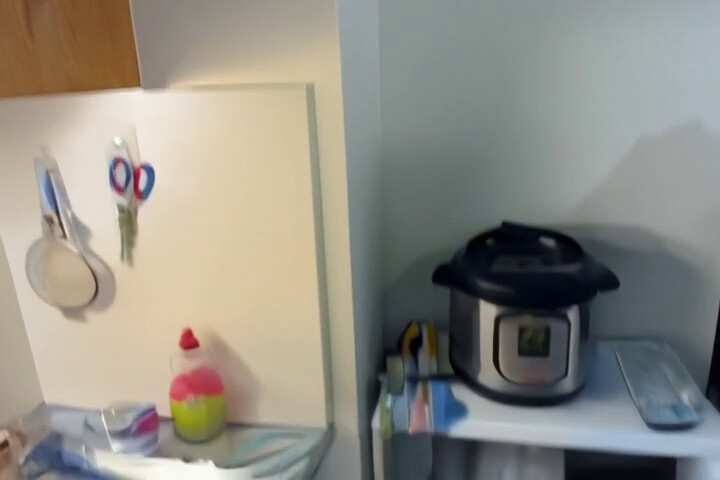} &
      \includegraphics[width=0.11\linewidth]{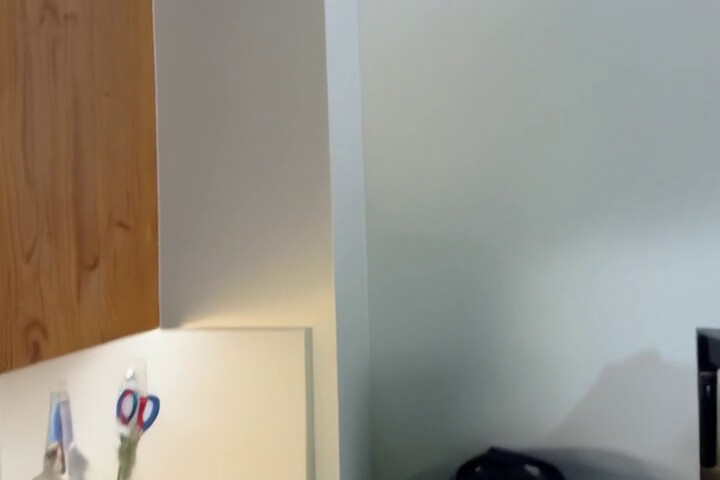} &
      \includegraphics[width=0.11\linewidth]{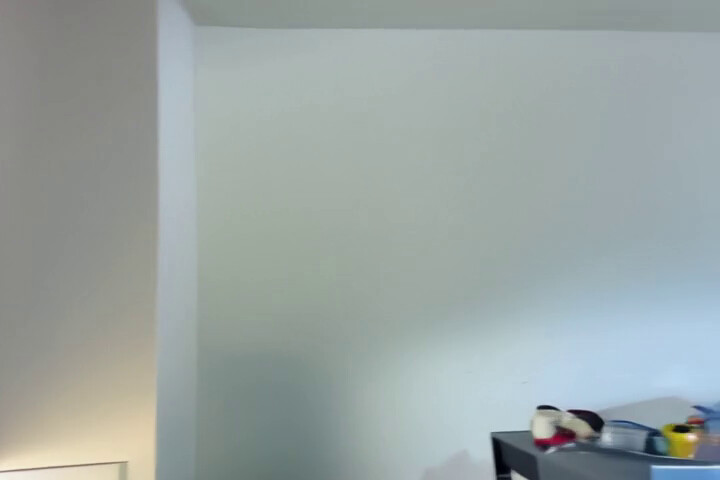} \\
  \end{tabular}
  \caption{Temporal stability of the generated content when looping the conditioning signal while running our method autoregressively. Frames with identical camera poses occur at t=1/98/196, t=25/74/123/172, and t=49/147, allowing comparison of scene consistency over long time horizons.}
  \label{fig:temporal_stab}
\end{figure}

Our model currently supports up to 49 frames per inference. Nevertheless, arbitrarily long video sequences can be generated by running the model autoregressively. In particular, inference is performed on blocks of 49 frames, where the conditioning frame for each subsequent block is set to the last output frame of the previously generated block. We demonstrate this procedure in \autoref{fig:temporal_stab}. 

Empirically, we observe that long video sequences can be generated without any noticeable degradation in quality. Camera poses remain highly consistent across time, with only minor deviations relative to the initial conditioning frame. This indicates that our model maintains temporal coherence and stability even over extended autoregressive rollouts, highlighting the robustness of both the generated content and the DINO feature representations.

\subsection{Conditioning Signal Augmentation}
\begin{figure}[htbp]
  \centering
  \setlength{\tabcolsep}{1pt}
  \renewcommand{\arraystretch}{1.0}
  \scriptsize
  \begin{tabular}{@{}c c c c @{}}
      \textbf{Voxel size 0.05} & \textbf{Voxel size 0.03} & \textbf{Point cloud} & \textbf{Features 30x45} \\
    \includegraphics[width=0.24\linewidth]{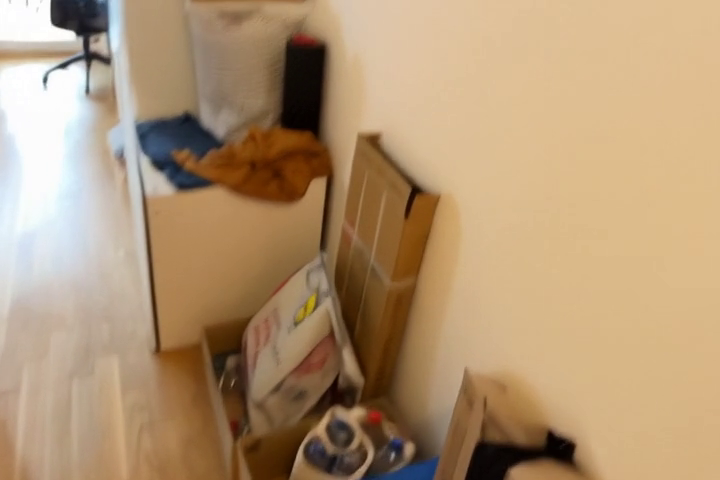} &
    \includegraphics[width=0.24\linewidth]{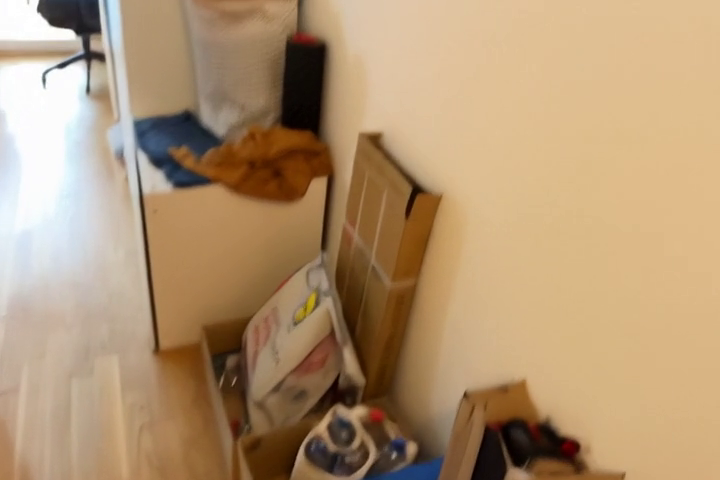} &
     \includegraphics[width=0.24\linewidth]{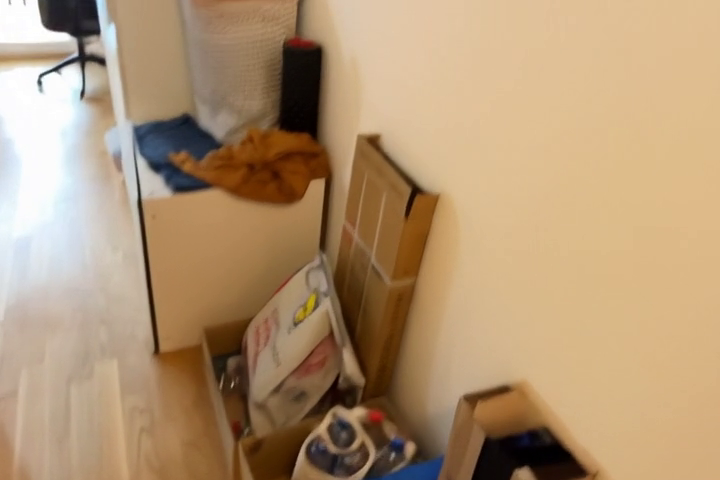} &
    \includegraphics[width=0.24\linewidth]{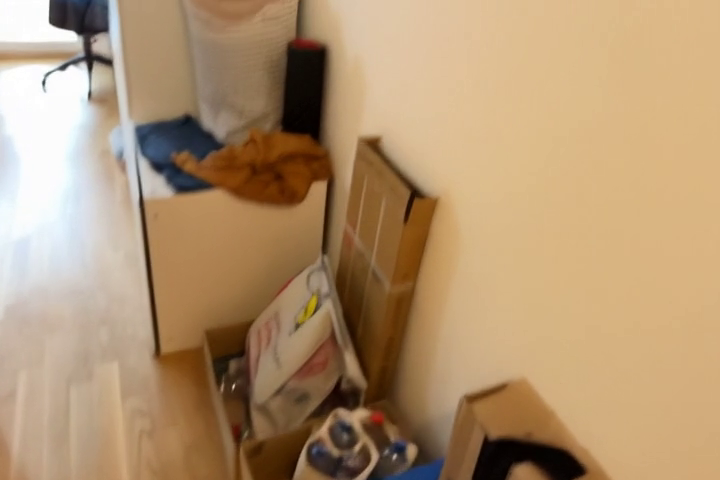} 
 \\
    \includegraphics[width=0.24\linewidth]{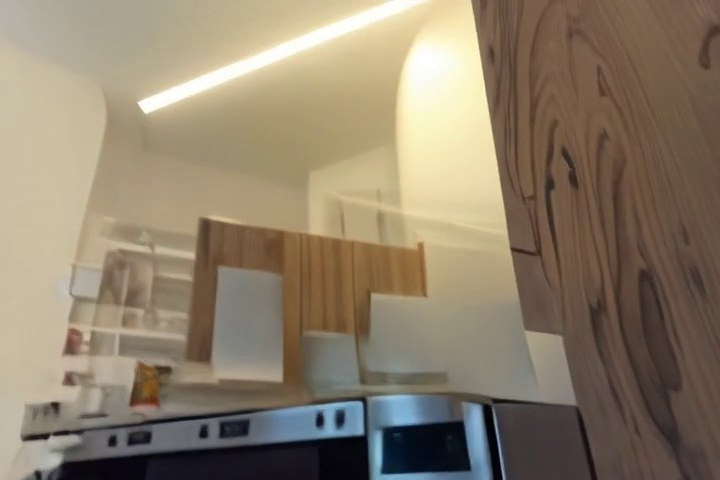} &
    \includegraphics[width=0.24\linewidth]{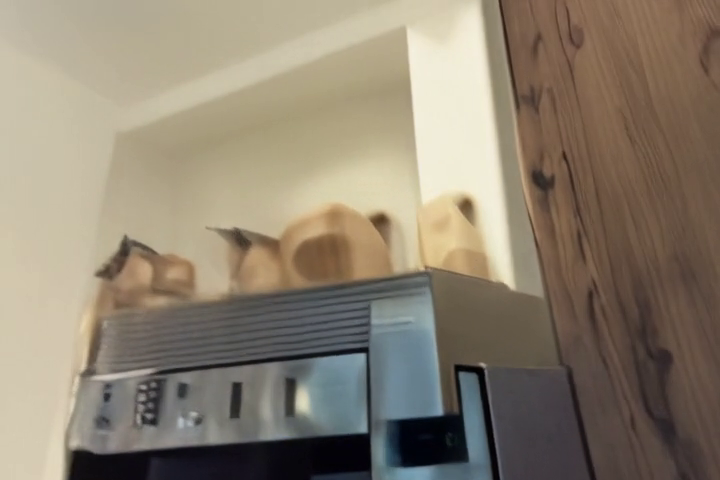} &
     \includegraphics[width=0.24\linewidth]{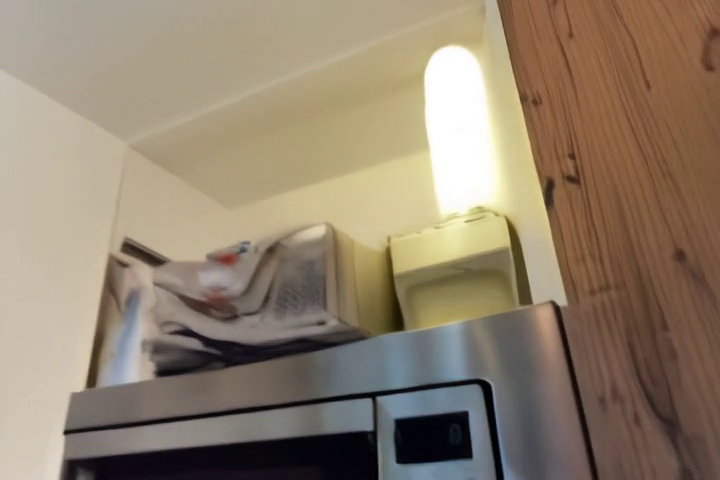} &
    \includegraphics[width=0.24\linewidth]{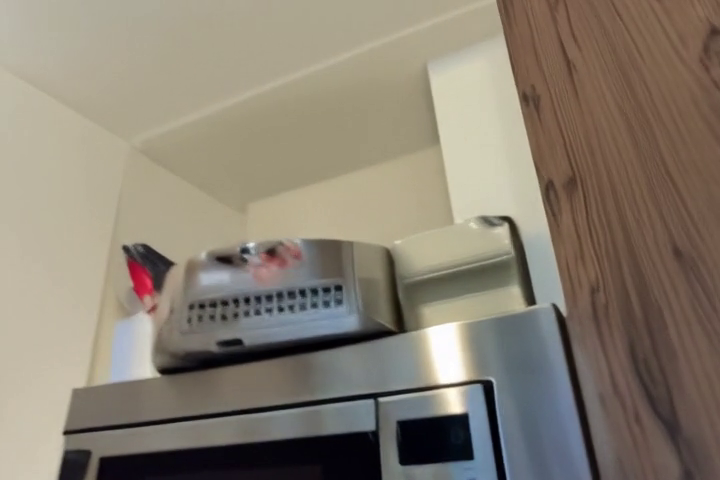} 
 \\
  \end{tabular}
\caption{Visual examples of augmentations to the conditioning signal: different voxel sizes(default 0.02), point clouds in place of voxels, and downscaled feature maps.}
\label{fig:snpp_data_augment}
\end{figure}

In \autoref{fig:snpp_data_augment}, we provide further insight into the robustness of our control network under changes to the conditioning signal. 
To do this, we modify the 3D structure in various ways, such as changing the original voxel size, performing point cloud rendering instead of using voxels, and downscaling the feature maps by reducing their resolution by a factor of 2. 
As can be seen, our model is not entirely unaffected by these changes, but it handles them reasonably well depending on the magnitude of the modifications.

\subsection{Feature Extraction Robustness}
In the 2D video transfer setting, our method relies on frame-by-frame feature extraction from the input video. Consequently, the quality of the extracted features can be affected by artifacts in the source video, including motion blur, occlusions, temporal instability, and other acquisition errors. Such degradations may not only impair the quality of individual feature representations but also introduce larger temporal variations between features extracted from consecutive frames.

\begin{figure}[htbp]
  \centering
  \setlength{\tabcolsep}{0.0pt}
  \renewcommand{\arraystretch}{0.0}
  \scriptsize
  \begin{tabular}{@{}c c c c c c@{}}
     & \textbf{\tiny \textbf{$t{=}36$}} & \textbf{\tiny $t{=}37$} & \textbf{\tiny $t{=}38$} & \textbf{\tiny \textbf{$t{=}39$}} & \textbf{\tiny $t{=}40$} \\
    \raisebox{4ex}{\rotatebox{90}{\tiny GT}} &
    \includegraphics[width=0.166\linewidth]{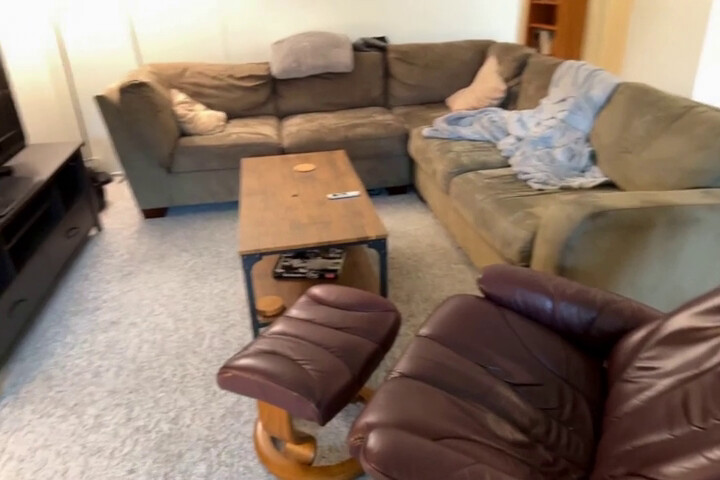} &
    \includegraphics[width=0.166\linewidth]{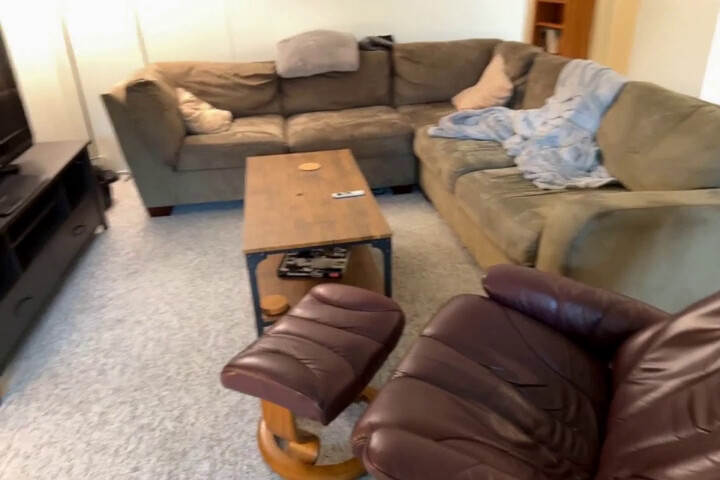} &
    \includegraphics[width=0.166\linewidth]{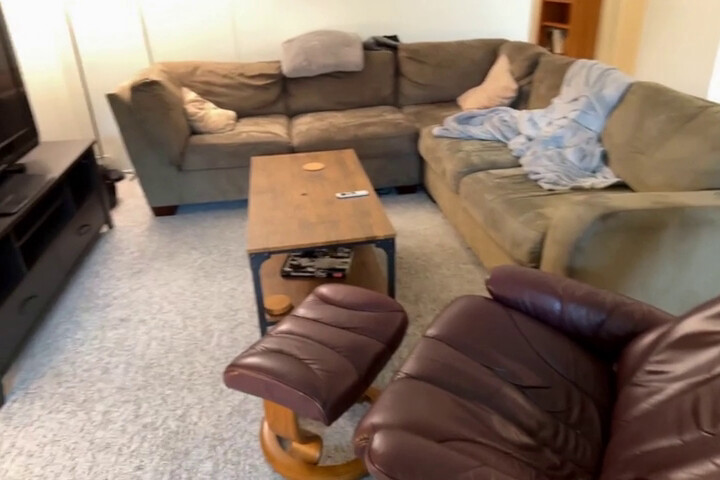} &
    \includegraphics[width=0.166\linewidth]{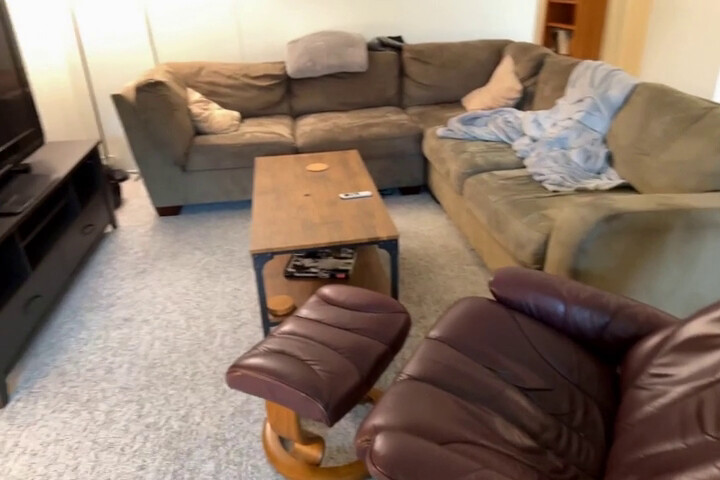} &
    \includegraphics[width=0.166\linewidth]{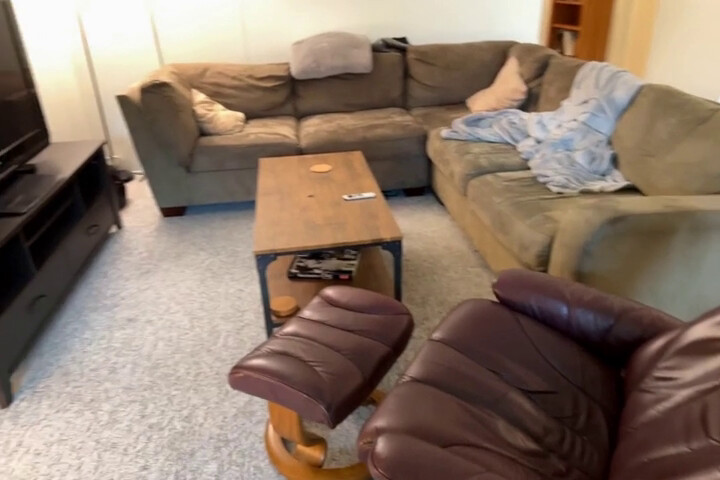} \\
    \raisebox{0ex}{\rotatebox{90}{ \shortstack{\tiny MB Input \\ (weak)}}} &
    \includegraphics[width=0.166\linewidth]{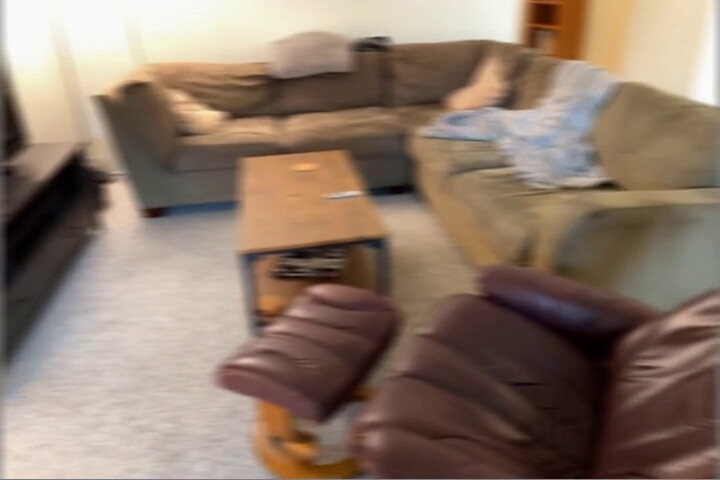} &
    \includegraphics[width=0.166\linewidth]{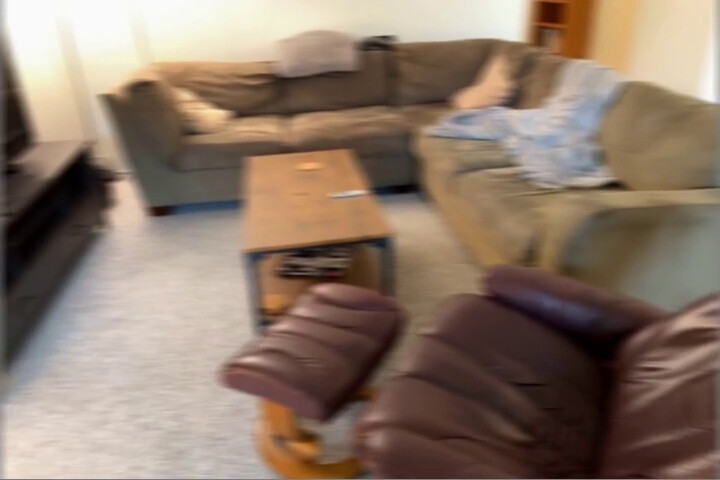} &
    \includegraphics[width=0.166\linewidth]{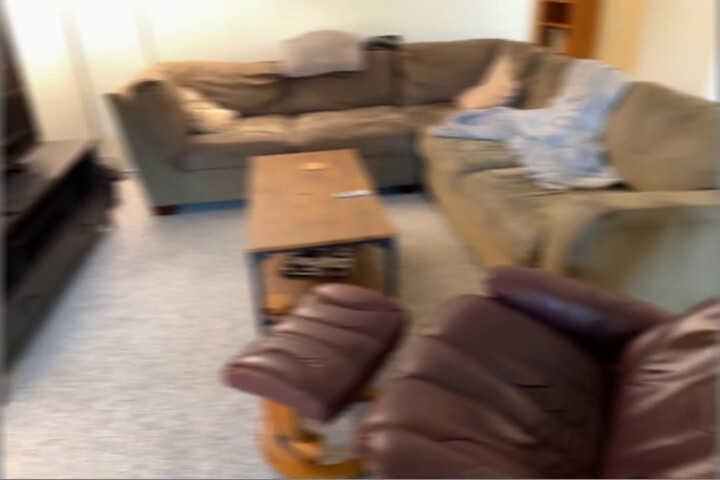} &
    \includegraphics[width=0.166\linewidth]{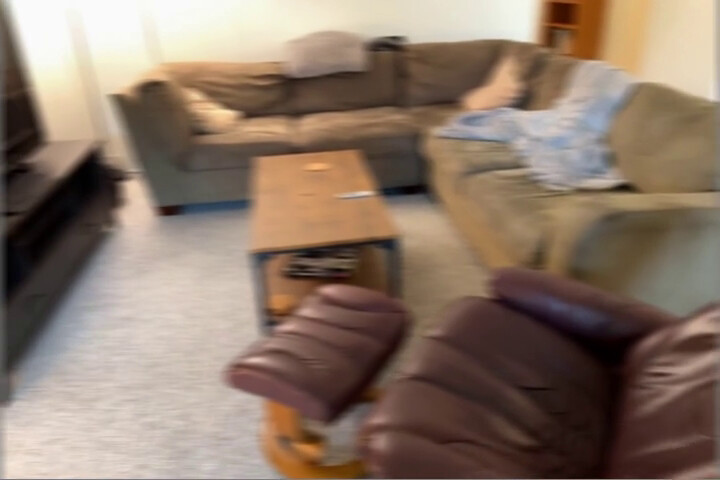} &
    \includegraphics[width=0.166\linewidth]{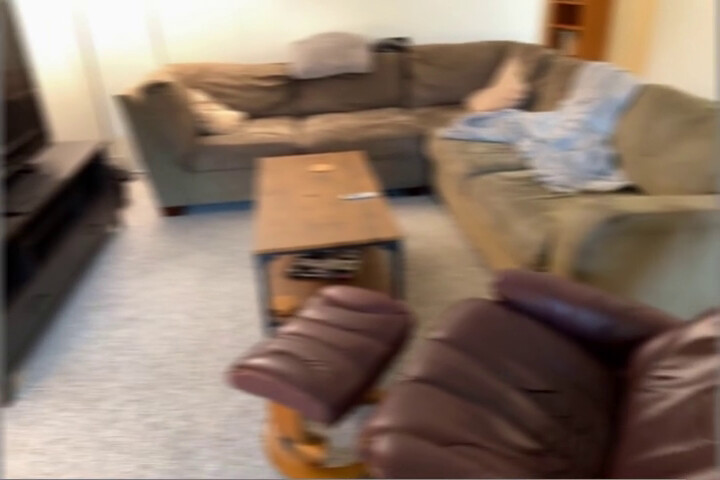} \\
    \raisebox{0ex}{\rotatebox{90}{ \shortstack{\tiny MB Pred \\ (weak)}}} &
    \includegraphics[width=0.166\linewidth]{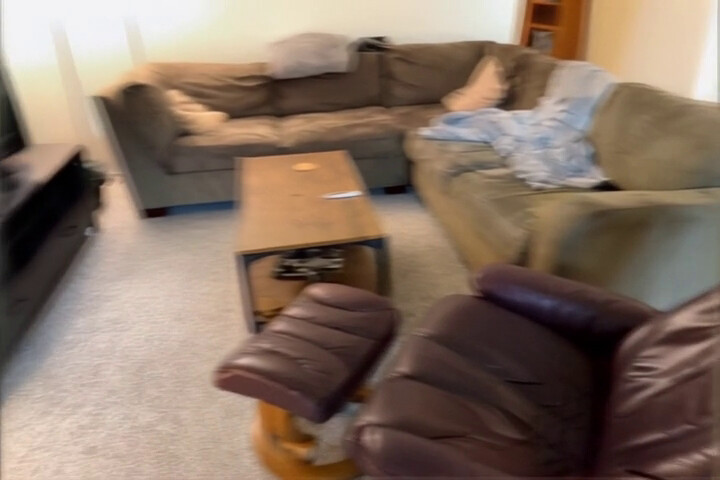} &
    \includegraphics[width=0.166\linewidth]{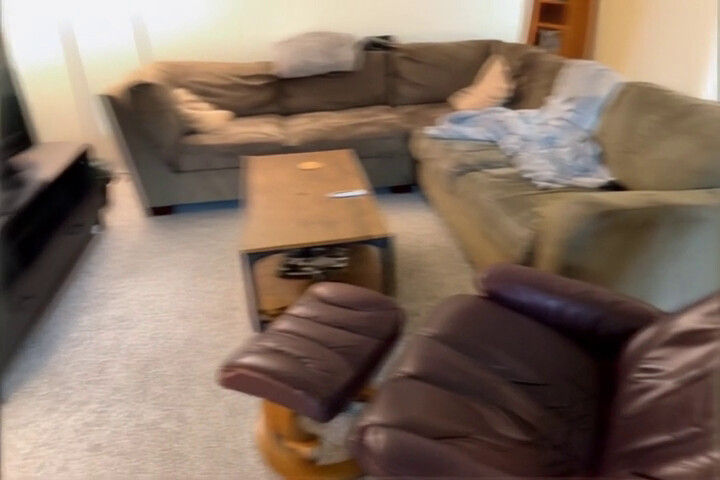} &
    \includegraphics[width=0.166\linewidth]{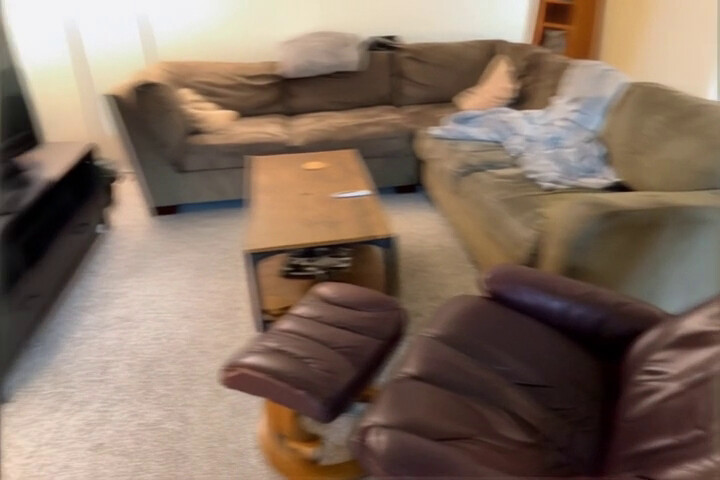} &
    \includegraphics[width=0.166\linewidth]{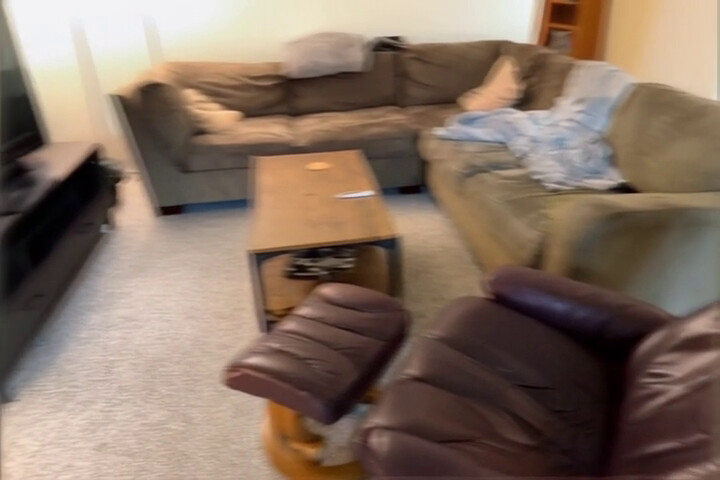} &
    \includegraphics[width=0.166\linewidth]{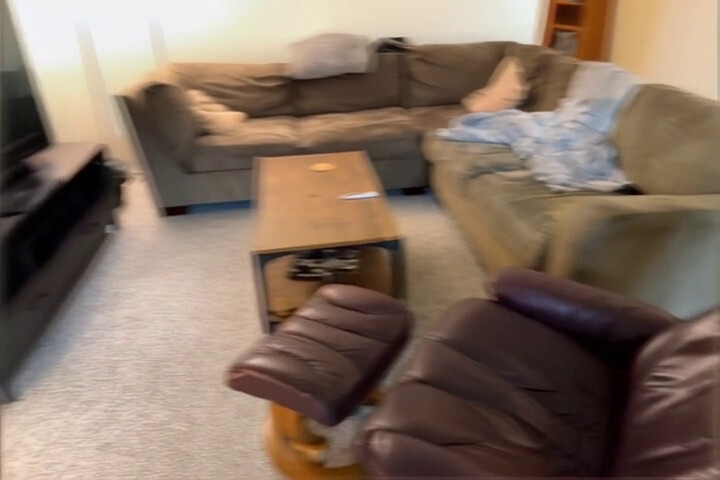} \\
    \raisebox{0ex}{\rotatebox{90}{ \shortstack{\tiny MB Input \\ (strong)}}} &
    \includegraphics[width=0.166\linewidth]{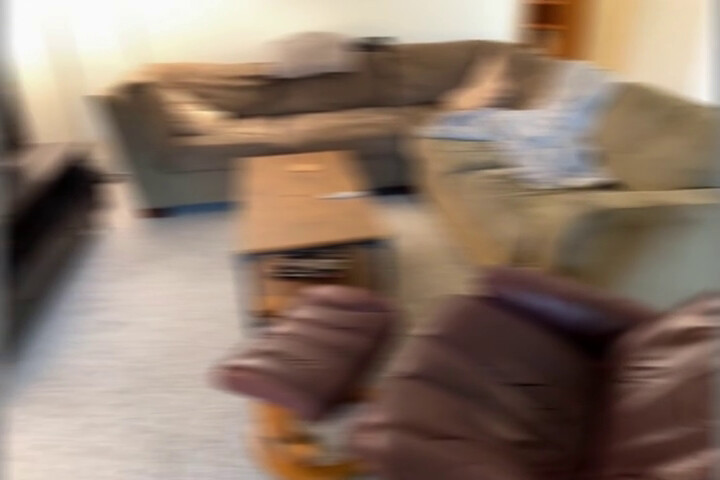} &
    \includegraphics[width=0.166\linewidth]{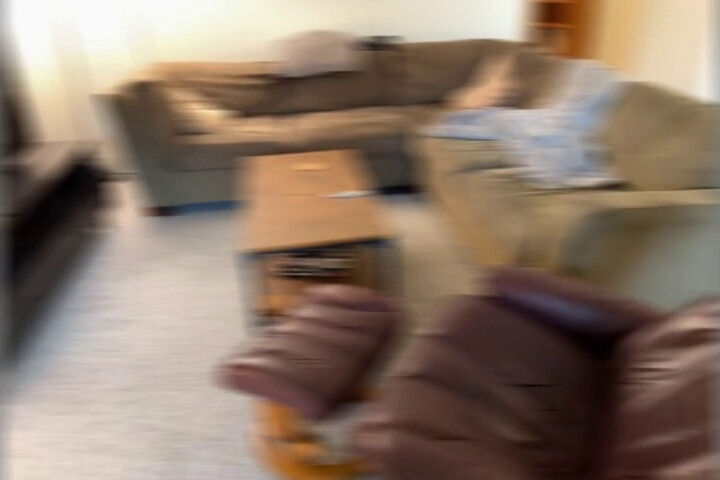} &
    \includegraphics[width=0.166\linewidth]{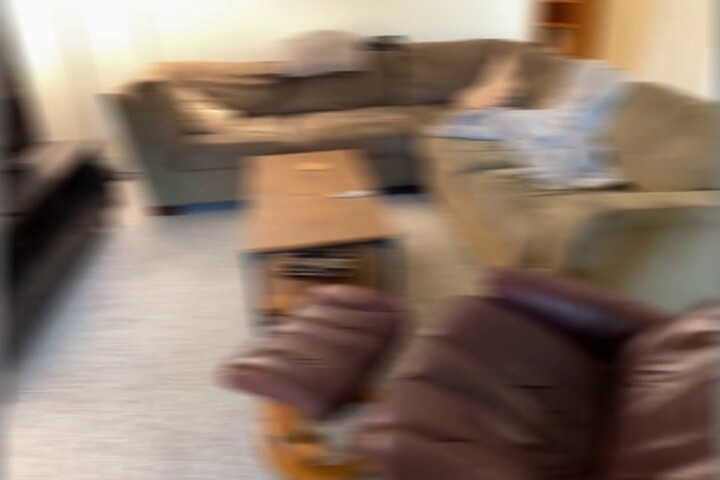} &
    \includegraphics[width=0.166\linewidth]{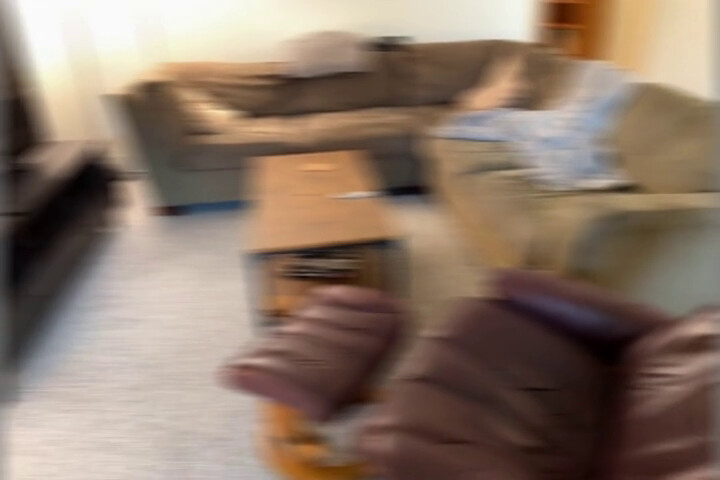} &
    \includegraphics[width=0.166\linewidth]{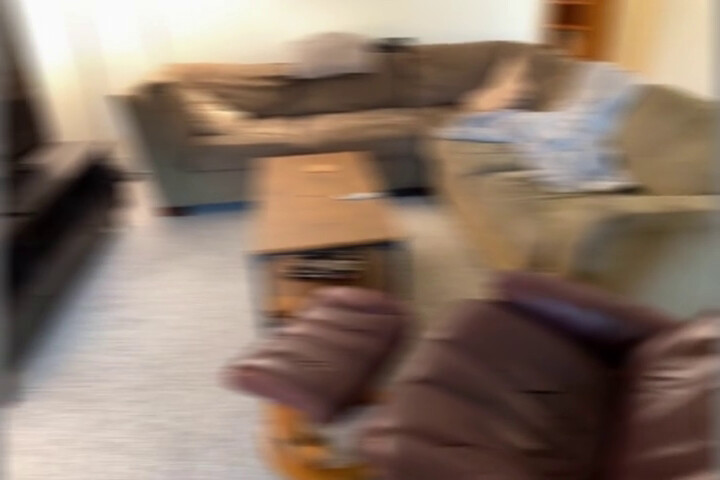} \\
    \raisebox{0ex}{\rotatebox{90}{ \shortstack{\tiny MB Pred \\ (strong)}}} &
    \includegraphics[width=0.166\linewidth]{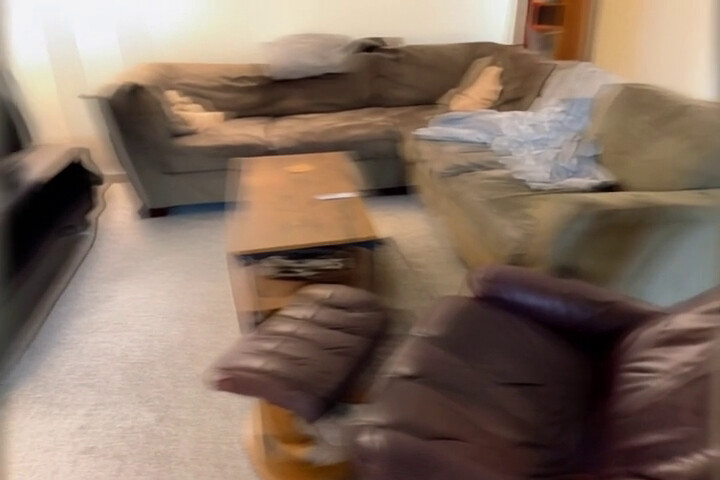} &
    \includegraphics[width=0.166\linewidth]{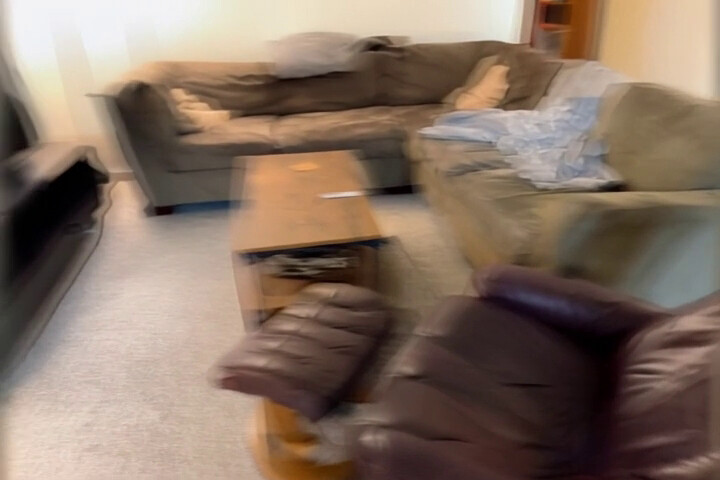} &
    \includegraphics[width=0.166\linewidth]{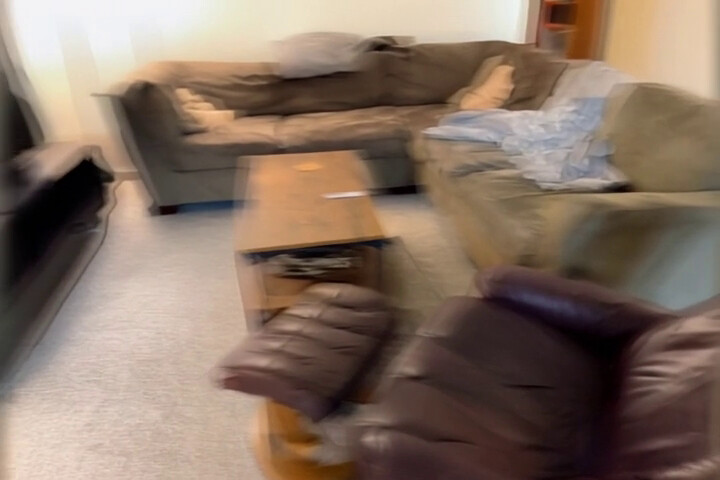} &
    \includegraphics[width=0.166\linewidth]{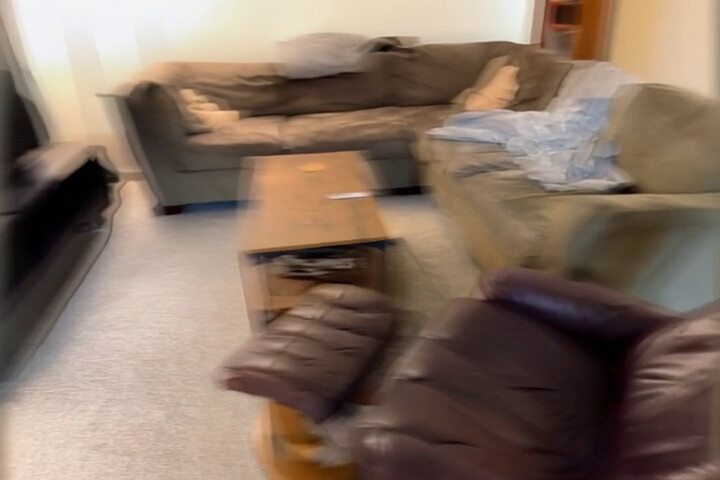} &
    \includegraphics[width=0.166\linewidth]{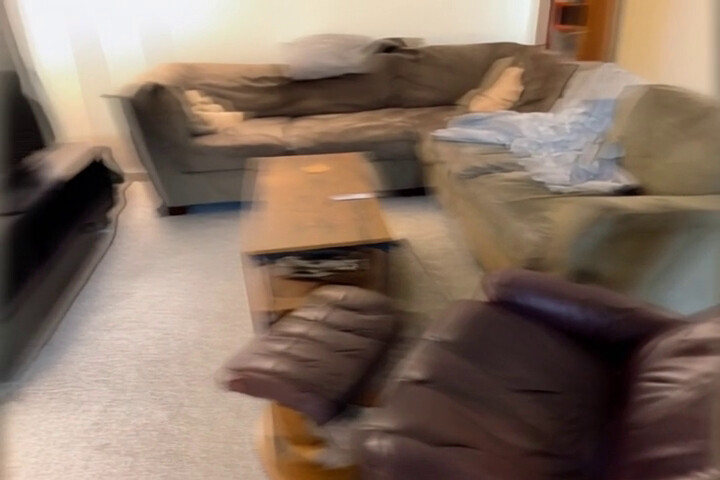} \\
    \raisebox{0ex}{\rotatebox{90}{ \shortstack{\tiny Flicker Input \\ (weak)}}} &
    \includegraphics[width=0.166\linewidth]{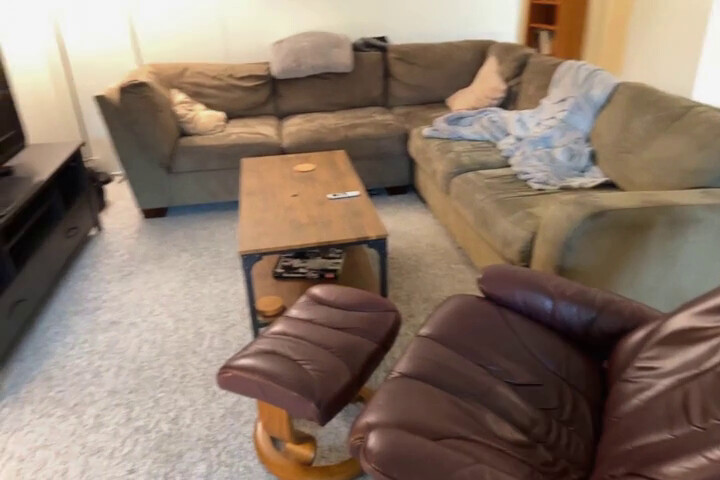} &
    \includegraphics[width=0.166\linewidth]{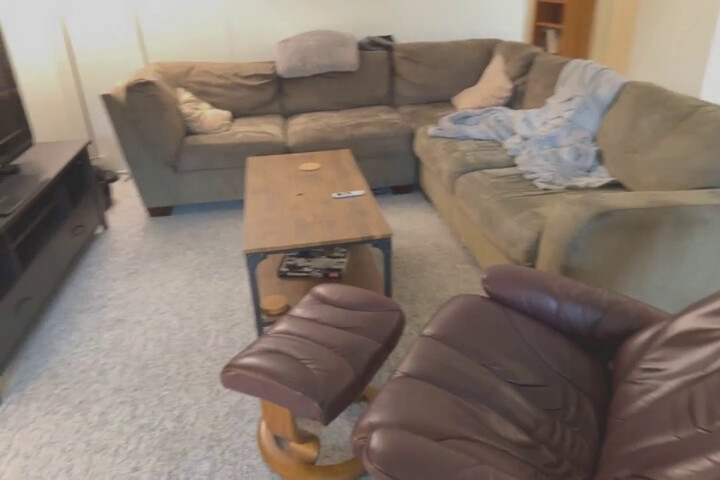} &
    \includegraphics[width=0.166\linewidth]{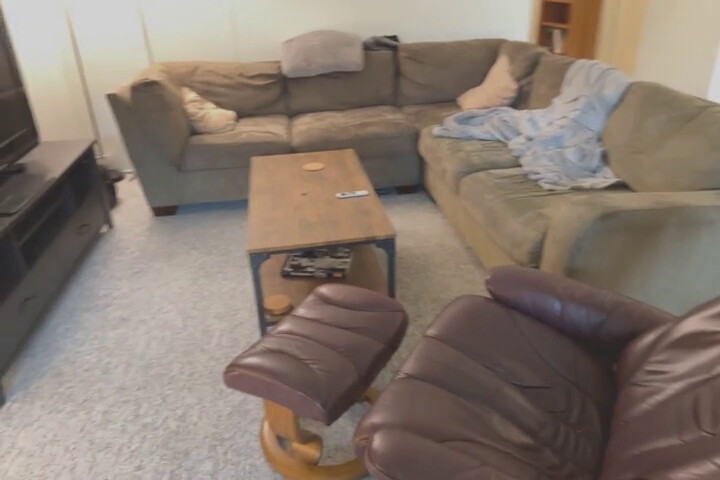} &
    \includegraphics[width=0.166\linewidth]{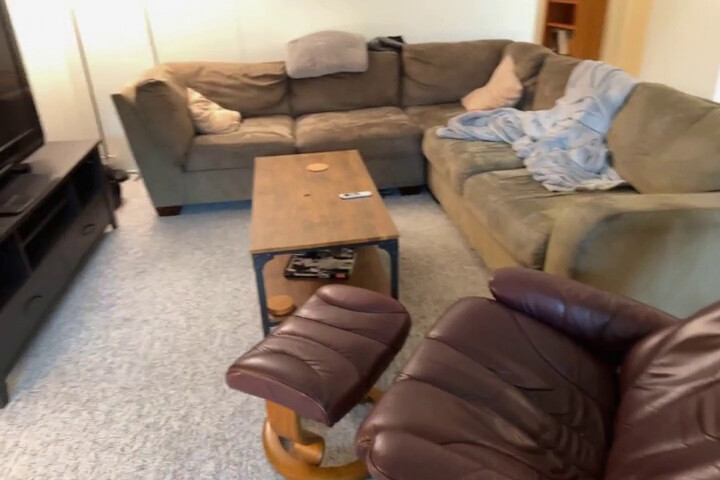} &
    \includegraphics[width=0.166\linewidth]{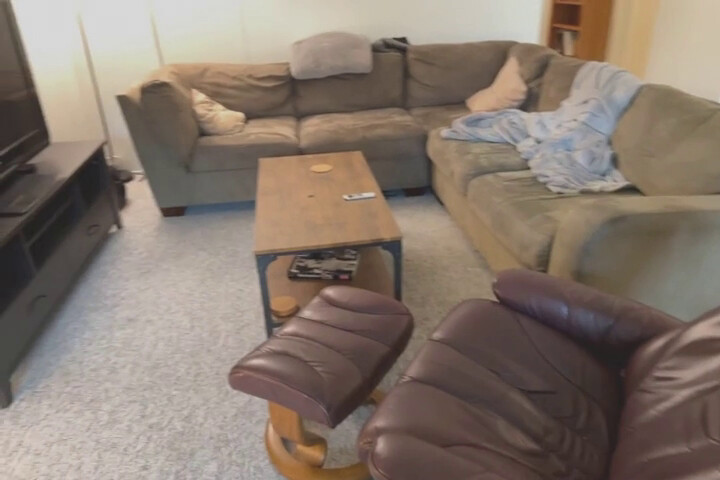} \\
    \raisebox{0ex}{\rotatebox{90}{ \shortstack{\tiny Flicker Pred \\ (weak)}}} &
    \includegraphics[width=0.166\linewidth]{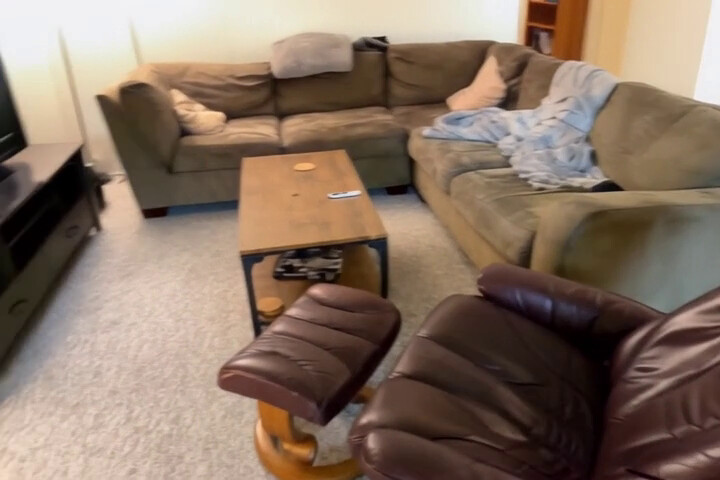} &
    \includegraphics[width=0.166\linewidth]{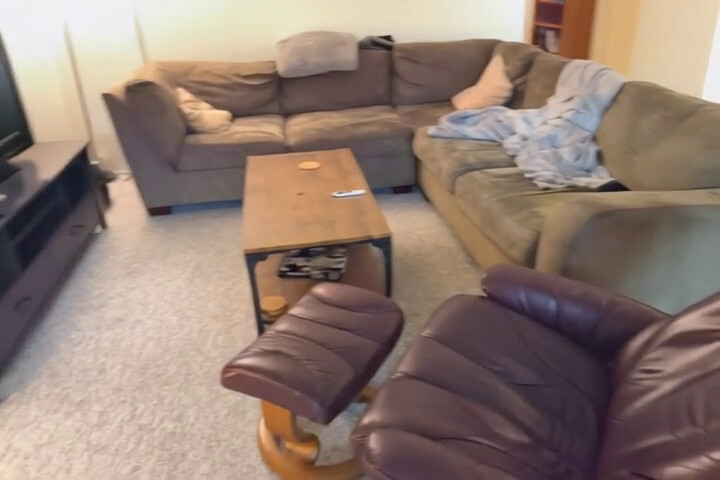} &
    \includegraphics[width=0.166\linewidth]{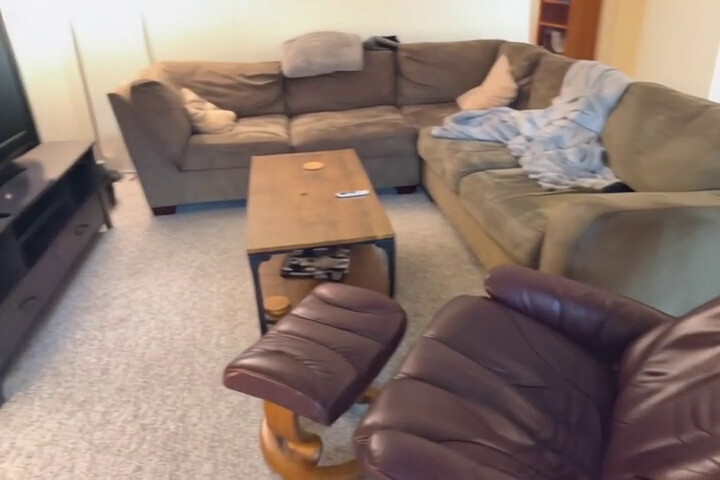} &
    \includegraphics[width=0.166\linewidth]{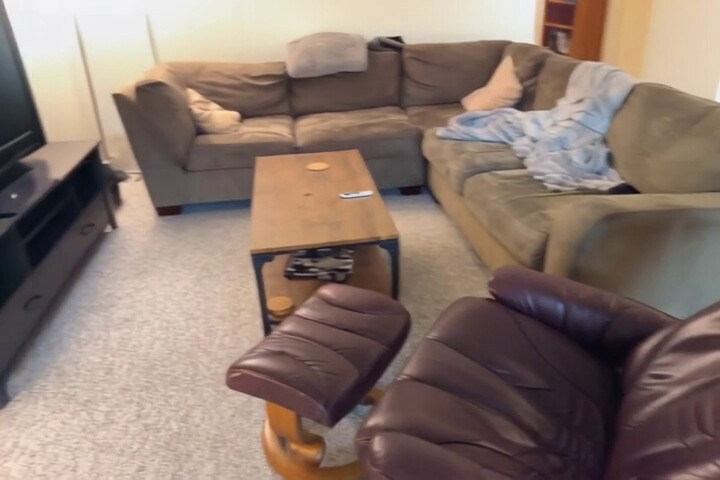} &
    \includegraphics[width=0.166\linewidth]{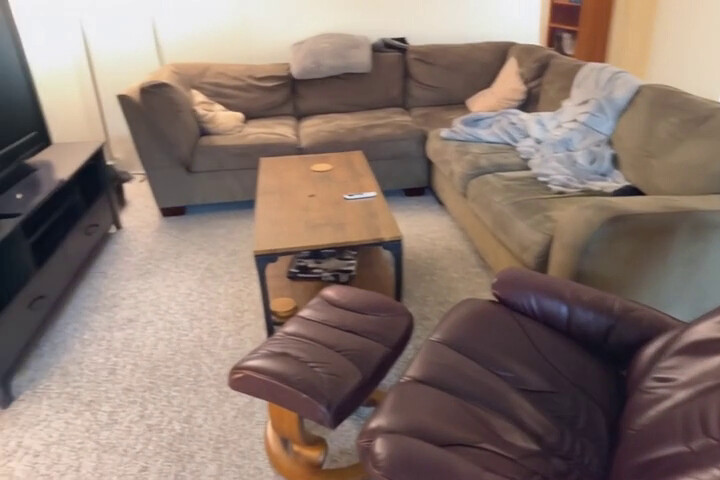} \\
    \raisebox{0ex}{\rotatebox{90}{ \shortstack{\tiny Flicker Input \\ (strong)}}} &
    \includegraphics[width=0.166\linewidth]{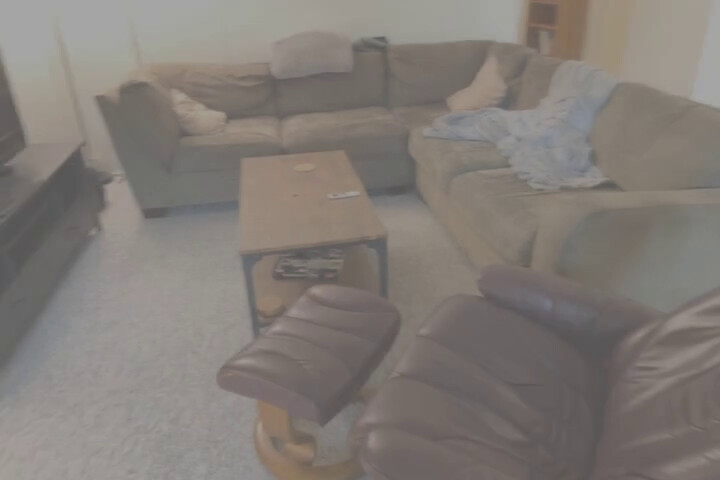} &
    \includegraphics[width=0.166\linewidth]{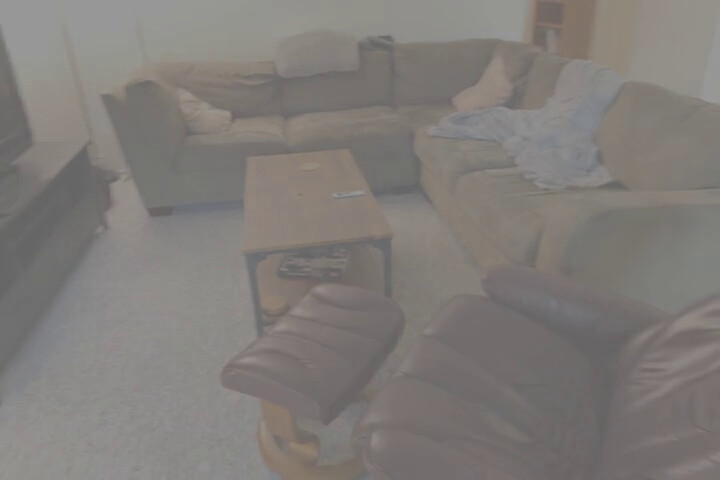} &
    \includegraphics[width=0.166\linewidth]{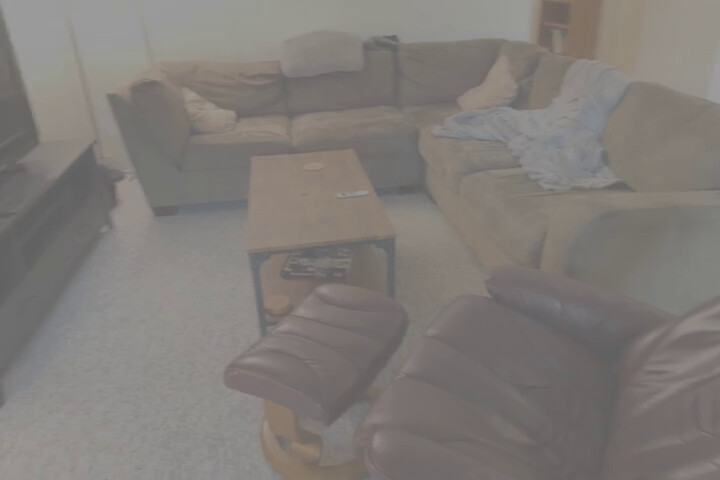} &
    \includegraphics[width=0.166\linewidth]{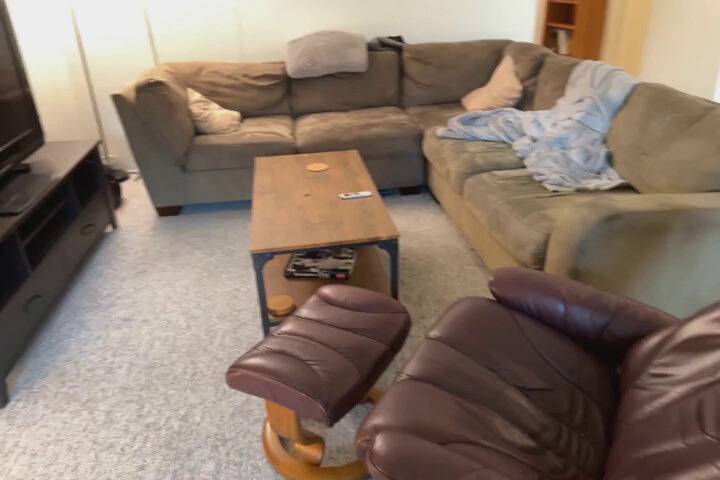} &
    \includegraphics[width=0.166\linewidth]{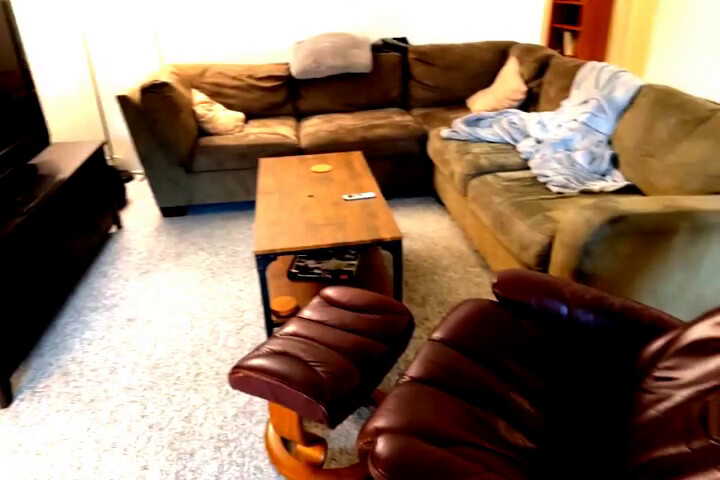} \\
    \raisebox{0ex}{\rotatebox{90}{ \shortstack{\tiny Flicker Pred \\ (strong)}}} &
    \includegraphics[width=0.166\linewidth]{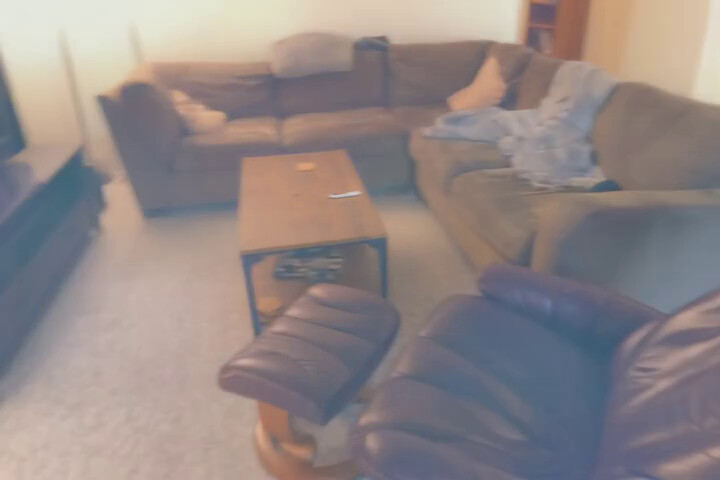} &
    \includegraphics[width=0.166\linewidth]{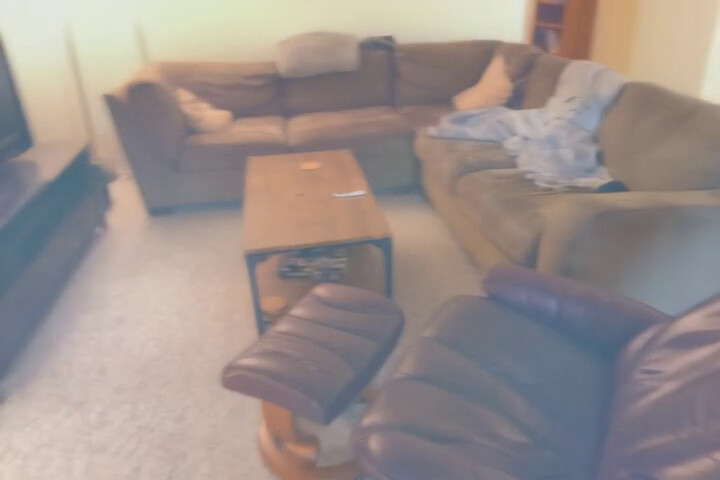} &
    \includegraphics[width=0.166\linewidth]{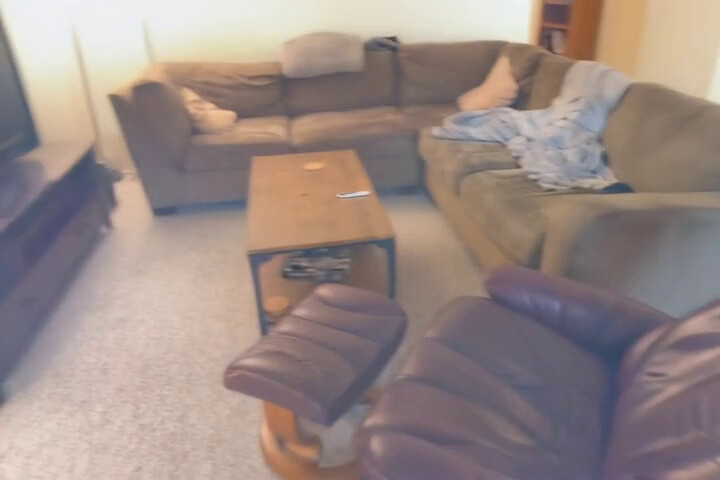} &
    \includegraphics[width=0.166\linewidth]{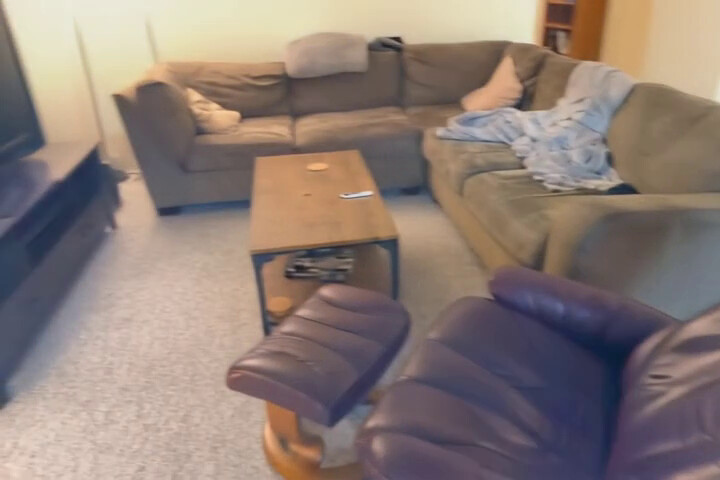} &
    \includegraphics[width=0.166\linewidth]{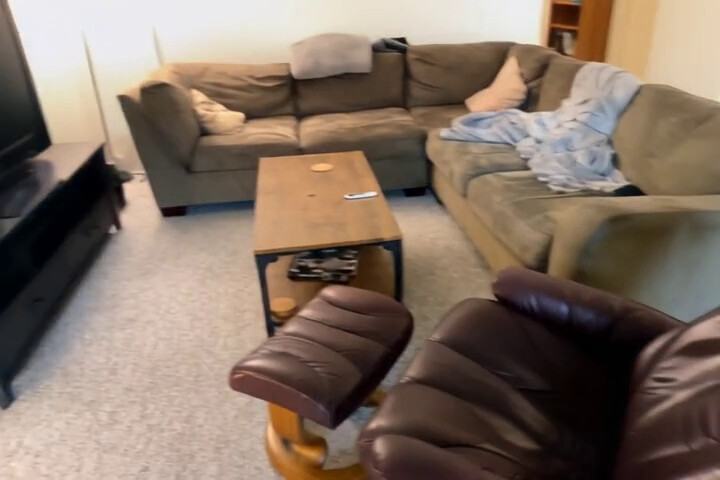} \\
    \raisebox{0ex}{\rotatebox{90}{ \shortstack{\tiny Occl Input \\ (weak)}}} &
    \includegraphics[width=0.166\linewidth]{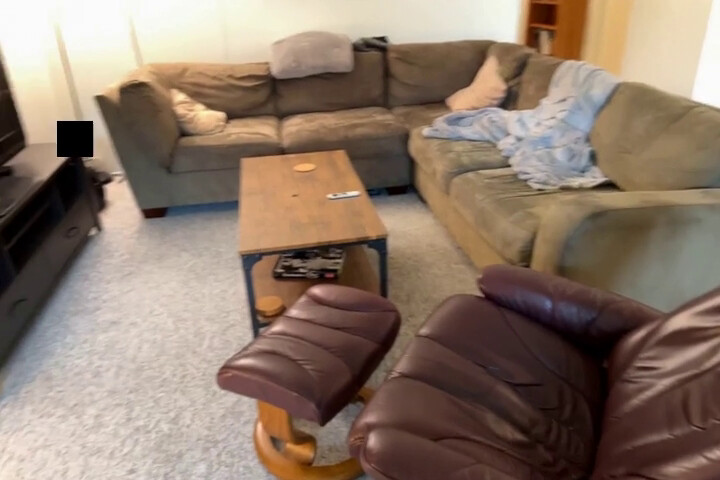} &
    \includegraphics[width=0.166\linewidth]{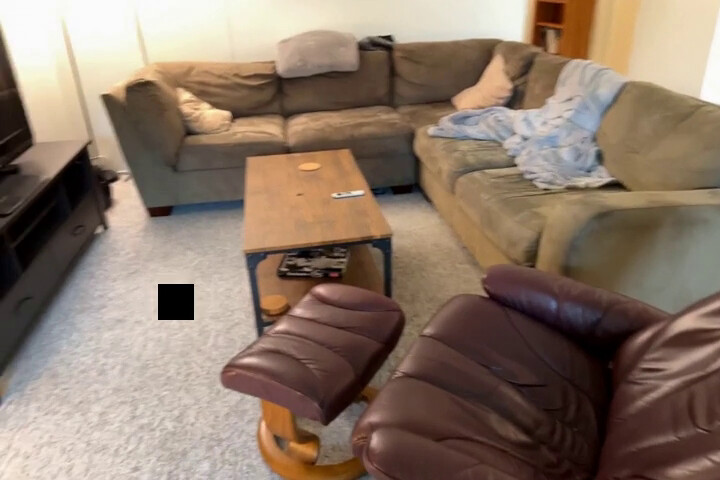} &
    \includegraphics[width=0.166\linewidth]{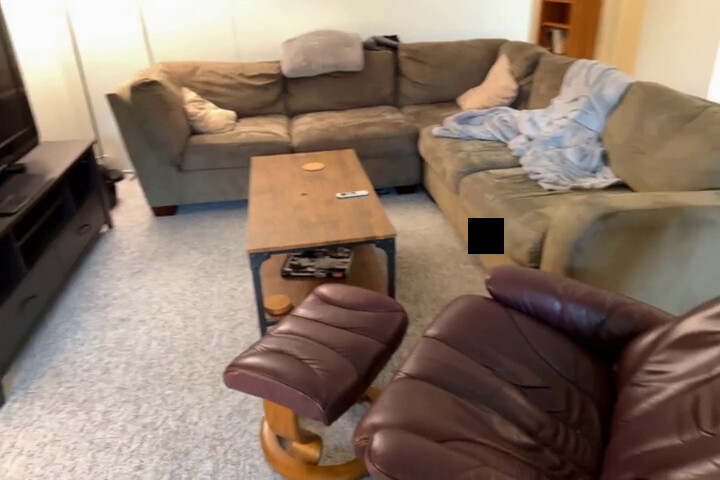} &
    \includegraphics[width=0.166\linewidth]{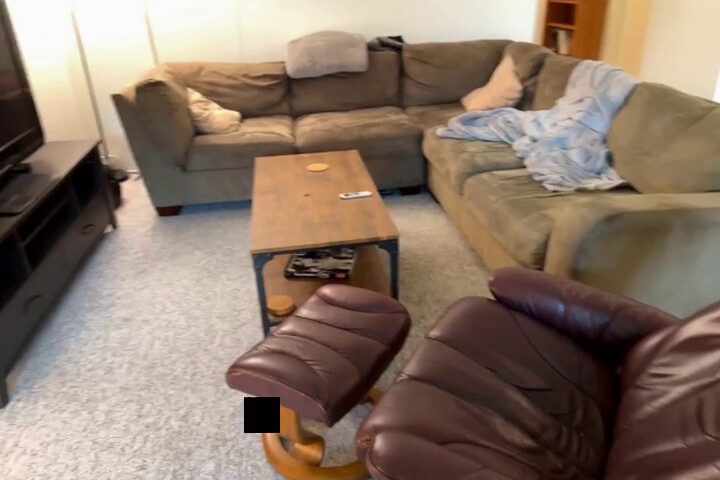} &
    \includegraphics[width=0.166\linewidth]{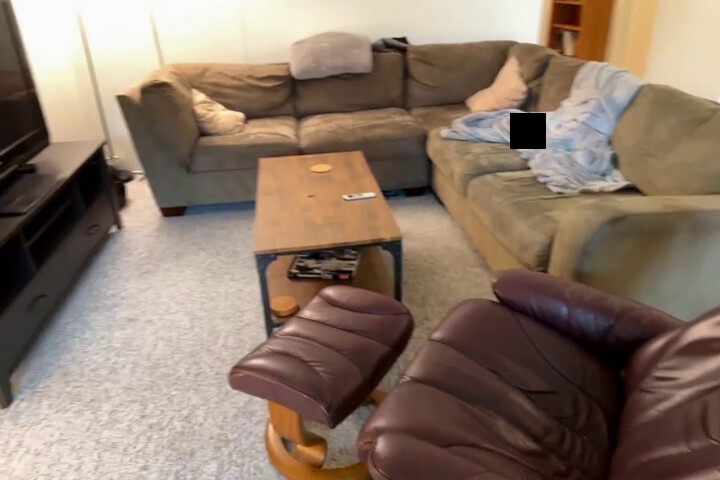} \\
    \raisebox{0ex}{\rotatebox{90}{ \shortstack{\tiny Occl Pred \\ (weak)}}} &
    \includegraphics[width=0.166\linewidth]{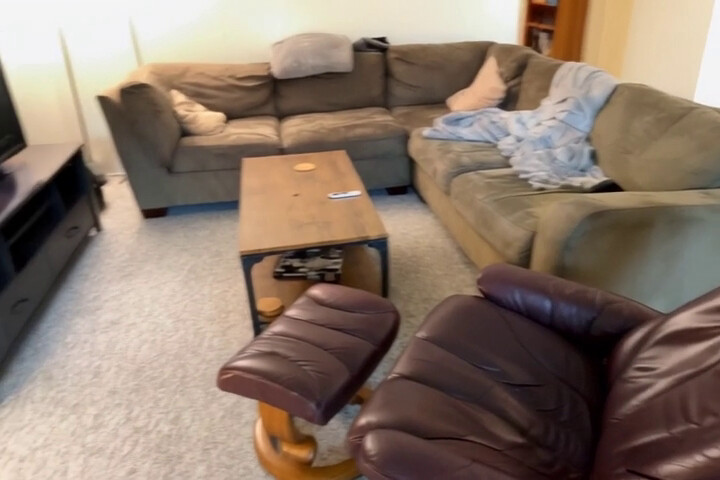} &
    \includegraphics[width=0.166\linewidth]{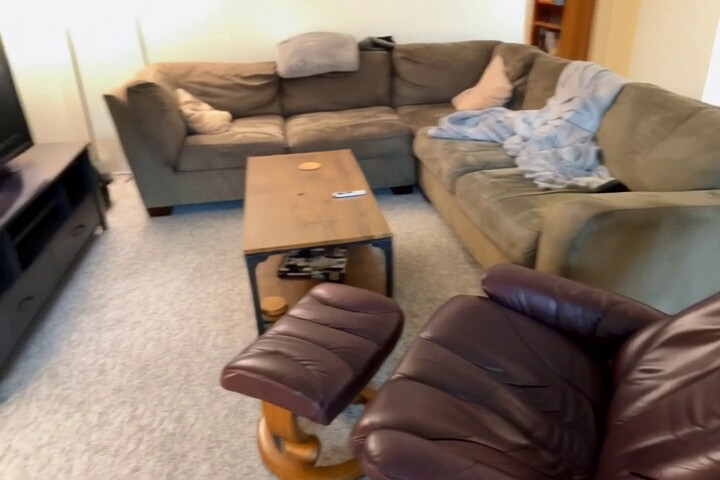} &
    \includegraphics[width=0.166\linewidth]{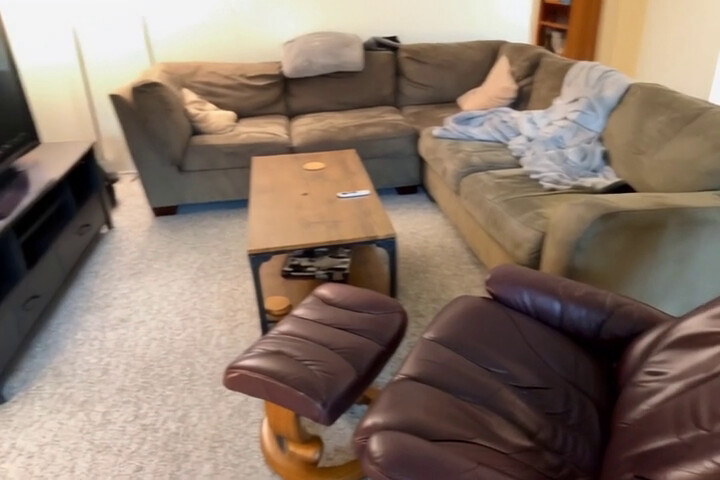} &
    \includegraphics[width=0.166\linewidth]{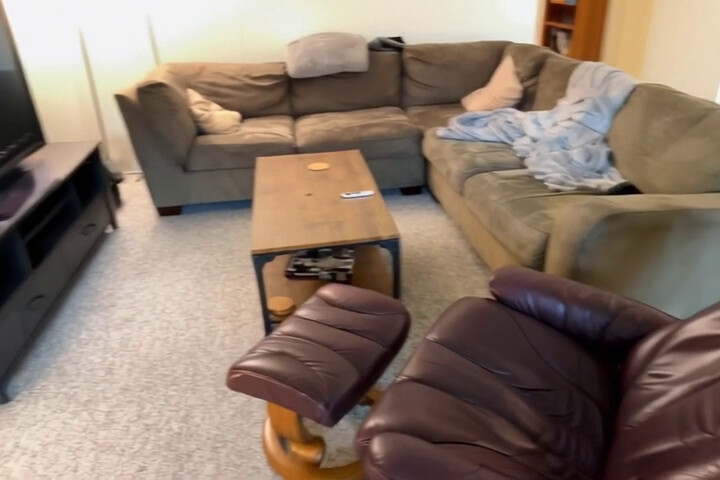} &
    \includegraphics[width=0.166\linewidth]{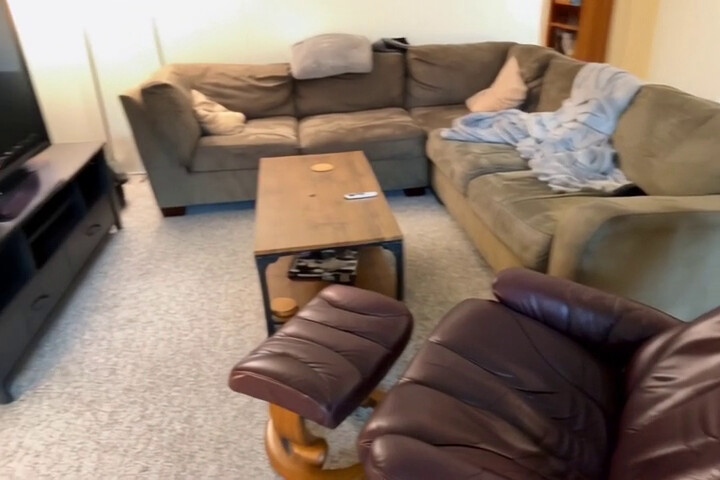} \\
    \raisebox{0ex}{\rotatebox{90}{ \shortstack{\tiny Occl Input \\ (strong)}}} &
    \includegraphics[width=0.166\linewidth]{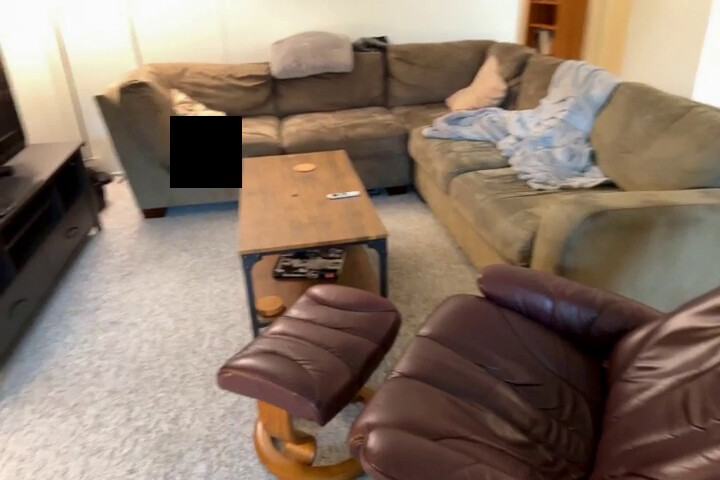} &
    \includegraphics[width=0.166\linewidth]{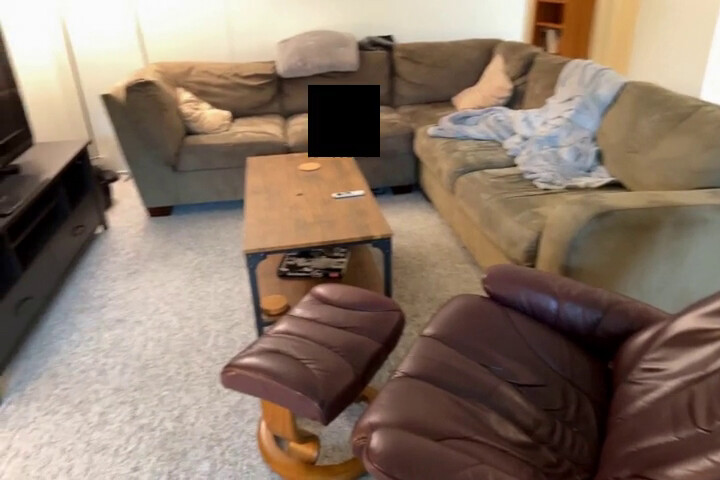} &
    \includegraphics[width=0.166\linewidth]{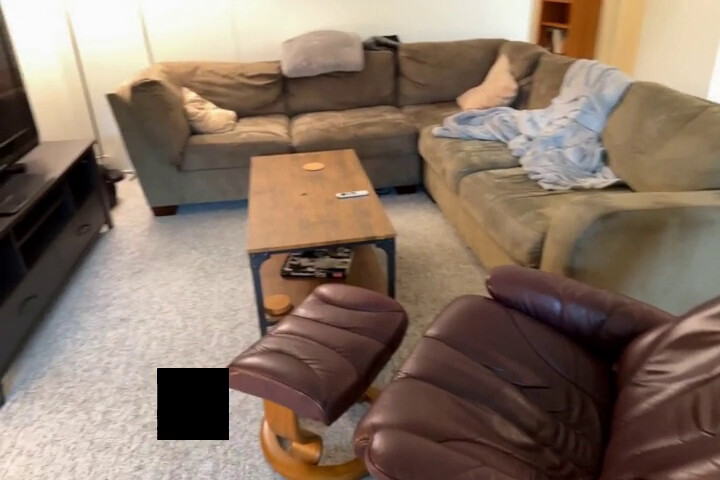} &
    \includegraphics[width=0.166\linewidth]{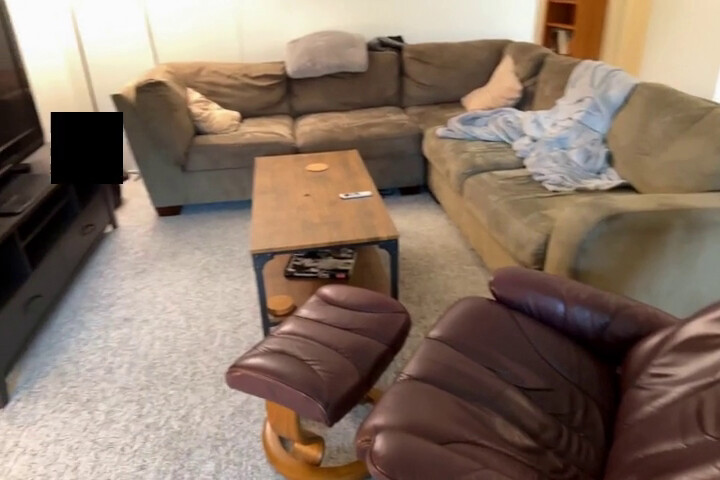} &
    \includegraphics[width=0.166\linewidth]{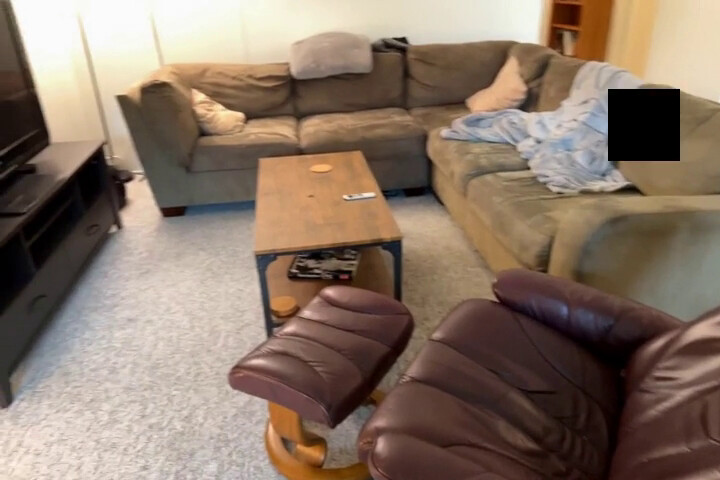} \\
    \raisebox{0ex}{\rotatebox{90}{ \shortstack{\tiny Occl. Pred \\ (strong)}}} &
    \includegraphics[width=0.166\linewidth]{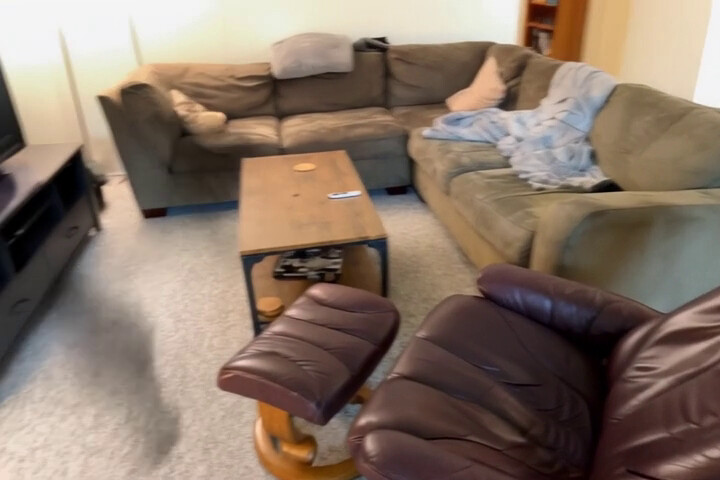} &
    \includegraphics[width=0.166\linewidth]{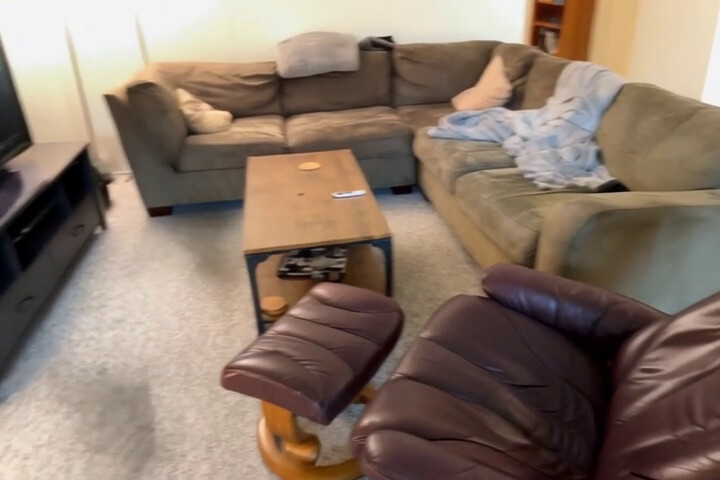} &
    \includegraphics[width=0.166\linewidth]{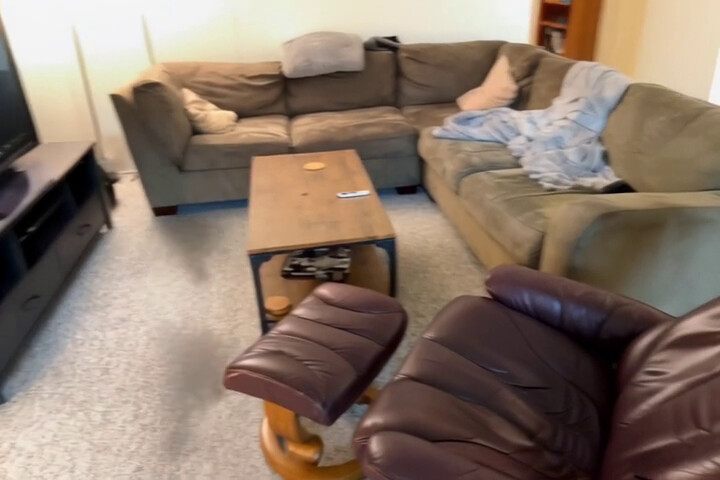} &
    \includegraphics[width=0.166\linewidth]{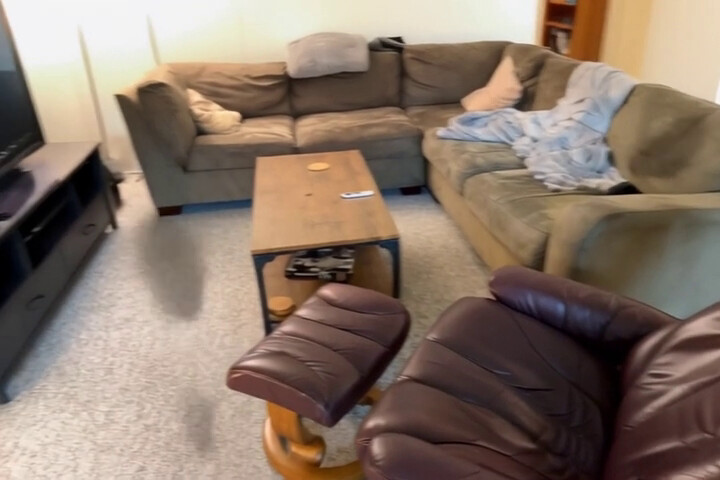} &
    \includegraphics[width=0.166\linewidth]{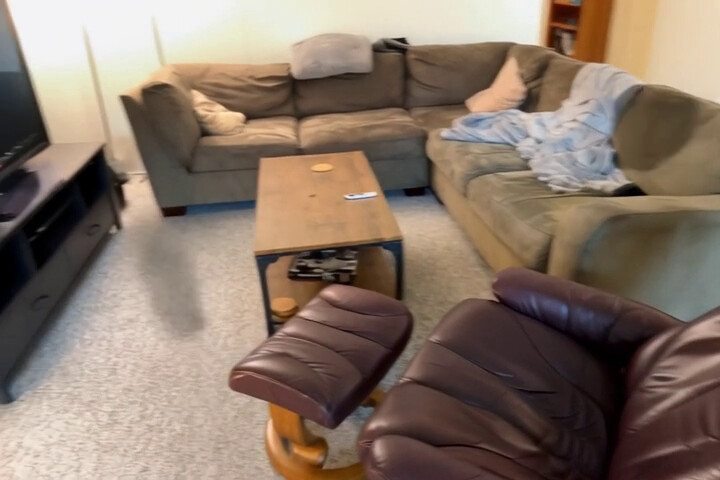} \\

  \end{tabular}
  \caption{Ablation of instability of frame-by-frame feature extraction}
  \label{fig:stabilty}
  \vspace{-0.5cm}
\end{figure}

The diffusion backbone is generally capable of mitigating and smoothing many of these artifacts, provided that similar degradations are represented in the training data. While our training pipeline does not explicitly incorporate augmentations targeting these effects, the ScanNet++(iPhone) videos used for training naturally contain artifacts such as flickering and motion blur. As shown in \autoref{fig:stabilty}, temporal flickering is substantially reduced in the generated results, suggesting that the model learns a degree of temporal stabilization. In contrast, motion blur is often preserved and reproduced, reflecting the statistics of the training data, where such artifacts are frequently present. Similarly, errors caused by occlusions can be partially smoothed out, although the extent of this correction depends on the severity and duration of the occlusion.

\subsection{3D Reconstruction Quality}
\begin{figure}[htbp]
  \centering
  \setlength{\tabcolsep}{0.0pt}
  \renewcommand{\arraystretch}{0.0}
  \scriptsize
  \begin{tabular}{@{}c c c c c c@{}}
    & \textbf{GT} & \textbf{Pred. sparse} & \textbf{Cond. sparse} & \textbf{Pred. I2V} & \textbf{Cond. I2V} \\
   \raisebox{5ex}{\rotatebox{90}{\tiny $t{=}1$}} &
       \includegraphics[width=0.20\linewidth]{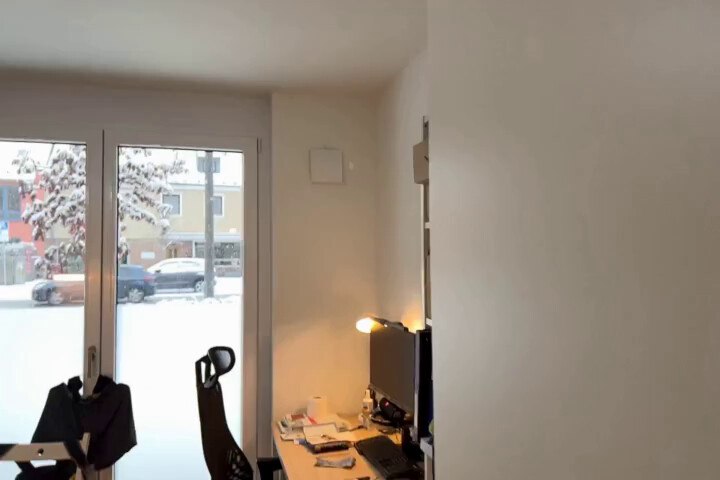} &
    \includegraphics[width=0.20\linewidth]{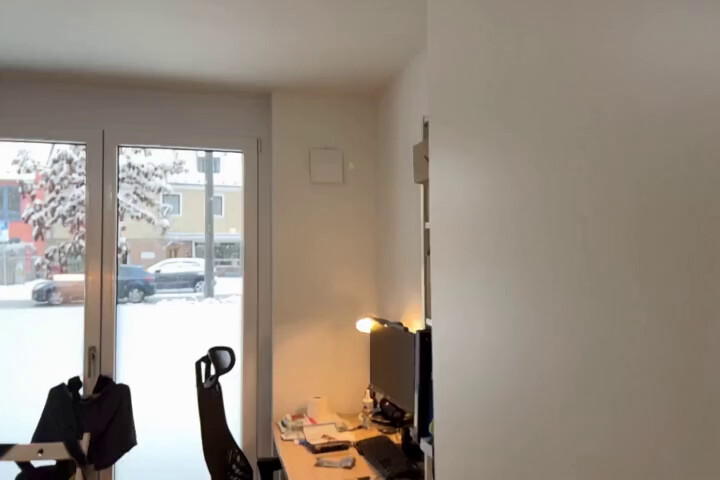} &
       \includegraphics[width=0.20\linewidth]{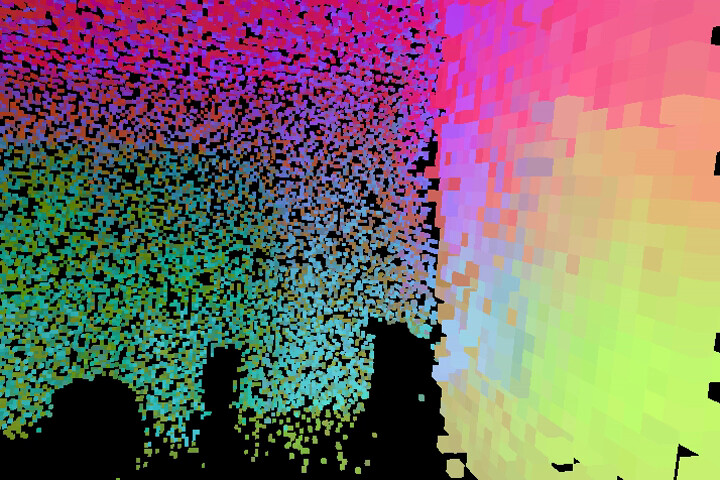} &
          \includegraphics[width=0.20\linewidth]{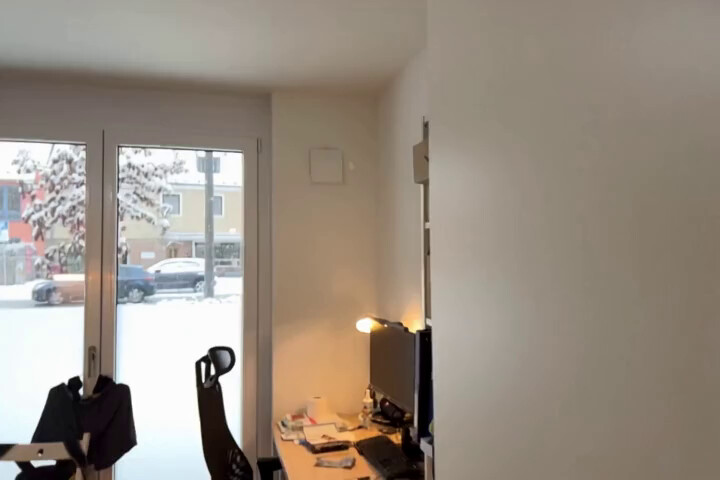} &
     \includegraphics[width=0.20\linewidth]{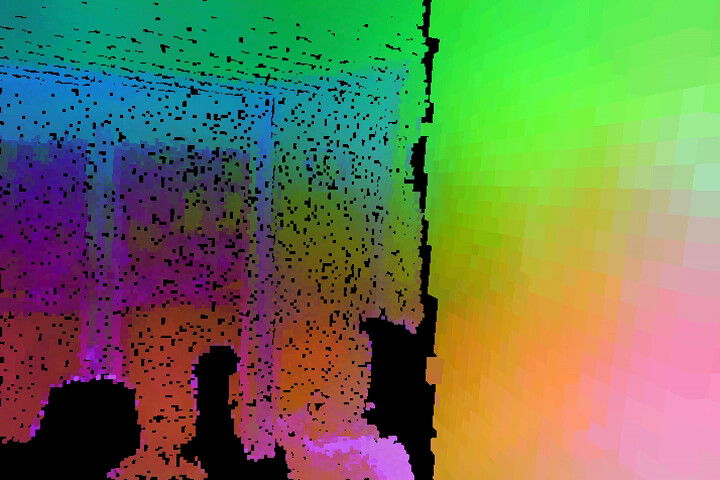}
    \\
    
    \raisebox{3.9ex}{\rotatebox{90}{\tiny $t{=}35$}} &
      \includegraphics[width=0.20\linewidth]{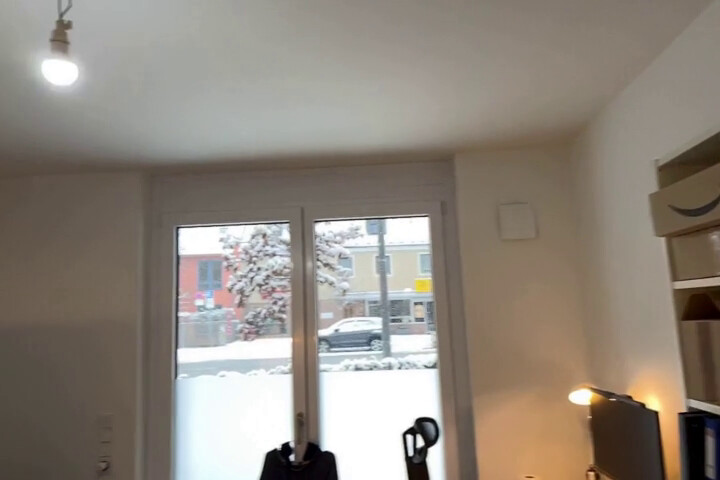} &
   \includegraphics[width=0.20\linewidth]{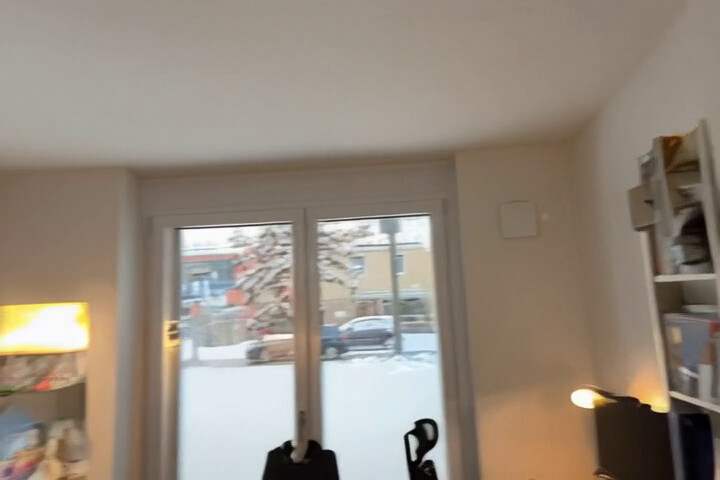} &
      \includegraphics[width=0.20\linewidth]{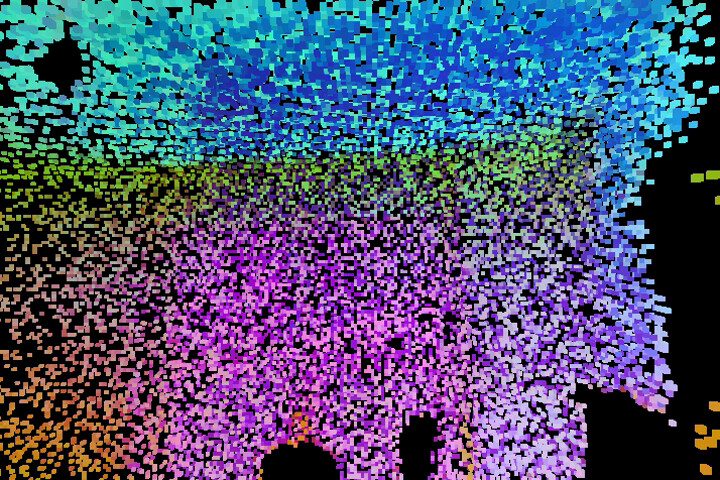} &
         \includegraphics[width=0.20\linewidth]{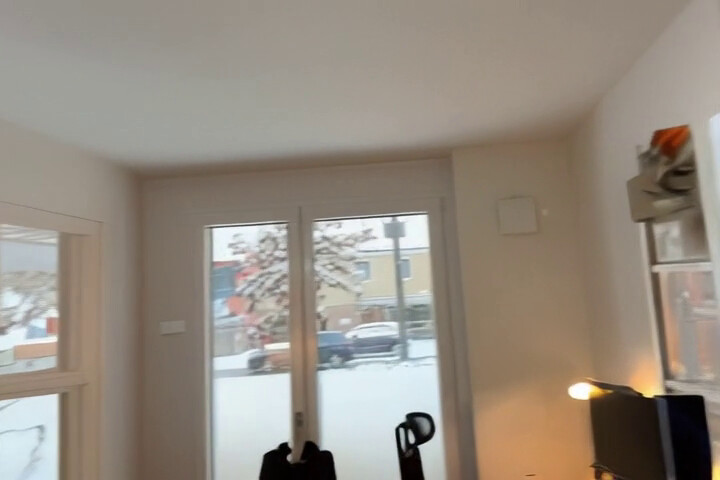} &
    \includegraphics[width=0.20\linewidth]{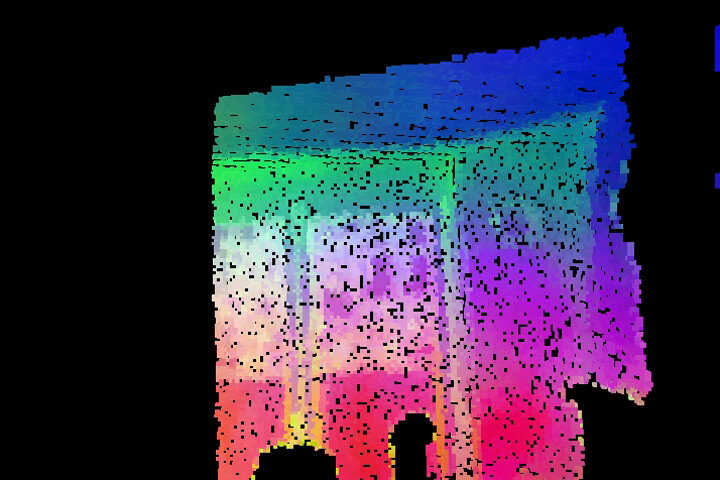}
\\
   \raisebox{4ex}{\rotatebox{90}{\tiny $t{=}49$}} &
     \includegraphics[width=0.20\linewidth]{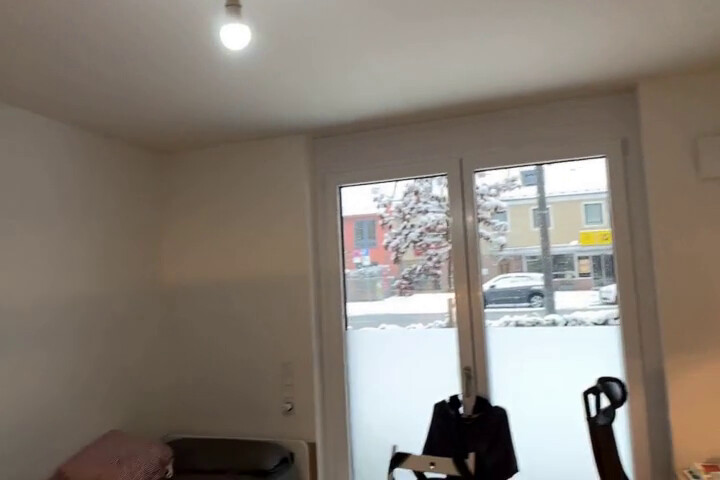} &
  \includegraphics[width=0.20\linewidth]{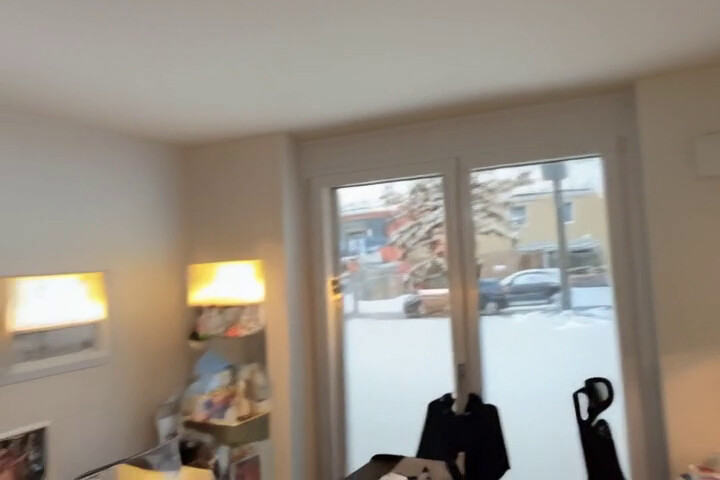} &
     \includegraphics[width=0.20\linewidth]{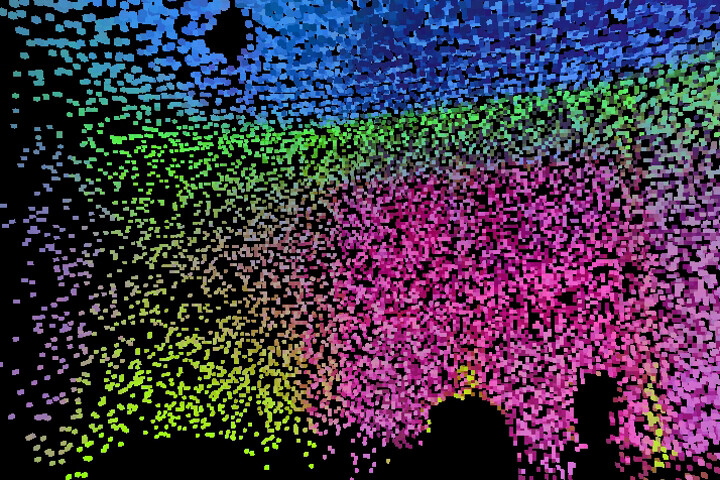} &
        \includegraphics[width=0.20\linewidth]{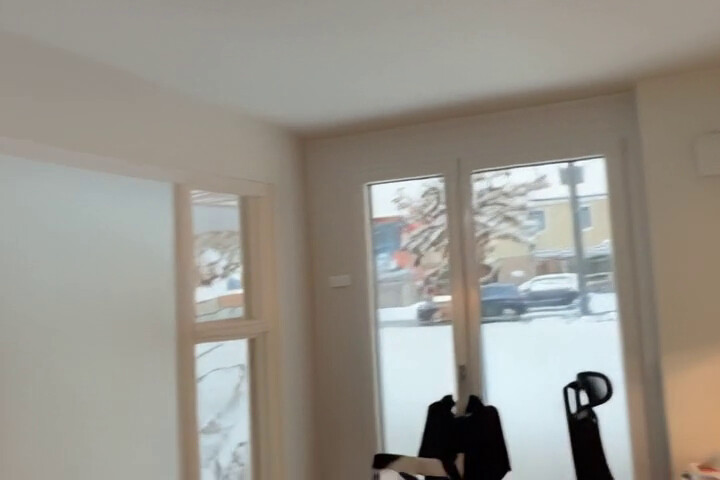} &
   \includegraphics[width=0.20\linewidth]{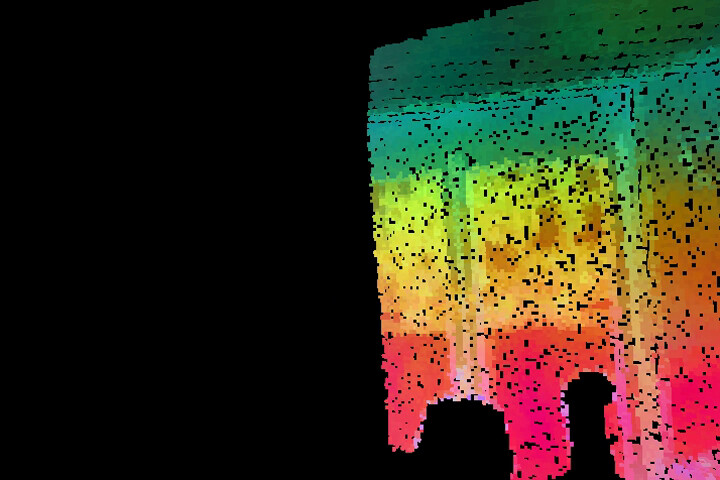}
  \end{tabular}
  \caption{Qualitative comparison using sparse and incomplete voxel conditioning (black regions will be masked), evaluating both sparse settings and single-frame I2V reconstruction scenario.}
  \label{fig:3d_robustness}
  \vspace{-0.5cm}
\end{figure}

\autoref{fig:3d_robustness} demonstrates the robustness of our approach under challenging input conditions. While our method is primarily tailored for dense 3D reconstruction and scene transformation, rather than novel scene generation from sparse observations, it does not suffer from severe degradation when presented with limited input data. Notably, the model can handle single-image-to-video settings (similar to \cite{ren2025gen3c}) and scenes with significant missing geometry, producing coherent and plausible hallucinations.

\subsection{ControlNet Layer Ablation}
We evaluate the reconstruction quality by varying the number of injected ControlNet layers. As expected, the reconstruction quality slightly decreases when injecting only $4$ layers (FID of  $39.82$) and LPIPS of $0.175$)) compared to injecting a higher number of layers (FID of $36.67$) and LPIPS of $0.136$)). The baseline version, which serves as a middle ground, achieves an FID of $37.67$) and an LPIPS of $0.165$). Overall, these results demonstrate that injecting even a few layers provides a strong conditioning signal.

\subsection{DINO Version Ablation}
As shown in \autoref{tab:dino_version}, distilling from DINO models of different sizes has only a marginal impact on the final reconstruction performance of Control-DINO. Given the comparable reconstruction results across model scales, we adopt the smallest model DINOv3-S in our Control-DINO framework due to its lower computational cost and improved efficiency.

\begin{table}[t]
\centering
\caption{Ablation of different DINOv3 Version used for Control-DINO on a small ScanNet++ subset, trained 1 epoch.}
\label{tab:dino_version}
\resizebox{\linewidth}{!}{
\begin{tabular}{lccc}
\toprule
Version & PSNR $\uparrow$ & SSIM $\uparrow$ & LPIPS $\downarrow$ \\
\midrule
DINOv3 ViT-S (Ours) & 27.51 dB & 0.9041 & 0.0991 \\
DINOv3 ViT-B  & 26.96 dB & 0.9070 & 0.0936 \\
DINOv3 ViT-L & 27.54 dB & 0.9064 & 0.0913  \\
DINOv2 ViT-S & 24.49 dB & 0.8841 & 0.1016  \\
\bottomrule
\end{tabular}
}
\end{table}

\end{document}